\documentclass{article}

\usepackage[dvipsnames]{xcolor}
\definecolor{darkpink}{RGB}{204, 0, 102}
\usepackage{microtype}
\usepackage{graphicx}
\usepackage{subfigure}
\usepackage{booktabs} 
\usepackage{xspace}
\usepackage{multirow}
\usepackage{mfirstuc}
\usepackage{wrapfig}
\usepackage{caption}

\usepackage{hyperref}

\usepackage[accepted]{icml2024}

\usepackage{amsmath}
\usepackage{amssymb}
\usepackage{mathtools}
\usepackage{amsthm}
\usepackage{bm}

\usepackage[capitalize,noabbrev]{cleveref}

\usepackage{enumitem}
\usepackage{makecell}
\usepackage[most]{tcolorbox}
\usepackage{wrapfig}

\renewcommand{\vec}[1]{\ensuremath{\boldsymbol{#1}}}

\DeclareMathOperator*{\ulp}{ulp}

\DeclareMathOperator*{\edq}{EDQ}
\DeclareMathOperator*{\oom}{OOM}
\DeclareMathOperator*{\expect}{\mathbb{E}}

\DeclareFixedFont{\ttb}{T1}{txtt}{b}{n}{9} 
\DeclareFixedFont{\ttm}{T1}{txtt}{m}{n}{9}  
\usepackage{color}
\definecolor{deepblue}{rgb}{0,0,0.5}
\definecolor{deepred}{rgb}{0.6,0,0}
\definecolor{deepgreen}{rgb}{0,0.5,0}

\definecolor{light-gray}{gray}{0.95}
\definecolor{bole}{rgb}{0.35, 0.22, 0.21}
\definecolor{brown(web)}{rgb}{0.65, 0.16, 0.16}
\usepackage{listings}

\newcommand\pythonstyle{\lstset{
language=Python,
basicstyle=\ttm\color{bole},
morekeywords={self, torch, size, 
            Size, fc, x, y, tensor, f, nc, grad, FloatTensor, HalfTensor, DoubleTensor, MCModel },  
keywordstyle=\ttm\color{deepblue},
emph={MyClass,__init__, Two_Sum, Split, Two_Prod, 
      MCModule, MCOptim,  MCLinear, MCEmbedding, 
      add, subtract, multiply, divide, exp, 
      dot, mv, mm, matmul, 
      MCSGD, MCAdam,
      Two_Prod_fma, Renormalize, Simple_Renorm,  
      Grow_ExpN, ScalingN, Add_MCN, Div_MCN, ScalingN, DivN,
      DivN, Mult_MCN, Mult_MCN_Slow, Exp_MCN, 
      Square_MCN, Dot_MCN, MV_MCN, MM_MCN, BMM_MCN,
      Exp−MCN, Square−MCN, Grow−ExpN, Add−MCN, Div−MCN, Mul−MCN,
      4DMM_MCN,  AddMM_MCN,  Matmul_MCN, MC-Linear},          
emphstyle=\ttm\color{brown(web)},    
stringstyle=\color{deepgreen},
frame=tb,                         
showstringspaces=false,
backgroundcolor=\color{light-gray}
}}

\lstnewenvironment{python}[1][]
{
\pythonstyle
\lstset{#1}
}
{}

\newcommand\pythoninline[1]{{\pythonstyle\lstinline!#1!}}
\theoremstyle{plain}
\newtheorem{theorem}{Theorem}[section]

\theoremstyle{definition}
\newtheorem{definition}[theorem]{Definition}

\theoremstyle{remark}
\newtheorem{remark}[theorem]{Remark}

\newcommand{\strname}{\textsc{Collage}\xspace}

\usepackage[textsize=tiny]{todonotes}

\icmltitlerunning{\strname: Light-Weight Low-Precision Strategy for LLM Training}

\begin{document}

\twocolumn[
\icmltitle{\strname: Light-Weight Low-Precision Strategy for LLM Training}



\icmlsetsymbol{wortAtAmazon}{*}
\icmlsetsymbol{majorContribution}{$\dagger$}
\begin{icmlauthorlist}
\icmlauthor{Tao Yu}{wortAtAmazon,cornell}
\icmlauthor{Gaurav Gupta}{majorContribution,ail}
\icmlauthor{Karthick Gopalswamy}{annap}
\icmlauthor{Amith Mamidala}{annap}
\icmlauthor{Hao Zhou}{sagem}
\icmlauthor{Jeffrey Huynh}{annap}
\icmlauthor{Youngsuk Park}{aire}
\icmlauthor{Ron Diamant}{annap}
\icmlauthor{Anoop Deoras}{ail}
\icmlauthor{Luke Huan}{ail}
\end{icmlauthorlist}

\icmlaffiliation{cornell}{Cornell University, Ithaca, NY}
\icmlaffiliation{ail}{AWS AI Labs, Santa Clara, CA}
\icmlaffiliation{annap}{AWS Annapurna Labs, Cupertino, CA}
\icmlaffiliation{sagem}{AWS Sagemaker, Santa Clara, CA}
\icmlaffiliation{aire}{AWS AI Research and Education, Santa Clara, CA}

\icmlcorrespondingauthor{Gaurav Gupta}{gauravaz@amazon.com}

\icmlkeywords{Machine Learning, ICML}

\vskip 0.3in
]


\printAffiliationsAndNotice{}  

\begin{abstract}
Large models training is plagued by the intense compute cost and limited hardware memory. 
A practical solution is low-precision representation but is troubled by loss in numerical accuracy and unstable training rendering the model less useful. We argue that low-precision floating points can perform well provided the error is properly compensated at the critical locations in the training process.
We propose \strname which utilizes multi-component float representation in low-precision to accurately perform operations with numerical errors accounted. 
To understand the impact of imprecision to training, 
we propose a simple and novel metric which tracks the lost information during training as well as differentiates various precision strategies.
Our method works with commonly used low-precision such as half-precision ($16$-bit floating points) and can be naturally extended to work with even lower precision such as $8$-bit. 
Experimental results show that pre-training using \strname removes the requirement of using $32$-bit floating-point copies of the model and attains similar/better training performance compared to $(16, 32)$-bit mixed-precision strategy, with up to $3.7\times$ speedup and $\sim 15\%$ to $23\%$ less memory usage in practice.
\end{abstract}

\section{Introduction}
\label{sec:introduction}
Recent success of large models using transformers backend has gathered the attention of community for generative language modeling (GPT-4 \citep{openai2023gpt4}, LaMDA \citep{thoppilan2022lamda}, LLaMa \citep{touvron2023llama}), image generation (e.g., Dall-E \cite{betker2023improving}), speech generation (such as Meta voicebox, OpenAI jukebox \cite{le2023voicebox, dhariwal2020jukebox}), and multimodality (e.g. gemini \cite{geminiteam2023gemini}) motivating to further scale such models to larger size and context lengths. However, scaling models is prohibited by the hardware memory and also incur immense compute cost in the distributed training, such as $\sim$1M GPU-hrs for LLaMA-$65$B \cite{touvron2023llama}, thus asking the question whether large model training could be made efficient while maintaining the accuracy?

Previous works have attempted to reduce the memory consumption and run models more efficiently by reducing precision of the parameter's representation, at training time \cite{zhang2022opt, kuchaiev2018mixedprecision, kuchaiev2019nemo, peng2023fp8lm} and post-training inference time 
\cite{courbariaux2016binarized, rastegari2016xnornet, MLSYS2019_c443e9d9}. The former one is directly relevant to our work using \textit{low-precision storages} at training time, but it suffers from issues such as numerical inaccuracies and narrow representation range. Researchers developed algorithms such as loss-scaling and mixed-precision \cite{micikevicius2018mixed, shoeybi2020megatronlm} to overcome these issues. Existing algorithms still face challenges in terms of memory efficiency as they require the presence of high-precision clones and computations in optimizations. One critical limitation of all the aforementioned methods is that such methods keep the ``standard format" for floating-points during computations and lose information with a reduced precision.

In this work, we elucidate that in the setting of low-precision (for example, 16-bit or lower) for floating point, using alternative representations such as multiple-component float (MCF) \cite{yu2022mctensor} helps in making reduced precision accurate in computations. MCF was introduced as  `expansion' \cite{priest1991Arithmetic} in C++ \cite{hida2008Cpp} and hyperbolic spaces \cite{Yu2021MCT} representation. Recently, MCF has been integrated with PyTorch in the MCTensor library \cite{yu2022mctensor}.

\begin{figure*}
\centering
\captionsetup{font=small}
\includegraphics[width=0.75\linewidth]{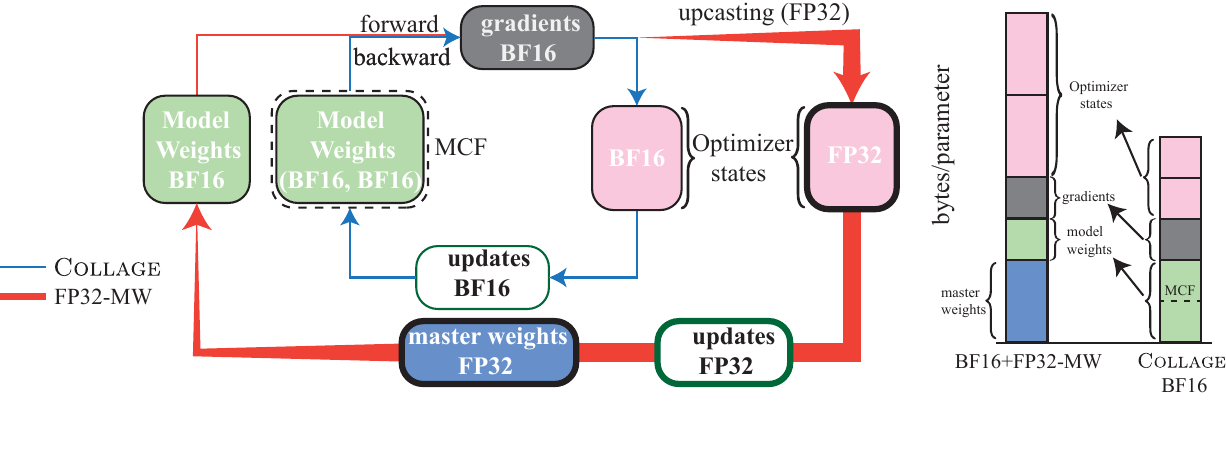}
\vspace*{-10pt}
\caption{\textbf{Left:} \strname uses a strict low-precision floating-point (such as BF16) optimization loop without ever needing to upcast to FP32 like in the mixed-precision with master weights (red thick loop). The model weights in \strname are represented as multi-component float (MCF) instead of ``standard float". \textbf{Right:} Total bytes/parameter savings for \strname without taking the FP32 upcasting route. The memory savings and uncompromising use of low-precision results in speed-up as seen in Table\,\ref{tab:speedup}.
}
\label{fig:main_algo_descr}
\end{figure*}

We propose \strname\footnote{Inspired from the multi-component nature of the algorithm.}, a new approach to deal with floating-point errors in low-precision to make LLM training accurate and efficient. Our primary objective is to develop a training loop with storage strict in low-precision without a need to maintain high-precision clones. We realize that when dealing with low-precision floats (such as Bfloat16), the ``standard" representation is not sufficient to avoid rounding errors which \textit{should not be ignored}. To solve these issues, we rather apply an existing technique of MCF to represent floats which (i) either encounters drastic rounding effects, (ii) the scale of the involved floats has a wide range such that arithmetic operations were lost. We implemented \strname as a plugin to be easily integrated with the well-known optimizers such as AdamW \cite{loshchilov2017decoupled} (extensions to SGD \cite{ruder2017overview} are straight-forward) using low-precision storage \& computations. By turning the optimizer to be more \textit{precision-aware}, even with additional low-precision components in MCF, we obtain faster training (upto $3.7\times$ better train throughput on $6.7$B GPT model, Table\,\ref{tab:speedup}) and also have less memory foot-print due to strict low-precision floats (see \figurename\,\ref{fig:main_algo_descr}\,right), compared to the most advanced mixed precision baseline.

We have developed a novel metric called ``effective descent quality" to trace the lost information in the optimizer model update step. Due to rounding and lost arithmetic (see definition in Section\,\ref{ssec:imprecision}), the effective update applied to the model is different from the intended update from optimizer, thus distracting the model training trajectory. Tracing this metric during the training enables to compare different precision strategies at a fine-grained level (see \figurename\,\ref{fig:metrics-bert-base-uncased}\,right).

In this work, we answer the critical question of where (which computation) with low-precision during training is severely impacting the performance and why?
The main contributions are outlined as follows.
\begin{itemize}[noitemsep,topsep=0pt]
    \item We provide \strname as a \textbf{plugin} which could be easily integrated with existing optimizer such as AdamW for low-precision training and make it \textit{precision-aware} by replacing critical floating-points with MCF. This avoids the path of high-precision master-weights and upcasting of variables, achieving memory efficiency (\figurename\,\ref{fig:main_algo_descr}\,right).
    \item By proposing the metric effective descent quality, we measure loss in the information at model update step during the training process and provide \textbf{better understanding} of the impact of precisions and \textbf{interpretation} for comparing precision strategies.
    \item \strname offers wall-clock time speedups by storing all variables in low-precision without upcasting. For GPT-$6.7$B and OpenLLaMA-$7$B, \strname using bfloat16 has {\bf up to }\bm{ $3.7\times$} speedup in the training throughput in comparison with mixed-precision strategy with FP$32$ master weights while following a similar training trajectory. The peak memory savings for GPTs ($125$M - $6.7$B) is on average of \bm{$22.8\%/14.9\%$} for \strname formations (light/plus), respectively.
    \item \strname \textbf{trains accurate} models using only low-precision storage compared with FP$32$ master-weights counterpart. For RoBERTa-base, the average GLUE accuracy scores differ by \bm{$+0.85\%$} among the best baseline in Table\,\ref{tab:bert-eval}. Similarly, for GPT of sizes $125$M, $1.3$B, $2.7$B, $6.7$B, \strname has \textbf{similar validation perplexity} as FP$32$ master weights in Table\,\ref{tab:gpt-all-ppl}.
    
\end{itemize}

\section{Background}
\label{sec:background}

\begin{figure*}
    \centering
    \includegraphics[width=0.8\linewidth]{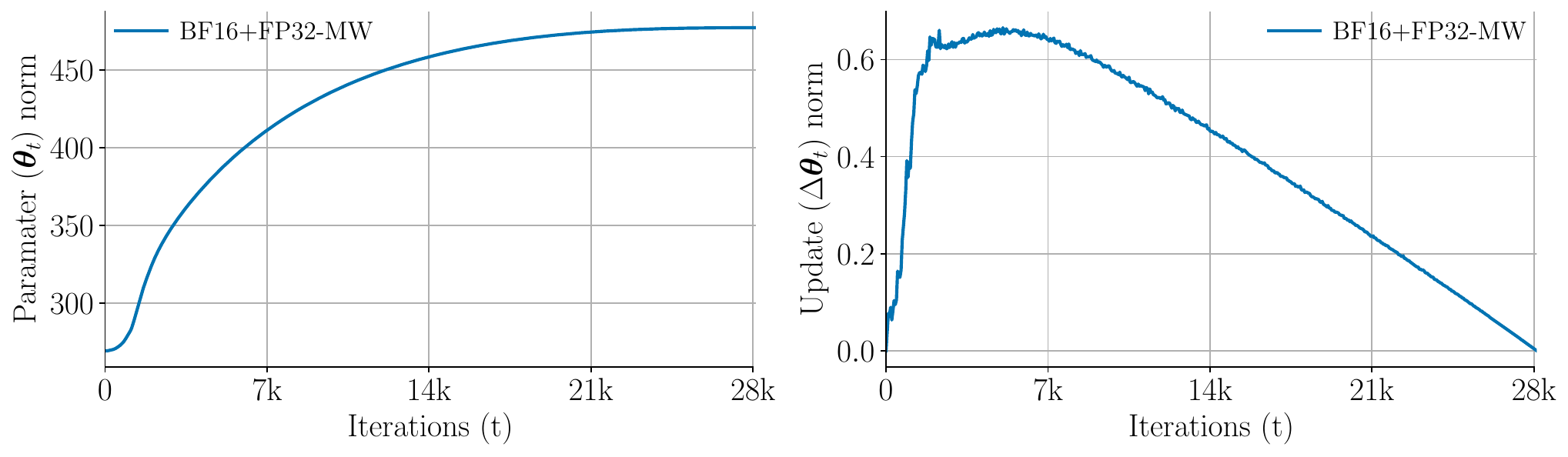}
    \caption{Bert-base-uncased phase-1 pretraining with settings as described in Section\,\ref{subsec:pretrain-bert-roberta}. \textbf{Left:} Model parameter L2 norm vs iterations for BF16 and FP32 master weights strategy. \textbf{Right:} update $\Delta\bm{\theta}_t$ L2 norm across iterations. The model parameter norm and update norm are at different scales, for example, $\sim 450$ vs $\sim 0.5$ at $14$k iterations, which is a factor of $900$ and causes lost arithmetic.}
    \label{fig:bert-base-param-norm}
\end{figure*}

We provide a survey on using different floating-points precision strategies for training LLM. We also introduce necessary background information on floating-point representations using a new structure, multi-component float. 
\subsection{Floats in LLM Training}
\label{ssec:floats_llm_backgr}
In LLM training, weights, activation, gradients are usually stored in low precision floating-points such as $16$-bit BF$16$ \cite{micikevicius2017mixed} for enhanced efficiency and optimized memory utilization.
The low-bits floating point units (FPUs) are appealing because of its low memory foot-print and computational efficiency. Due to numerical inaccuracies, popular choices of training strategies using FPUs are as follows.

\textbf{Mixed-precision} refers to operations executed in low precision ($16$-bit) with minimal interactions with high precision ($32$-bits) floats, thus offering wall-clock speedups. For example, in GEMM (Generalized Matrix Multiplication), matrix multiplication is performed in 16-bit while accumulation in done in 32-bit through tensor cores in NVIDIA A100 \cite{jia2021A100} and V100 \cite{jia2018dissecting}. 

\paragraph{Mixed-precision with Master Weights.} Mixed-precision computations of the activations and gradients are not sufficient to ensure a stable training due to encountered numerical inaccuracies, especially, when gradients and model parameters are at different scale, which is the case with large models (see \figurename\,\ref{fig:bert-base-param-norm}
). A standard workaround is to use the master weight (MW), which refers to maintaining an additional high-precision version (such as $32$-bit float) copy of the model (\figurename\,\ref{fig:main_algo_descr} left) and then performing model update (optimizer step) in high-precision to the master weight \cite{micikevicius2018mixed}. To our knowledge, this approach has the state-of-the-art performance among mixed-precision strategies.

Note that, we also use mixed-precision for GEMM (activations and gradients) in our work. In addition to ``standard single float" representation which is used in the above strategies, an alternate form is discussed below in Section\,\ref{ssec:mcf_backgr}.

\subsection{Multiple-Component Floating-point}
\label{ssec:mcf_backgr}

Precise computations can be achieved with one of two approaches in numerical computing.
\begin{enumerate}[noitemsep,topsep=-2pt]
    \item[(i)] \textit{multiple-bit}, i.e., using ``standard single float" with more bits in the mantissa/fraction, such as 32-, 64-bit floats, and even Bigfloat \cite{granlund2004gnu};
    \item[(ii)] \textit{multiple-component} representation using unevaluated sum $x_1+x_2+\cdots+x_n$ of multiple floats usually in low-precision such as BF$16$, FP$16$, or even FP$8$.
\end{enumerate}

Multiple-bit approach has an advantage of large representation range, while the multiple-component floating-point (MCF) has an advantage in speed, as it consists of only low-precision floating-point computations. Additionally, rounding is often required in $p$-bit ``standard single float" arithmetic due to output requiring additional bits to express and store exactly, while in MCF, the rounding error could be circumvent and accounted via appending additional components. A basic structure in MCF is expansion:

\begin{definition}
\label{def:mcf}
\cite{priest1991Arithmetic}. A length-$n$ expansion ($x_1, \ldots, x_n$) represents the unevaluated exact sum $x = x_1 + x_2 + \cdots + x_n$, where components $x_i$ are non-overlapping $p$-bit floating-points in decreasing order, i.e., for $i<j$, the least significant non-zero bit of $x_i$ is more significant than the most significant non-zero bit of $x_j$ or vice versa.
\end{definition}

Exact representations of real numbers such as $0.999$ is usually muddled in low-precision, such as BF16, with rounding-to-the-nearest (RN); $0.999\xrightarrow[\text{BF16}]{\text{RN}}1.0$, but can be represented accurately as a length-$2$ expansion $(1.0, -0.001)$ in MCF with two BF$16$ components. The first component serves as an approximation to the value, while the second accounts for the roundoff error. This problem is further aggravated in weighted averaging (see Section\,\ref{ssec:collage_mcf_algo}), such that instead of the average, a monotonic increasing sum is produced causing reduced step size and poor learning. We aim to alleviate such problems by using expansions to represent numbers and parameters accurately (e.g., Table\,\ref{tab:beta2_mcf}). Since speed and scalability is critical for LLM training, we are particularly interested in utilizing low-precision MCF (e.g., BF$16$ and FP$16$) as low-bit FPUs are faster than their high-bit counterparts such as FP$32$. For rest of the work, we consider only length-$2$ expansion for MCF as it suffices for our purpose.
\section{Imprecision Issues}
\label{sec:imprecision_issue}

To motivate the work, in this section, we formalize the issue of imprecision in floating point units. Afterwards, we introduce a novel metric to monitor the information loss. Next, we show its impact via a case study on BERT-like models \cite{devlin2019bert, liu2019roberta}.  Unless specified otherwise, the low-precision FPU is referred to bfloat$16$, and the same analogy can be easily extended for other low-precision FPUs such as float$16$, float$8$.

\subsection{Imprecision with Bfloat16}
\label{ssec:imprecision}
A commonly encountered problem of computations using low-precision arithmetic is \textit{imprecision}, where an exact representation of a real-number $x$ either requires more mantissa bits (see Appendix\,\ref{appsec:FPU} for definitions) beyond the limit (for example, $7$ bits in bfloat$16$), or is not possible (for example, $x=0.1$, is rounded to $0.1001$ in BF16). As a result, the given number $x$ will be rounded to a representable floating-point value, causing numerical quantization errors. An important concept for FPU rounding is unit in the last place ($\ulp$), which is the spacing between two consecutive representable floating-point numbers, i.e., the value the least significant (rightmost) bit represents if it is $1$.

\begin{definition}[$\ulp$ \cite{muller2018handbook}]
\label{def:ulp}
    In radix $2$ with precision $P$, if $2^e\leq |x|< 2^{e+1}$ for some integer $e$, then $\ulp(x)=2^{\max{(e, e_{\min})} - P}$, where $e_{\min}$ is the zero offset in the IEEE 754 standard. 
\end{definition}
Broadly speaking, two numbers for a given FPU are separated by its $\ulp$, hence the worst case rounding error for any given $x$ is $\ulp(x)/2$ \cite{goldberg1991FPU} assumed rounding-to-the-nearest is used. 
Next, lets denote 
$\mathcal{F}^{\text{BF16}}(a\,\varkappa\,b)$ 
as bfloat16 floating-point operation between $a,b$, where $\varkappa$ could be $\oplus$ addition, $\odot$ multiplication, etc. Such operations can be computationally inaccurate and as a consequence, we identify below a problematic behavior with RN. 
\begin{definition}[Lost Arithmetic]
\label{def:lost-arithmetic} 
Given the input floating-point numbers $a, b$ and precision $P$. A floating operation $\mathcal{F}^P(a\,\varkappa\,b)$ is lost if 
\[
|\mathcal{F}^P(a\,\varkappa\,b) - a| \leq \frac{\ulp(a)}{2},\,\text{or}\,|\mathcal{F}^P(a\,\varkappa\,b) - b| \leq \frac{\ulp(b)}{2}.
\]
Consequently, $\mathcal{F}^P(a\,\varkappa\,b)=a,\,\text{or}\,b$, respectively.
\end{definition}

\textbf{Remark:} For any non-zero bfloat16 number, if $|b|\leq \text{ulp}(a)/2$, then $\mathcal{F}^{\text{BF16}}(a\,\oplus\,b) = a$. As an example, if $a=200, b=0.1$, then $\mathcal{F}^{\text{BF16}}(200\,\oplus\,0.1) = 200$, since $\text{ulp}(200) = 1$. Next, we discuss these concepts in the context of LLM training.

\subsection{Loss of Information in LLM Training}
\label{ssec:inf_loss}
The situation of `adding two numbers at different scale' is very common in LLM training. See \figurename\,\ref{fig:bert-base-param-norm} 
, where due to different scales of model parameter and updates, $\oplus$ in bfloat16 becomes an \textit{lost arithmetic}. A pseudocode of model parameter ($\vec{\theta}$) update using bfloat16 at iteration $t$ is written as
\begin{equation}
    \vec{\theta}_{t} \leftarrow \mathcal{F}^{\text{BF16}}(\vec{\theta}_{t-1} \oplus \Delta\vec{\theta}_{t}),
    \label{eqn:model_update_FPU}
\end{equation}
where, $\Delta\vec{\theta}_{t}$ is the aggregated update from an optimizer (for example, including learning rate, momentum, etc.) at step $t$. With a possibility of lost arithmetic in Equation~\eqref{eqn:model_update_FPU}, the actual updated parameter could be different from expected. Hence, we define the effective update at step $t$ as
\begin{equation}
    \widehat{\Delta\vec{\theta}}_t = \mathcal{F}^{\text{BF16}}(\vec{\theta}_{t-1} \oplus \Delta\vec{\theta}_{t}) - \vec{\theta}_{t-1}.
    \label{eqn:effective_update_float}
\end{equation}
Note that in the event of no lost arithmetic, $\widehat{\Delta\vec{\theta}}_t = \Delta\vec{\theta}_{t}$. While, when $\widehat{\Delta\vec{\theta}}_t \neq \Delta\vec{\theta}_{t}$ which is usually the case with low-precision FPUs, there is a loss in information as $\leq\ulp/2$ values are simply ignored (see \figurename\,\ref{fig:imprecison-percentage-bert-base-uncased}). To better capture this information loss, we introduce a novel metric.

\begin{definition}[\textbf{Effective Descent Quality}] Given the current parameter, aggregated update at step $t$ as $\vec{\theta}_t$, $\Delta\vec{\theta}_t$, respectively. The effective descent quality for a given floating-pint precision is defined as 
\begin{equation}
    \edq(\Delta\vec{\theta}_t, \widehat{\Delta\vec{\theta}}_t;\vec{\theta}_t, P) 
    = \Big\langle  \frac{\Delta\vec{\theta}_t}{\vert\vert\Delta\vec{\theta}_t\vert\vert}, \widehat{\Delta\vec{\theta}}_t\Big\rangle,
    \label{eqn:edq_defn}
\end{equation}
where, $\widehat{\Delta\vec{\theta}}_t$ is defined in eq.~\eqref{eqn:effective_update_float} for a given precision $P$.
\end{definition}

In other words, $\edq$ in eq. \eqref{eqn:edq_defn} is projection of the effective update along the desired update. In the absence of any imprecision, $\edq$ will be simply the norm of original update. We show in Section~\ref{subsec:pretrain-bert-roberta} and Figure~\ref{fig:metrics-bert-base-uncased} how $\edq$ relates to the learning and helps understanding impacts of different precision strategies. 

To remedy the imprecision and lost arithmetic in the model parameter update step (Equation~\eqref{eqn:model_update_FPU}), works such as Kahan summation \cite{zamirai2020revisiting,park2018training} exist (see Appendix~\ref{app-para:kahan}), however, we see in \figurename~\ref{fig:metrics-bert-base-uncased} (Middle) that although Kahan-based BF$16$ approach improves over `BF$16$' training but it still could not match with the commonly used FP$32$ master weights approach.
\section{\strname: Low-Precision MCF Optimizer}
\label{sec:methodology}

In this section, we present \strname, a low precision strategy \& optimizer implementation to solve aforementioned imprecision and lost arithmetic issues in Section\,\ref{sec:imprecision_issue} without upcasting to a higher precision, using the multiple-component floating-point (MCF) structure.

\subsection{Computing with MCF}
\label{ssec:mcf_collage_intro}
Precise computing with exact numbers stored as MCF expansions is easy with some basic algorithms\footnote{The correctness of algorithms presented herein rely on the assumption that standard rounding-to-the-nearest is used.}. For example, \pythoninline{Fast2Sum} captures the roundoff error for the float addition $\oplus$ and outputs an expansion of length $2$.

\begin{theorem}[Fast2Sum \cite{dekker1971float}]
\label{thm:fast2sum}
Let two floating-point numbers $a,b$ be $|a|\geq |b|$, \pythoninline{Fast2Sum} produces a MCF expansion $(x,y)$ such that $a+b=x+y$, where $x\gets\mathcal{F}^{P}(a\oplus b)$ is the floating-point sum with precision $P$, $y\gets\mathcal{F}^{P}\left(b\ominus\mathcal{F}^{P}(x\ominus a)\right)=a+b-\mathcal{F}^{P}(a\oplus b)$ is the rounding error. Also, $y$ is upper-bounded such that $|y|<\ulp(x)/2$.
\end{theorem}

\setcounter{algorithm}{1}
\begin{algorithm*}[ht]
\caption{\strname: Bfloat$16$ \colorbox{darkpink!15}{MCF} Adam{W} Optimization}
\begin{algorithmic}[1]
    \label{alg:mcfadamw}
    \STATE Given $\alpha$ (learning rate), $\beta_1$, $\beta_2$, $\epsilon$, $\lambda\in\mathbb{R}$
    \STATE Initialize time step: $t \leftarrow 0$, BF$16$ parameter vector $\vec{\theta}_{t=0}\in\mathbb{R}^n$, BF$16$ first moment vector: $\vec{m}_{t=0} \leftarrow \vec{0}$, BF$16$ second moment vector: $\vec{v}_{t=0} \leftarrow \vec{0}$
    \STATE \colorbox{darkpink!15}{Initialize 2nd component $\vec{\delta\theta}_{t=0}\leftarrow \vec{0}$ in BF$16$ for parameter}
    \STATE \colorbox{darkpink!15}{(optional) Represent $\beta_2$ as expansion $(\hat{\beta}_2, \delta\beta_2)$, initialize 2nd component $\vec{\delta v}_{t=0}\!\leftarrow\!\vec{0}$ in BF$16$ for second moment}
    \REPEAT
        \STATE $t \leftarrow t + 1$
        \STATE $\vec{g}_t \leftarrow \nabla f_t(\vec{\theta}_{t-1})$
        \STATE $\vec{m}_t \leftarrow \beta_1 \cdot \vec{m}_{t-1} + (1 - \beta_1) \cdot \vec{g}_t$
        \STATE $\vec{v}_t \leftarrow \beta_2 \cdot \vec{v}_{t-1} + (1 - \beta_2) \cdot \vec{g}_t^2$ ~~~~\colorbox{darkpink!15}{$\Longrightarrow~~~~(\vec{v}_t, \vec{\delta v}_t) \leftarrow \textbf{Grow}(\textbf{Mul}(\hat{\beta}_2,\delta\beta_2), (\vec{v}_{t-1}, \vec{\delta v}_{t-1})), (1 - \beta_2) \cdot \vec{g}_t^2)$}
        \STATE $\hat{\vec{m}}_t \leftarrow \vec{m}_t / (1 - \beta_1^t)$
        \STATE $\hat{\vec{v}}_t \leftarrow \vec{v}_t / (1 - \beta_2^t)$
        \STATE $\vec{\Delta\theta}_t \leftarrow -\alpha(\hat{\vec{m}}_t/(\sqrt{\hat{\vec{v}}_t + \epsilon}) + \lambda\vec{\theta}_{t-1})$
        \STATE $\vec{\theta}_t \leftarrow \vec{\theta}_{t-1}+ \vec{\Delta\theta}_t$ ~~~~\colorbox{darkpink!15}{$\Longrightarrow~~~~(\vec{\theta}_t, \vec{\delta\theta}_t) \leftarrow \textbf{Grow}((\vec{\theta}_{t-1}, \vec{\delta\theta}_{t-1}), \vec{\Delta\theta}_t)$}
    \UNTIL stopping criterion is met
    \STATE {\bfseries return:} optimized parameters $\vec{\theta}_t$
\end{algorithmic}
\end{algorithm*}

Note that, particularly for LLM training, we are able to add using \pythoninline{Fast2Sum} without any sorting since parameter weights $\vec{\theta}$ are usually larger than the gradients and updates $\vec{\Delta\theta}$ in absolute value at the parameter update step Equation~\eqref{eqn:model_update_FPU} (See \figurename\,\ref{fig:bert-base-param-norm}
).
Similar basic algorithms exist for the multiplication of two floats, which produces in the same way a length-2 expansion. Using the basic algorithms, an exhaustive set of advanced algorithms are developed \cite{yu2022mctensor}. We refer the reader to Appendix\,\ref{appsec:mcf_algs} for more details. Particularly, for the optimizer update step \eqref{eqn:model_update_FPU}, a useful algorithm to introduce is \pythoninline{Grow} (see Algorithm~\ref{alg:grow}) which adds a float to a MCF expansion of length $2$.

\setcounter{algorithm}{0}
\begin{algorithm}[H]
   \caption{\textbf{Grow}}
   \label{alg:grow}
\begin{algorithmic}[1]
    \STATE {\bfseries Input:} an expansion $(x,y)$ and a float $a$ with $|x|\geq |a|$
    \STATE $(u, v) \gets \textbf{Fast2Sum}(x, a)$
    \STATE $(u, v) \gets \textbf{Fast2Sum}(u, y + v)$
    \STATE {\bfseries Return:} $(u, v)$
\end{algorithmic}
\end{algorithm}

\subsection{\strname: Bfloat$16$ MCF AdamW}
\label{ssec:collage_mcf_algo}
Using the basic components from Section\,\ref{ssec:mcf_collage_intro} and Appendix\,\ref{appsec:mcf_algs}, we now provide plugins to modify a given optimizer such as AdamW \cite{loshchilov2017decoupled} to be \textit{precision-aware} and \textbf{store entirely with low-precision} floats, specifically bfloat$16$ in Algorithm\,\ref{alg:mcfadamw}. Note that, mixed-precision is still used in GEMM for obtaining gradients and activations but are stored in bfloat$16$ only. The required changes are highlighted in pink, and are discussed individually as follows.

\paragraph{Model Parameters} We substitute the bfloat$16$ model parameter $\vec{\theta}_t$ with a length-$2$ MCF expansion $(\vec{\theta}_t, \vec{\delta\theta}_{t})$ by appending an additional bfloat$16$ variable $\vec{\delta\theta}_{t}$ in line-$3$ which does not require any gradients. Next, to update the model parameter expansion, we use \pythoninline{Grow} in line-13 to add a float $\Delta\vec{\theta}_t$ to the expansion.

\begin{wraptable}{r}{4cm}
\vspace{-1em}
\captionsetup{font=small}
\caption{length-$2$ expansions for $\beta_2$ in Bfloat$16$.}
\label{tab:beta2_mcf}
\resizebox{0.8\linewidth}{!}{
\begin{tabular}{@{}c|c@{}}
$\beta_2$ & BF$16$ MCF \\\hline
$0.999$ & $(1, -0.001)$ \\ 
$0.99$ & $(0.9893, 0.0017)$ \\  
$0.95$ & $(0.9492, 0.0008)$ \\  
\end{tabular}
}
\vspace{-1em}
\end{wraptable}
\paragraph{Optimizer States}\label{para:mcf_optim_states} With Adam-like algorithms, unlike the first moment $\vec{m}_t$, the second moment $\vec{v}_t$ update suffers from severe imprecision and lost arithmetic due to smaller accumulation, $\vec{g}_t$ vs $\vec{g}_t^2$. To make the matter worse, default choice of $\beta_2$ such as $0.999$ \cite{devlin2019bert} are simply rounded to $1.0$ in bfloat$16$, thus resulting in a monotonic increase in second momentum. This in turn makes the update $\Delta\vec{\theta}_t$ smaller and hence slower learning as we see in \figurename\,\ref{fig:metrics-bert-base-uncased}. To alleviate this issue, we propose switching $\beta_2$ from standard single float to a MCF expansion as $(\beta_2, \delta\beta_2)$, and also for second momentum as $(\vec{v}_{t}, \vec{\delta v}_{t})$. Doing so, we have an exact representation of $\beta_2$ as shown in Table\,\ref{tab:beta2_mcf}. We then perform a multiplication of two expansions using \pythoninline{Mul} (see Appendix\,\ref{appsec:mcf_algs}).

\begin{table*}
 \centering
 \caption{Precision breakdown of various training strategies applied to the given optimizer. The strategies are ranked from top to bottom in the order of byte/parameter occupancy.}
    \resizebox{0.7\linewidth}{!}{
    \begin{tabular}{lcccc} 
    \toprule
    \multirow{3}{*}{Precision Option} & \multicolumn{3}{c}{Stages \& Components} & \multirow{3}{*}{\makecell{Memory \\ (bytes/parameter)}} \\
    \cmidrule{2-4}
    & \makecell{Parameter \\ \&  Gradient} & \makecell{Optimizer \\ States} & \makecell{MCF or \\ Master Weight} &  \\
    \midrule
    A (BF$16$) & BF$16\times2$ & BF$16\times2$ & NA & $8$ \\
    B (\strname-light)\,(ours) & BF$16\times2$ & BF$16\times2$ & BF$16\times1$ & $10$ \\
    C (\strname-plus)\,(ours) & BF$16\times2$ & BF$16\times2$ & BF$16\times2$ & $12$ \\
    D (BF$16$ + FP$32$Optim + FP$32$MW) & BF$16\times2$ & FP$32\times2$ & FP$32\times1$ & $16$ \\
    \bottomrule
    \end{tabular}
    }
    \label{tab:precision-strategies-breakdown}
\end{table*}

For the sake of simplicity in notations, we denote \strname-light as using MCF expansions only for model parameters and \strname-plus for both model parameters and optimizer states. It's worthy to note that imprecision and lost arithmetic are common and sometimes hard to notice. We only identify places when they hurt training accuracies. A rule of thumb is to do as many scalar computations in high precision as possible before casting them to low precision (e.g., PyTorch BFloat$16$ Tensor). Worthy to note, existing Kahan-based optimizers are special cases of \strname-light under a magnitude assumption, we defer this discussion and other places of imprecision and lost arithmetic such as weight decay that exist in the algorithm to Appendix~\ref{appsec:further_discussions_algorithm}.
\begin{table*}
\vspace{-1.5em}
 \centering
    \caption{\centering Pre-training perplexity of BERT (both phases) and RoBERTa for all precision strategies as listed in Table\,\ref{tab:precision-strategies-breakdown}. Lower values are better, with the best results in bold. D$^{-\text{MW}}$ with FP$32$Optim with same bytes/parameter as \strname could not match its performance.}
    \resizebox{0.7\linewidth}{!}{
    \begin{tabular}{lccccc} 
    \toprule
     \multirow{3}{*}{Precision Option} & \multicolumn{4}{c}{$\beta_2=0.999$} & $\beta_2=0.98$\\
     & \multicolumn{2}{c}{BERT-base} & \multicolumn{2}{c}{BERT-large} & \multirow{2}{*}{RoBERTa-base} \\
    & Phase-$1$ & Phase-$2$ & Phase-$1$ & Phase-$2$ &  \\
    \hline
    A  & $8.67$ & $7.61$ & $6.05$ & $5.47$ & $3.82$ \\
    \hline
    B (\strname-light) & $5.99$ & $5.26$ & $4.39$ & $3.90$ & $3.49$\\
    C (\strname-plus) & $\bf{5.26}$ & $\bf{4.66}$ & $\bf{3.94}$ & $\bf{3.53}$ & $3.49$\\
    \hline
    D$^{-\text{MW}}$ (BF$16$ + FP$32$Optim) & $6.23$ & $5.64$ & $4.66$ & $4.22$ & $3.82$\\
    D & $\bf{5.26}$ & ${4.71}$ & $4.06$ & $3.63$ & $\bf{3.46}$ \\
    \bottomrule
    \end{tabular}
    }
    \vspace{-0.5em}
    \label{tab:bert-roberta-pretraining-ppl}
\end{table*}
\section{Empirical Evaluation}
\label{sec:experiments}
We evaluate \strname formations against the existing precision strategies on pretraining LLMs at different scales, including BERT~\cite{devlin2019bert}, RoBERTa~\cite{liu2019roberta}, GPT~\cite{gpt-neox-library}, and OpenLLaMA~\cite{touvron2023llama}. Specifically, we compared the following precision strategies in our experiments, which are ordered in an increasing number of byte/parameter (see Table\,\ref{tab:precision-strategies-breakdown}).
\begin{itemize}[nosep, leftmargin=*]
    \item Option A: Bfloat$16$ parameters
    \item Option B: Bfloat$16$ + \strname-light
    \item Option C: Bfloat$16$ + \strname-plus
    \item Option D: Bfloat$16$ + FP$32$ Optimizer states + FP$32$ master weights
\end{itemize}

Since option D is the best-known baseline with state-of-the-art quality among mixed-precision strategies, we aim to outperform, or at least match the quality of option D with \strname throughout our experiments. We show that \strname matching the quality of option D, has \textbf{orders-magnitude higher} performance (speed, see Table~\ref{tab:speedup}). All strategies are evaluated using AdamW \cite{loshchilov2017decoupled} optimizer with standard $\beta_1=0.9$ while varying $\beta_2$ as per different experiments. We use \textit{aws.p4.24xlarge} compute instances for all of our experiments.

\subsection{Pre-training BERT \& RoBERTa}
\label{subsec:pretrain-bert-roberta}
We demonstrate that BF$16$-\strname can be used to obtain an accurate model, comparable to heavy-weighted FP$32$ master weights strategy. 

\paragraph{Precision options.} 
In addition to options A, B, C, D, we further augment our experiments with another baseline strategy D$^{-\text{MW}}$, where we disabled the FP$32$ master weights but only used FP$32$ optimizer states. This strategy saves $4$ bytes/parameter in comparison to Option D and has the same bytes/parameter as option C (\strname-plus).

\paragraph{Model and Dataset.} We first pre-train the BERT-base-uncased, BERT-large-uncased, and RoBERTa-base model with HuggingFace (HF) \cite{wolf2019huggingface} configuration on the Wikipedia-en corpus \cite{Wikiextractor2015}, preprocessed with BERT Wordpiece tokenizer. We execute the following pipeline to pretrain, i) BERT in two phases with phase-1 on $128$ sequence length, and then phase-2 with $512$ sequence length; and ii) RoBERTa with sequence length $512$. We adopt $\beta_2=0.999$ for BERT and $\beta_2=0.98$ for RoBERTa following the configs from HF. We defer more training details to Appendix\,\ref{appssec:bert_roberta}.

\begin{figure*}
\centering
\subfigure{\label{fig:imprecison-percentage-bert-base-uncased}\includegraphics[width=56mm]{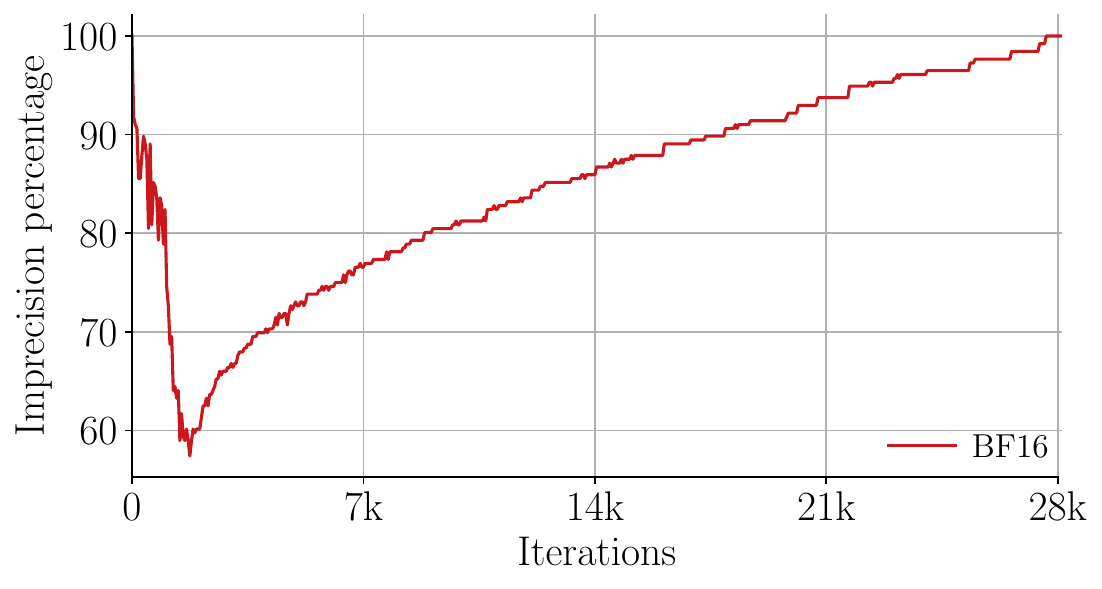}}
\subfigure{\label{fig:loss-bert-base-uncased}\includegraphics[width=56mm]{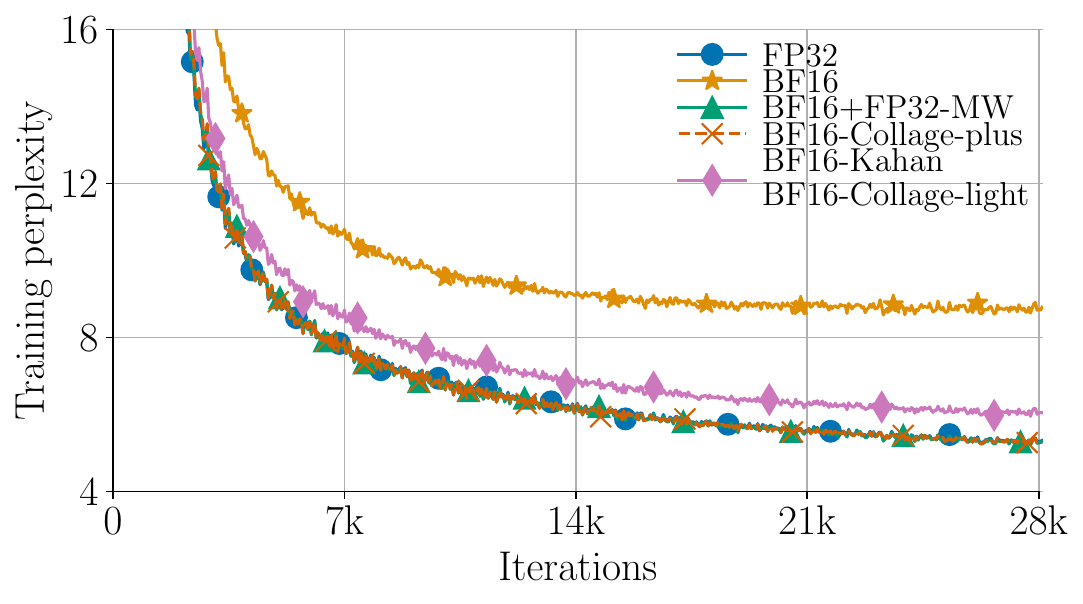}}
\subfigure{\label{fig:dqa-bert-base-uncased}\includegraphics[width=56mm]{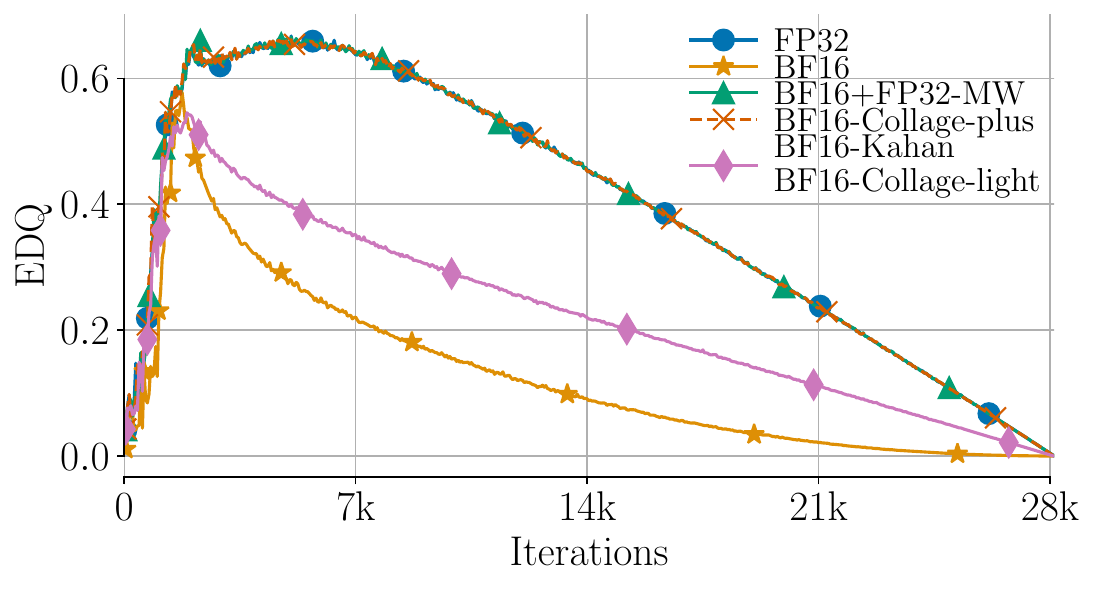}}
\vspace{-1em}
\caption{BERT phase-$1$ pre-training (see Appendix\,\ref{appssec:bert_roberta} for details). \textbf{Left:} 
Imprecision percentage
($\%$) measured as the percentage of lost arithmetic for all model parameters, i.e., not updated, vs iterations for BF$16$. \textbf{Middle:} Training perplexity vs iterations for various precision strategies (see Table\,\ref{tab:precision-strategies-breakdown}). Additionally, we evaluate ``FP32" as 32-bit counterpart of option A, and BF16-Kahan as Kahan-sum \cite{zamirai2020revisiting} with BF16 parameters. \textbf{Right:} Effective descent quality ($\edq$) in \eqref{eqn:edq_defn} vs iterations to measure loss in information at the optimizer step for different precision strategies. BF16-\strname-plus training perplexity and $\edq$ \textbf{overlaps} with the best ``FP32", and ``BF16 + FP32 MW" with less bytes/parameter.
}
\label{fig:metrics-bert-base-uncased}
\end{figure*}

\begin{table*}[h]
\vspace{-1.5em}
 \centering
     \caption{GLUE benchmark for BERT-base-uncased and RoBERTa-base pre-trained using different precision strategies. See Appendix\,\ref{appssec:bert_roberta} for experimental details. BF$16$-\strname training strategy matches/exceeds the finetuning quality over several metrics.
     }
    \resizebox{0.9\linewidth}{!}{
    \begin{tabular}{llccccccccc} 
    \toprule
     Model & Precision & MRPC & QNLI & SST-2 & CoLA & RTE & STS-B & QQP & MNLI & Avg\\
    \midrule
    \multirow{4}{*}{BERT-base} & A & $0.8210$ & $0.8832$ & $0.8890$ & $0.3522$ & $0.6462$  & $0.8666/0.8618$ & $0.8973$ & $0.7993$ & $0.7796$  \\
    & B (ours)& $0.8431$ & $0.8974$ & $0.9071$ & $0.4149$ & $0.6606$ & $0.8837/0.8785$ & $0.9031$ & $0.8184$ & $0.8007$\\
    & C (ours) & $0.8602$ & $\bf{0.9090}$ & $\bf{0.9128}$ & $\bf{0.4314}$ & $0.6698$ & $0.8851/0.8821$ & $\bf{0.9069}$ & $\bf{0.8330}$ & $\bf{0.8100}$\\
    & D & $\bf{0.8651}$ & $0.9071$ & $0.9036$ & ${0.4212}$ & $\bf{0.6714}$ & $\bf{0.8890/0.8849}$ & ${0.9064}$ & $\bf{0.8330}$ & ${0.8090}$\\
    \hline
    \multirow{4}{*}{RoBERTa-base} & A & ${0.8504}$ & $0.8914$ & $0.9000$ & $0.3866$ & ${0.6281}$ & $0.8636/0.8625$ & $0.8981$ & $0.8155$ & $0.7884$\\
    & B (ours) & $0.8455$ & ${0.9000}$ & ${0.9025}$ & ${0.4460}$ & ${0.6281}$ & ${0.8636/0.8635}$ & ${0.9002}$ & $0.8182$ & ${0.7964}$\\
    & C (ours)& $\bf{0.8529}$ & $\bf{0.9040}$ &$\bf{0.9048}$ & $\bf{0.4588}$ & $0.6137$ & $\bf{0.8658/0.8647}$ & $\bf{0.9005}$ & $\bf{0.8230}$ & $\bf{0.7986}$\\
    & D & $0.8406$ & $0.8993$ & $0.9002$ & $0.3870$ & $\bf{0.6389}$ & $0.8622/0.8631$ & $0.8999$ & ${0.8203}$ & $0.7901$\\
    \bottomrule
    \end{tabular}
    }
    \vspace{-0.5em}
    \label{tab:bert-eval}
\end{table*}

\paragraph{Results.} 
The final pretraining perplexity of various precision strategies are summarized in Table\,\ref{tab:bert-roberta-pretraining-ppl} and for BERT-base, the complete phase-$1$ training loss trajectory is shown in \figurename\,\ref{fig:metrics-bert-base-uncased}~middle. Additionally, we did finetuning of the pre-trained models on the GLUE benchmark \cite{wang2019glue} for eight tasks in Table~\ref{tab:bert-eval} with the same configurations specified in Appendix\,\ref{appssec:bert_roberta}. \strname-plus although using only BF16 parameters, outperforms the vanilla BF16 option A and matches/exceeds option D for both pre-training and finetuning experiments. For BERT-base \strname-plus exceeds on \textbf{5/8} tasks with \bm{$+0.1\%$} lead in average, while for roberta-base its exceeds on \textbf{7/8} tasks with \bm{$+0.85\%$} in average. Note that, although D$^{-\text{MW}}$ has FP$32$ optimizer states and same/more byte/parameter complexity as \strname-plus/light, respectively, it could not match the quality \textbf{showing the importance of MCF} in the AdamW through \strname. This shows that simply having higher-precision is not enough to obtain better models but requires a \textit{careful consideration of the floating errors}.

Interestingly, \strname-light suffices to closely match the option D in the RoBERTa pretraining experiments with $\beta_2=0.98$, while lagging to match with the $\beta_2=0.999$ BERT pretraining experiments. Our proposed metric, the effective descent quality ($\edq$) provides a nuanced understanding of this phenomenon in Figure~\ref{fig:metrics-bert-base-uncased}~(Right). \strname-light and Kahan-based approach improve $\edq$ upon BF$16$ option A at the parameter update step, yet cannot achieve the optimal $\edq$ due to lost arithmetic at the exponential moving averaging step. In contrast, \strname-plus achieves better $\edq$ by taking it into considerations and thereby outperforms the best-known baseline, Option D.

\subsection{Pretraining multi-size GPTs \& OpenLLaMA 7B}
\label{ssec:llama_gpt_results}
\paragraph{Model and Dataset.} We conduct following pretraining experiments; 1) GPT with different sizes ranging from $125$M, $1.3$B, $2.7$B to $6.7$B, and 2) OpenLLaMA-$7$B using NeMo Megatron \cite{kuchaiev2019nemo} with the provided configs. The GPTs are trained on the Wikipedia corpus \cite{Wikiextractor2015} with GPT$2$ BPE tokenizer, and OpenLLaMA-$7$B on the LLaMA tokenizer, respectively. Additional training and hyerparameter details are described in Appendix\,\ref{appssec:gpt_llama}.

\paragraph{Results.} Using the recommended $\beta_2=0.95$ \cite{gpt-neox-library}, Table~\ref{tab:gpt-all-ppl} summarizes the train \& validation perplexity after pre-training GPT models and OpenLLaMA-$7$B under various options. Our \strname formations are able to \textbf{match} the quality of the \textbf{best-known} baseline, FP$32$ MW option D, most of the time \emph{for all models} with the only exception on the smallest GPT-$125$M, while having the same validation perplexity. 

\begin{table*}[ht]
    \centering
    \caption{\textbf{Left:} Train $|$ Validation perplexity of pre-trained GPT with $\beta_2=0.95$. \textbf{Right:} OpenLLaMA-$7$B with $\beta_2=0.95$ and $0.99$.}
    \hspace{-2em}
    \resizebox{0.9\linewidth}{!}{
    \begin{minipage}[t]{0.77\linewidth}
      \centering
        \begin{tabular}{l|cccc} 
        \toprule
        Model & \multicolumn{4}{c}{GPT} \\
        Precision Option & $125$M & $1.3$B & $2.7$B & $6.7$B \\
        \hline
        A (BF16)  & $14.73~|~15.64$ & \hspace{-0.5em}$10.28~|~12.43$ & $9.97~|~12.18$ & $9.87~|~12.18$ \\
        B (\strname-light) & $14.01~|~15.03$ & $8.50~|~17.70$ & $8.33~|~11.36$ & $8.17~|~11.13$ \\
        C (\strname-plus) & $14.01~|~15.03$ & $8.50~|~17.70$ & $8.33~|~11.36$ & $8.17~|~11.13$ \\
        D (BF$16$ + FP$32$Optim + FP$32$MW) & $13.87~|~15.03$ & $8.50~|~17.70$ & $8.33~|~11.36$ & $8.17~|~11.13$ \\
        \bottomrule
        \end{tabular}
    \end{minipage}%
    \hspace{3em}
    \begin{minipage}[t]{0.2\linewidth}
      \centering
        \begin{tabular}{cc}
            \toprule
            \multicolumn{2}{c}{OpenLLaMA-7B} \\
            $\beta_2=0.95$ & $\beta_2=0.99$ \\
            \hline
            $6.36~|~4.81$ & $15.96~|~12.55$ \\
            $5.99~|~4.53$ & $8.00~|~5.99$ \\
            $5.99~|~4.57$ & ${6.11~|~4.62}$ \\
            $5.99~|~4.57$ & $8.58~|~6.42$ \\
            \bottomrule
        \end{tabular}
    \end{minipage} 
    }
    \vspace{-1em}
    \label{tab:gpt-all-ppl}
\end{table*}

\begin{table*}[htbp]
    \centering
     \caption{\centering Train $|$ Validation perplexity of GPT-$125$M pre-trained with $\beta_2\in\{0.95, 0.99, 0.999\}$ and Global BatchSize $\in\{1024, 2048\}$.
        }
    \label{tab:gpt125M-ppl}
    \resizebox{0.9\linewidth}{!}{
    \begin{tabular}{l|ccc|ccc} 
    \toprule
     \multirow{2}{*}{Precision Option} & \multicolumn{3}{c|}{Global BatchSize~$=1024$} & \multicolumn{3}{c}{Global BatchSize~$=2048$} \\
      & $\beta_2=0.95$ & $\beta_2=0.99$ & $\beta_2=0.999$ & $\beta_2=0.95$ & $\beta_2=0.99$ & $\beta_2=0.999$ \\
    \midrule
    A (BF16)  & $14.73~|~15.64$ & $14.88~|~15.80$ & $17.29~|~18.17$ & $14.73~|~15.18$ & $14.88~|~15.33$ & $17.64~|~15.33$ \\
    B (\strname-light) & $14.01~|~15.03$ & $14.01~|~15.03$ & $14.88~|~15.80$ & $13.87~|~14.44$ & $13.87~|~14.44$ & $14.59~|~15.18$ \\
    C (\strname-plus) & $14.01~|~15.03$ & $14.01~|~15.03$ & $14.15~|~15.18$ & $13.87~|~14.44$ & $13.87~|~14.44$ & $14.01~|~14.59$  \\
    D (BF$16$ + FP$32$Optim + FP$32$MW) & $13.87~|~15.03$ & $14.01~|~15.03$ & $14.01~|~15.03$ & $13.87~|~14.44$ & $13.87~|~14.44$ & $14.01~|~14.59$ \\
    \bottomrule
    \end{tabular}
    }
    \vspace{-1em}
\end{table*}

\paragraph{Ablation: Impact of \bm{$\beta_2$}.} 
We conduct ablation experiments to illustrate the impact of $\beta_2$ on the quality of precision strategies by further pre-training the GPT-$125$M model using $\beta_2=0.99$ and $0.999$, with a global batchsize $1024$, $2048$ and the same micro-batchsize $16$, as summarized in Table~\ref{tab:gpt125M-ppl}. Similar to the BERT and RoBERTa pre-training experiments, \strname-light is able to closely match Option D when $\beta_2=0.95$ or $0.99$ and remain unaffected by changes in the global batchsize. 

However, with $\beta_2=0.999$, \strname-light underperforms option D while \strname-plus is still able to closely match option D. As analyzed in Section\,\ref{para:mcf_optim_states}, low precision (Bfloat$16$) arithmetic fails to represent and compute with $\beta_2=0.999$ due to rounding errors.
In fact, we observed the same phenomenon as pre-training BERT \& RoBERTa in Section~\ref{subsec:pretrain-bert-roberta}, including i) a high imprecision percentage of lost additions with low-precision BF$16$ arithmetic; ii) a reduced $\edq$ for \strname-light and a better $\edq$ for \strname-plus. These together rationalize the utility and significance of our proposed metric $\edq$ and the necessity of \strname-plus for quality models. We defer figures of these metrics for GPTs to Appendix\,\ref{appssec:gpt_pt}.

We also pretrain OpenLLaMA-$7$B with $\beta_2=0.99$ in Table~\ref{tab:gpt-all-ppl}~(right), where both \strname formations outperform option D. In fact, we observe that $\beta_2=0.99$ can easily lead to gradient explosion (see Figure\,\ref{fig:openllama-7B-beta2_0p99}\,right in Appendix\,\ref{appssec:openllama7B_pt}), while \strname-plus provides stable training. The training perplexity trajectories in Figure\,\ref{fig:openllama-7B-beta2_0p95},\ref{fig:openllama-7B-beta2_0p99} (in Appendix\,\ref{appssec:openllama7B_pt}) show that \strname-plus effectively solves the imprecision issue and produces quality models.

\remark{The optimal choice of $\beta_2$ differs case-by-case. To our best knowledge, there is no clear conclusion between $\beta_2$ and the converged performance of the pre-trained models. Showing \strname works with different $\beta_2$'s, enable LLM training to be not limited by such precision issues.}

\subsection{Performance and Memory} 
\label{subsec:efficiency_memory}
\paragraph{Throughput.} We record the mean training throughput of precision strategies for pre-training GPTs and OpenLLaMA-$7$B in a simple setting for fair comparisons: one \textit{aws.p4.24xlarge} node with sequence parallel~\cite{korthikanti2023reducing} turned off\footnote{We observed similar throughputs for precision strategies when sequence parallel is turned on}, and present relative speed-up in Table~\ref{tab:speedup}. Both \strname formations are able to maintain the efficiency of option A. Moreover, the speed factor for \strname increases with an increase in the model size, obtaining up to $\bm{3.74\times}$ for GPT-$6.7$B model.
\begin{table}[ht]
\vspace{-1em}
\centering
\caption{Relative speed-up compared to the option D.}
\label{tab:speedup}
\resizebox{0.8\linewidth}{!}{
\begin{tabular}{l|cccc}
\toprule
Precision & \multicolumn{3}{c}{GPT} & \multirow{2}{*}{\makecell{OpenLlama \\ $7$B}}  \\
Option & $1.3$B & $2.7$B & $6.7$B & \\
\hline
A & 1.78$\times$ & 2.59$\times$ & 3.82$\times$ & 3.15$\times$\\
B (ours) & 1.74$\times$ & 2.57$\times$ & 3.74$\times$ & 3.14$\times$ \\
C (ours) & 1.67$\times$ & 2.48$\times$ & 3.57$\times$ & 3.05$\times$ \\
D & 1$\times$ & 1$\times$ & 1$\times$ & 1$\times$ \\
\bottomrule
\end{tabular}
}
\vspace{-2em}
\end{table}

\begin{figure}[h]
\centering
\includegraphics[width=65mm]{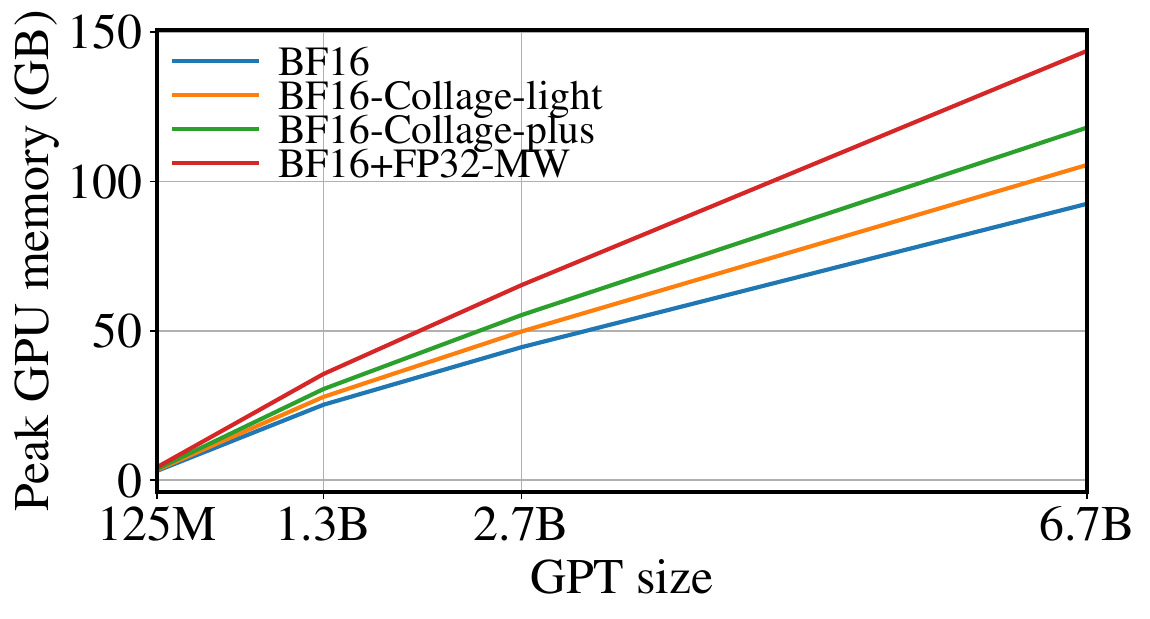}
\vspace{-1em}
\captionof{figure}{GPU peak memory in GB vs model size. GPT-$125$M is hosted on $1$ NVIDIA A$100$ $40$GB, while all other models were hosted on $8\times$ A$100$ $40$GB using tensor-parallelism $8$.}
\label{fig:peak_gpu_mem}
\vspace{-1em}
\end{figure}

\paragraph{Memory.} We probe the peak GPU memory of all training precision strategies during practical runs on $8\times$NVIDIA A$100$s ($40$GB) with the same hyper-parameters for a fair comparison: sequence length $2048$, global batchsize $128$ and micro (per-device) batchsize $1$. \figurename\,\ref{fig:peak_gpu_mem} visualizes the peak memory usage of GPTs vs model sizes.
During real runs, on average, \strname formations (light/plus) use \bm{$23.8\%/15.6\%$} less peak memory compared to option D. The best savings are for the largest model OpenLLaMA-$7$B, with savings \bm{$27.8\%/18.5\%$}, respectively.

\paragraph{Increased Sequence Length and Micro BatchSize.}
\begin{table}
\centering
\caption{Memory compatibility of pre-training GPT-NeoX-$30$B using precision options with different micro batchsize (UBS) and sequence length.}
\label{tab:gpt30B-ubs-seq}
\resizebox{1.0\linewidth}{!}{
\begin{tabular}[width=\linewidth]{@{}l|cc|cc@{}}
    \toprule
    Precision &\multicolumn{2}{c|}{UBS$=1$} & \multicolumn{2}{c}{UBS$=2$}  \\
    option ~~/~~ SeqLen & $1,024$ & $2,048$ & $1,024$ & $2,048$ \\
    \midrule 
    A (BF$16$)  & $\checkmark$ & $\checkmark$ & $\checkmark$ & $\checkmark$ \\
    B (\strname-light)  & $\checkmark$ & $\checkmark$ & $\checkmark$ & $\oom$ \\
    C (\strname-plus)  & $\checkmark$ & $\checkmark$ & $\checkmark$ & $\oom$ \\
    D (BF$16$ + FP$32$Optim + FP$32$MW)  & $\checkmark$ & $\oom$ & $\oom$ & $\oom$ \\
    \bottomrule
    \end{tabular}
    }
\end{table}
We study the benefits of \strname's reduced memory foot-print (as shown in \figurename\,\ref{fig:peak_gpu_mem}), with a demonstration on pre-training a large GPT-$30$B model with tensor-parallelism=$8$, pipeline-parallelism=$2$ on two \textit{aws.p4.24xlarge} ($8\times$A100s $40$GB) instances. Specifically, we identify the maximum sequence length and micro batchsize for all precision strategies to be able to run without $\oom$, as summarized in Table~\ref{tab:gpt30B-ubs-seq}. \strname enables training with an increased sequence length and micro batchsize compared to option D, thus providing a smooth trade-off between quality and performance.

\remark{Further improvements on throughput and memory can be achieved for \strname with specialized fused kernels.}
\section{Conclusion and Future Work}
\label{sec:conclusion}
We provide novel understandings on how low-precision memory and computations affect LLM training, with which, we propose a low-precision plugin \strname using multiple-component floating-point. \strname matches/exceeds the quality of (BF$16$, FP$32$) mixed-precision strategy with master weight while achieving an enhanced execution speed and optimized memory utilization.

\strname can be used in a drop-in manner with any optimization algorithm and float-type. An interesting future work is the direct extension to even lower precision such as $8$-bit FPUs, e.g., FP$8$, removing the usage of FP$16$ in (FP$8$, FP$16$) mixed-precision strategy \cite{peng2023fp8lm}. It's also intriguing to discern when \strname with MCF expansions is more suitable than (BF$16$, FP$32$) mixed-precision strategy with FP$32$ MW, as elucidated in pretraining OpenLLaMA-$7$B.
\section*{Impact Statements}
Our proposed \strname speeds up LLM training with reduced memory usage, without hurting model performances. It can be easily integrated to the existing optimization frameworks. We believe that our method advances the field of LLM and enables efficient-training of even larger and more scalable language models with less carbon foot-print.

\section*{Authors Contributions}
TY conceived the ideas/algorithms, wrote the \strname optimization code, conducted BERT/RoBERTa and GPT training experiments, and wrote the manuscript. 
GG conceived the ideas, conducted evaluation experiments, and wrote the manuscript. 
KG, AM participated in the research discussions and wrote the manuscript. 
HZ helped conduct OpenLLaMA-7B training experiments. 
JH, YP, RD, AP, LH contributed in writing the manuscript.

\nocite{dettmers2023qlora, peng2023fp8lm, perez2023training, yu2022mctensor, zamirai2020revisiting, micikevicius2018mixed}

\nocite{micikevicius2022fp8, desa2018highaccuracy, zhang2022opt}

\nocite{dettmers2023case, dettmers2022llmint8, frantar2023gptq, xiao2023smoothquant}

\nocite{wei2023greener, gupta2015deep, wortsman2023stable, rae2022scaling}
\nocite{hoffmann2022training}
\nocite{priest1992Arithmetic, dekker1971float}
\nocite{goldberg1991FPU}

\nocite{jia2018dissecting, jia2021A100}

\nocite{kahan2006futile}

\nocite{kalamkar2019study, li2017training, hou2018analysis, wang20188bit, xiao2019hybrid8bit, gupta2015deep, wu2018training, sakr2018pertensor}

\clearpage
\bibliography{references}

\begin{thebibliography}{85}
\providecommand{\natexlab}[1]{#1}
\providecommand{\url}[1]{\texttt{#1}}
\expandafter\ifx\csname urlstyle\endcsname\relax
  \providecommand{\doi}[1]{doi: #1}\else
  \providecommand{\doi}{doi: \begingroup \urlstyle{rm}\Url}\fi

\bibitem[Andonian et~al.(2023)Andonian, Anthony, Biderman, Black, Gali, Gao, Hallahan, Levy-Kramer, Leahy, Nestler, Parker, Pieler, Phang, Purohit, Schoelkopf, Stander, Songz, Tigges, Thérien, Wang, and Weinbach]{gpt-neox-library}
Andonian, A., Anthony, Q., Biderman, S., Black, S., Gali, P., Gao, L., Hallahan, E., Levy-Kramer, J., Leahy, C., Nestler, L., et~al.
\newblock {GPT-NeoX: Large Scale Autoregressive Language Modeling in PyTorch}, 9 2023.
\newblock URL \url{https://www.github.com/eleutherai/gpt-neox}.

\bibitem[Attardi(2015)]{Wikiextractor2015}
Attardi, G.
\newblock Wikiextractor.
\newblock \url{https://github.com/attardi/wikiextractor}, 2015.

\bibitem[Banner et~al.(2018)Banner, Hubara, Hoffer, and Soudry]{banner2018scalable}
Banner, R., Hubara, I., Hoffer, E., and Soudry, D.
\newblock Scalable methods for 8-bit training of neural networks, 2018.

\bibitem[Betker et~al.(2023)Betker, Goh, Jing, Brooks, Wang, Li, Ouyang, Zhuang, Lee, Guo, et~al.]{betker2023improving}
Betker, J., Goh, G., Jing, L., Brooks, T., Wang, J., Li, L., Ouyang, L., Zhuang, J., Lee, J., Guo, Y., et~al.
\newblock Improving image generation with better captions.
\newblock \emph{Computer Science. https://cdn. openai. com/papers/dall-e-3. pdf}, 2\penalty0 (3), 2023.

\bibitem[Chen et~al.(2020)Chen, Gai, Yao, Mahoney, and Gonzalez]{chen2020statistical}
Chen, J., Gai, Y., Yao, Z., Mahoney, M.~W., and Gonzalez, J.~E.
\newblock A statistical framework for low-bitwidth training of deep neural networks.
\newblock \emph{Advances in neural information processing systems}, 33:\penalty0 883--894, 2020.

\bibitem[Choi et~al.(2019)Choi, Venkataramani, Srinivasan, Gopalakrishnan, Wang, and Chuang]{MLSYS2019_c443e9d9}
Choi, J., Venkataramani, S., Srinivasan, V.~V., Gopalakrishnan, K., Wang, Z., and Chuang, P.
\newblock Accurate and efficient 2-bit quantized neural networks.
\newblock In Talwalkar, A., Smith, V., and Zaharia, M. (eds.), \emph{Proceedings of Machine Learning and Systems}, volume~1, pp.\  348--359, 2019.
\newblock URL \url{https://proceedings.mlsys.org/paper_files/paper/2019/file/c443e9d9fc984cda1c5cc447fe2c724d-Paper.pdf}.

\bibitem[Courbariaux et~al.(2016)Courbariaux, Hubara, Soudry, El-Yaniv, and Bengio]{courbariaux2016binarized}
Courbariaux, M., Hubara, I., Soudry, D., El-Yaniv, R., and Bengio, Y.
\newblock Binarized neural networks: Training deep neural networks with weights and activations constrained to +1 or -1, 2016.

\bibitem[Croci et~al.(2022)Croci, Fasi, Higham, Mary, and Mikaitis]{croci2022stochastic}
Croci, M., Fasi, M., Higham, N.~J., Mary, T., and Mikaitis, M.
\newblock Stochastic rounding: implementation, error analysis and applications.
\newblock \emph{Royal Society Open Science}, 9\penalty0 (3):\penalty0 211631, 2022.

\bibitem[Dekker(1971)]{dekker1971float}
Dekker, T.~J.
\newblock A floating-point technique for extending the available precision.
\newblock \emph{Numer. Math.}, 18\penalty0 (3):\penalty0 224–242, jun 1971.
\newblock ISSN 0029-599X.
\newblock \doi{10.1007/BF01397083}.
\newblock URL \url{https://doi.org/10.1007/BF01397083}.

\bibitem[Dettmers \& Zettlemoyer(2023)Dettmers and Zettlemoyer]{dettmers2023case}
Dettmers, T. and Zettlemoyer, L.
\newblock The case for 4-bit precision: k-bit inference scaling laws, 2023.

\bibitem[Dettmers et~al.(2022)Dettmers, Lewis, Belkada, and Zettlemoyer]{dettmers2022llmint8}
Dettmers, T., Lewis, M., Belkada, Y., and Zettlemoyer, L.
\newblock Llm.int8(): 8-bit matrix multiplication for transformers at scale, 2022.

\bibitem[Dettmers et~al.(2023)Dettmers, Pagnoni, Holtzman, and Zettlemoyer]{dettmers2023qlora}
Dettmers, T., Pagnoni, A., Holtzman, A., and Zettlemoyer, L.
\newblock Qlora: Efficient finetuning of quantized llms, 2023.

\bibitem[Devlin et~al.(2019)Devlin, Chang, Lee, and Toutanova]{devlin2019bert}
Devlin, J., Chang, M.-W., Lee, K., and Toutanova, K.
\newblock Bert: Pre-training of deep bidirectional transformers for language understanding, 2019.

\bibitem[Dhariwal et~al.(2020)Dhariwal, Jun, Payne, Kim, Radford, and Sutskever]{dhariwal2020jukebox}
Dhariwal, P., Jun, H., Payne, C., Kim, J.~W., Radford, A., and Sutskever, I.
\newblock Jukebox: A generative model for music.
\newblock \emph{arXiv preprint arXiv:2005.00341}, 2020.

\bibitem[Frantar \& Alistarh(2023)Frantar and Alistarh]{frantar2023sparsegpt}
Frantar, E. and Alistarh, D.
\newblock Sparsegpt: Massive language models can be accurately pruned in one-shot, 2023.

\bibitem[Frantar et~al.(2022)Frantar, Ashkboos, Hoefler, and Alistarh]{frantar2022gptq}
Frantar, E., Ashkboos, S., Hoefler, T., and Alistarh, D.
\newblock Gptq: Accurate post-training quantization for generative pre-trained transformers.
\newblock \emph{arXiv preprint arXiv:2210.17323}, 2022.

\bibitem[Frantar et~al.(2023)Frantar, Ashkboos, Hoefler, and Alistarh]{frantar2023gptq}
Frantar, E., Ashkboos, S., Hoefler, T., and Alistarh, D.
\newblock Gptq: Accurate post-training quantization for generative pre-trained transformers, 2023.

\bibitem[Goldberg(1991)]{goldberg1991FPU}
Goldberg, D.
\newblock What every computer scientist should know about floating-point arithmetic.
\newblock \emph{ACM Comput. Surv.}, 23\penalty0 (1):\penalty0 5–48, mar 1991.
\newblock ISSN 0360-0300.
\newblock \doi{10.1145/103162.103163}.
\newblock URL \url{https://doi.org/10.1145/103162.103163}.

\bibitem[Granlund(2004)]{granlund2004gnu}
Granlund, T.
\newblock Gnu mp: The gnu multiple precision arithmetic library.
\newblock \emph{http://gmplib. org/}, 2004.

\bibitem[Guo et~al.(2023)Guo, Greengard, Xing, and Kim]{guo2023lq}
Guo, H., Greengard, P., Xing, E.~P., and Kim, Y.
\newblock Lq-lora: Low-rank plus quantized matrix decomposition for efficient language model finetuning.
\newblock \emph{arXiv preprint arXiv:2311.12023}, 2023.

\bibitem[Gupta et~al.(2015)Gupta, Agrawal, Gopalakrishnan, and Narayanan]{gupta2015deep}
Gupta, S., Agrawal, A., Gopalakrishnan, K., and Narayanan, P.
\newblock Deep learning with limited numerical precision, 2015.

\bibitem[Han et~al.(2015)Han, Mao, and Dally]{han2015deep}
Han, S., Mao, H., and Dally, W.~J.
\newblock Deep compression: Compressing deep neural networks with pruning, trained quantization and huffman coding.
\newblock \emph{arXiv preprint arXiv:1510.00149}, 2015.

\bibitem[Hida et~al.(2008)Hida, Li, and Bailey]{hida2008Cpp}
Hida, Y., Li, S., and Bailey, D.
\newblock Library for double-double and quad-double arithmetic.
\newblock 01 2008.

\bibitem[Hinton et~al.(2015)Hinton, Vinyals, and Dean]{hinton2015distilling}
Hinton, G., Vinyals, O., and Dean, J.
\newblock Distilling the knowledge in a neural network, 2015.

\bibitem[Hoffmann et~al.(2022)Hoffmann, Borgeaud, Mensch, Buchatskaya, Cai, Rutherford, de~Las~Casas, Hendricks, Welbl, Clark, Hennigan, Noland, Millican, van~den Driessche, Damoc, Guy, Osindero, Simonyan, Elsen, Rae, Vinyals, and Sifre]{hoffmann2022training}
Hoffmann, J., Borgeaud, S., Mensch, A., Buchatskaya, E., Cai, T., Rutherford, E., de~Las~Casas, D., Hendricks, L.~A., Welbl, J., Clark, A., et~al.
\newblock Training compute-optimal large language models, 2022.

\bibitem[Hou et~al.(2019)Hou, Zhang, and Kwok]{hou2018analysis}
Hou, L., Zhang, R., and Kwok, J.~T.
\newblock Analysis of quantized models.
\newblock In \emph{International Conference on Learning Representations}, 2019.
\newblock URL \url{https://openreview.net/forum?id=ryM_IoAqYX}.

\bibitem[Hsieh et~al.(2023)Hsieh, Li, Yeh, Nakhost, Fujii, Ratner, Krishna, Lee, and Pfister]{hsieh2023distilling}
Hsieh, C.-Y., Li, C.-L., Yeh, C.-K., Nakhost, H., Fujii, Y., Ratner, A., Krishna, R., Lee, C.-Y., and Pfister, T.
\newblock Distilling step-by-step! outperforming larger language models with less training data and smaller model sizes, 2023.

\bibitem[Hu et~al.(2021)Hu, Shen, Wallis, Allen-Zhu, Li, Wang, Wang, and Chen]{hu2021lora}
Hu, E.~J., Shen, Y., Wallis, P., Allen-Zhu, Z., Li, Y., Wang, S., Wang, L., and Chen, W.
\newblock Lora: Low-rank adaptation of large language models.
\newblock \emph{arXiv preprint arXiv:2106.09685}, 2021.

\bibitem[Jacob et~al.(2018)Jacob, Kligys, Chen, Zhu, Tang, Howard, Adam, and Kalenichenko]{jacob2018quantization}
Jacob, B., Kligys, S., Chen, B., Zhu, M., Tang, M., Howard, A., Adam, H., and Kalenichenko, D.
\newblock Quantization and training of neural networks for efficient integer-arithmetic-only inference.
\newblock In \emph{Proceedings of the IEEE conference on computer vision and pattern recognition}, pp.\  2704--2713, 2018.

\bibitem[Jia \& Sandt(2021)Jia and Sandt]{jia2021A100}
Jia, Z. and Sandt, P.~V.
\newblock Zhe jia and peter van sandt. dissecting the ampere gpu architecture via microbenchmarking. gpu technology conference, 2021.
\newblock In \emph{GTC}, 2021.

\bibitem[Jia et~al.(2018)Jia, Maggioni, Staiger, and Scarpazza]{jia2018dissecting}
Jia, Z., Maggioni, M., Staiger, B., and Scarpazza, D.~P.
\newblock Dissecting the nvidia volta gpu architecture via microbenchmarking, 2018.

\bibitem[Kahan(2006)]{kahan2006futile}
Kahan, W.
\newblock How futile are mindless assessments of roundoff in floating-point computation.
\newblock \emph{Preprint, University of California, Berkeley}, 2006.

\bibitem[Kalamkar et~al.(2019)Kalamkar, Mudigere, Mellempudi, Das, Banerjee, Avancha, Vooturi, Jammalamadaka, Huang, Yuen, Yang, Park, Heinecke, Georganas, Srinivasan, Kundu, Smelyanskiy, Kaul, and Dubey]{kalamkar2019study}
Kalamkar, D., Mudigere, D., Mellempudi, N., Das, D., Banerjee, K., Avancha, S., Vooturi, D.~T., Jammalamadaka, N., Huang, J., Yuen, H., et~al.
\newblock A study of bfloat16 for deep learning training, 2019.

\bibitem[Korthikanti et~al.(2023)Korthikanti, Casper, Lym, McAfee, Andersch, Shoeybi, and Catanzaro]{korthikanti2023reducing}
Korthikanti, V.~A., Casper, J., Lym, S., McAfee, L., Andersch, M., Shoeybi, M., and Catanzaro, B.
\newblock Reducing activation recomputation in large transformer models.
\newblock \emph{Proceedings of Machine Learning and Systems}, 5, 2023.

\bibitem[Kuchaiev et~al.(2018)Kuchaiev, Ginsburg, Gitman, Lavrukhin, Li, Nguyen, Case, and Micikevicius]{kuchaiev2018mixedprecision}
Kuchaiev, O., Ginsburg, B., Gitman, I., Lavrukhin, V., Li, J., Nguyen, H., Case, C., and Micikevicius, P.
\newblock Mixed-precision training for nlp and speech recognition with openseq2seq, 2018.

\bibitem[Kuchaiev et~al.(2019)Kuchaiev, Li, Nguyen, Hrinchuk, Leary, Ginsburg, Kriman, Beliaev, Lavrukhin, Cook, Castonguay, Popova, Huang, and Cohen]{kuchaiev2019nemo}
Kuchaiev, O., Li, J., Nguyen, H., Hrinchuk, O., Leary, R., Ginsburg, B., Kriman, S., Beliaev, S., Lavrukhin, V., Cook, J., et~al.
\newblock Nemo: a toolkit for building ai applications using neural modules, 2019.

\bibitem[Kurtic et~al.(2022)Kurtic, Campos, Nguyen, Frantar, Kurtz, Fineran, Goin, and Alistarh]{kurtic2022optimal}
Kurtic, E., Campos, D., Nguyen, T., Frantar, E., Kurtz, M., Fineran, B., Goin, M., and Alistarh, D.
\newblock The optimal bert surgeon: Scalable and accurate second-order pruning for large language models, 2022.

\bibitem[Kwon et~al.(2022)Kwon, Kim, Bae, Yoo, Kim, Park, Kim, Ha, Sung, and Lee]{kwon2022alphatuning}
Kwon, S.~J., Kim, J., Bae, J., Yoo, K.~M., Kim, J.-H., Park, B., Kim, B., Ha, J.-W., Sung, N., and Lee, D.
\newblock Alphatuning: Quantization-aware parameter-efficient adaptation of large-scale pre-trained language models.
\newblock \emph{arXiv preprint arXiv:2210.03858}, 2022.

\bibitem[Lagunas et~al.(2021)Lagunas, Charlaix, Sanh, and Rush]{lagunas2021block}
Lagunas, F., Charlaix, E., Sanh, V., and Rush, A.~M.
\newblock Block pruning for faster transformers.
\newblock \emph{arXiv preprint arXiv:2109.04838}, 2021.

\bibitem[Le et~al.(2023)Le, Vyas, Shi, Karrer, Sari, Moritz, Williamson, Manohar, Adi, Mahadeokar, and Hsu]{le2023voicebox}
Le, M., Vyas, A., Shi, B., Karrer, B., Sari, L., Moritz, R., Williamson, M., Manohar, V., Adi, Y., Mahadeokar, J., and Hsu, W.-N.
\newblock Voicebox: Text-guided multilingual universal speech generation at scale, 2023.

\bibitem[Li et~al.(2017)Li, De, Xu, Studer, Samet, and Goldstein]{li2017training}
Li, H., De, S., Xu, Z., Studer, C., Samet, H., and Goldstein, T.
\newblock Training quantized nets: A deeper understanding, 2017.

\bibitem[Liu et~al.(2019)Liu, Ott, Goyal, Du, Joshi, Chen, Levy, Lewis, Zettlemoyer, and Stoyanov]{liu2019roberta}
Liu, Y., Ott, M., Goyal, N., Du, J., Joshi, M., Chen, D., Levy, O., Lewis, M., Zettlemoyer, L., and Stoyanov, V.
\newblock Roberta: A robustly optimized bert pretraining approach, 2019.

\bibitem[Loshchilov \& Hutter(2017)Loshchilov and Hutter]{loshchilov2017decoupled}
Loshchilov, I. and Hutter, F.
\newblock Decoupled weight decay regularization.
\newblock \emph{arXiv preprint arXiv:1711.05101}, 2017.

\bibitem[Micikevicius et~al.(2017)Micikevicius, Narang, Alben, Diamos, Elsen, Garcia, Ginsburg, Houston, Kuchaiev, Venkatesh, et~al.]{micikevicius2017mixed}
Micikevicius, P., Narang, S., Alben, J., Diamos, G., Elsen, E., Garcia, D., Ginsburg, B., Houston, M., Kuchaiev, O., Venkatesh, G., et~al.
\newblock Mixed precision training.
\newblock \emph{arXiv preprint arXiv:1710.03740}, 2017.

\bibitem[Micikevicius et~al.(2018)Micikevicius, Narang, Alben, Diamos, Elsen, Garcia, Ginsburg, Houston, Kuchaiev, Venkatesh, and Wu]{micikevicius2018mixed}
Micikevicius, P., Narang, S., Alben, J., Diamos, G., Elsen, E., Garcia, D., Ginsburg, B., Houston, M., Kuchaiev, O., Venkatesh, G., and Wu, H.
\newblock Mixed precision training, 2018.

\bibitem[Micikevicius et~al.(2022)Micikevicius, Stosic, Burgess, Cornea, Dubey, Grisenthwaite, Ha, Heinecke, Judd, Kamalu, Mellempudi, Oberman, Shoeybi, Siu, and Wu]{micikevicius2022fp8}
Micikevicius, P., Stosic, D., Burgess, N., Cornea, M., Dubey, P., Grisenthwaite, R., Ha, S., Heinecke, A., Judd, P., Kamalu, J., et~al.
\newblock Fp8 formats for deep learning, 2022.

\bibitem[Muller et~al.(2018)Muller, Brisebarre, De~Dinechin, Jeannerod, Lefevre, Melquiond, Revol, Stehl{\'e}, Torres, et~al.]{muller2018handbook}
Muller, J.-M., Brisebarre, N., De~Dinechin, F., Jeannerod, C.-P., Lefevre, V., Melquiond, G., Revol, N., Stehl{\'e}, D., Torres, S., et~al.
\newblock \emph{Handbook of floating-point arithmetic}.
\newblock Springer, 2018.

\bibitem[OpenAI et~al.(2023)OpenAI, :, Achiam, Adler, Agarwal, Ahmad, Akkaya, Aleman, Almeida, Altenschmidt, Altman, Anadkat, Avila, Babuschkin, Balaji, Balcom, Baltescu, Bao, Bavarian, Belgum, Bello, Berdine, Bernadett-Shapiro, Berner, Bogdonoff, Boiko, Boyd, Brakman, Brockman, Brooks, Brundage, Button, Cai, Campbell, Cann, Carey, Carlson, Carmichael, Chan, Chang, Chantzis, Chen, Chen, Chen, Chen, Chen, Chess, Cho, Chu, Chung, Cummings, Currier, Dai, Decareaux, Degry, Deutsch, Deville, Dhar, Dohan, Dowling, Dunning, Ecoffet, Eleti, Eloundou, Farhi, Fedus, Felix, Fishman, Forte, Fulford, Gao, Georges, Gibson, Goel, Gogineni, Goh, Gontijo-Lopes, Gordon, Grafstein, Gray, Greene, Gross, Gu, Guo, Hallacy, Han, Harris, He, Heaton, Heidecke, Hesse, Hickey, Hickey, Hoeschele, Houghton, Hsu, Hu, Hu, Huizinga, Jain, Jain, Jang, Jiang, Jiang, Jin, Jin, Jomoto, Jonn, Jun, Kaftan, Łukasz Kaiser, Kamali, Kanitscheider, Keskar, Khan, Kilpatrick, Kim, Kim, Kim, Kirchner, Kiros, Knight, Kokotajlo, Łukasz Kondraciuk,
  Kondrich, Konstantinidis, Kosic, Krueger, Kuo, Lampe, Lan, Lee, Leike, Leung, Levy, Li, Lim, Lin, Lin, Litwin, Lopez, Lowe, Lue, Makanju, Malfacini, Manning, Markov, Markovski, Martin, Mayer, Mayne, McGrew, McKinney, McLeavey, McMillan, McNeil, Medina, Mehta, Menick, Metz, Mishchenko, Mishkin, Monaco, Morikawa, Mossing, Mu, Murati, Murk, Mély, Nair, Nakano, Nayak, Neelakantan, Ngo, Noh, Ouyang, O'Keefe, Pachocki, Paino, Palermo, Pantuliano, Parascandolo, Parish, Parparita, Passos, Pavlov, Peng, Perelman, de~Avila Belbute~Peres, Petrov, de~Oliveira~Pinto, Michael, Pokorny, Pokrass, Pong, Powell, Power, Power, Proehl, Puri, Radford, Rae, Ramesh, Raymond, Real, Rimbach, Ross, Rotsted, Roussez, Ryder, Saltarelli, Sanders, Santurkar, Sastry, Schmidt, Schnurr, Schulman, Selsam, Sheppard, Sherbakov, Shieh, Shoker, Shyam, Sidor, Sigler, Simens, Sitkin, Slama, Sohl, Sokolowsky, Song, Staudacher, Such, Summers, Sutskever, Tang, Tezak, Thompson, Tillet, Tootoonchian, Tseng, Tuggle, Turley, Tworek, Uribe, Vallone,
  Vijayvergiya, Voss, Wainwright, Wang, Wang, Wang, Ward, Wei, Weinmann, Welihinda, Welinder, Weng, Weng, Wiethoff, Willner, Winter, Wolrich, Wong, Workman, Wu, Wu, Wu, Xiao, Xu, Yoo, Yu, Yuan, Zaremba, Zellers, Zhang, Zhang, Zhao, Zheng, Zhuang, Zhuk, and Zoph]{openai2023gpt4}
OpenAI, :, Achiam, J., Adler, S., Agarwal, S., Ahmad, L., Akkaya, I., Aleman, F.~L., Almeida, D., Altenschmidt, J., et~al.
\newblock Gpt-4 technical report, 2023.

\bibitem[Park et~al.(2018)Park, Lee, Oh, Ha, and Lee]{park2018training}
Park, H., Lee, J.~H., Oh, Y., Ha, S., and Lee, S.
\newblock Training deep neural network in limited precision, 2018.

\bibitem[Peng et~al.(2023)Peng, Wu, Wei, Zhao, Yang, Liu, Xiong, Yang, Ni, Hu, Li, Zhang, Li, Ning, Wang, Zhang, Liu, Chau, Hu, and Cheng]{peng2023fp8lm}
Peng, H., Wu, K., Wei, Y., Zhao, G., Yang, Y., Liu, Z., Xiong, Y., Yang, Z., Ni, B., Hu, J., et~al.
\newblock Fp8-lm: Training fp8 large language models, 2023.

\bibitem[Perez et~al.(2023)Perez, Zhang, Briggs, Blake, Levy-Kramer, Balanca, Luschi, Barlow, and Fitzgibbon]{perez2023training}
Perez, S.~P., Zhang, Y., Briggs, J., Blake, C., Levy-Kramer, J., Balanca, P., Luschi, C., Barlow, S., and Fitzgibbon, A.~W.
\newblock Training and inference of large language models using 8-bit floating point, 2023.

\bibitem[Priest(1991)]{priest1991Arithmetic}
Priest, D.
\newblock Algorithms for arbitrary precision floating point arithmetic.
\newblock In \emph{[1991] Proceedings 10th IEEE Symposium on Computer Arithmetic}, pp.\  132--143, 1991.
\newblock \doi{10.1109/ARITH.1991.145549}.

\bibitem[Priest(1992)]{priest1992Arithmetic}
Priest, D.~M.
\newblock \emph{On properties of floating point arithmetics: numerical stability and the cost of accurate computations}.
\newblock PhD thesis, University of California at Berkeley, USA, 1992.
\newblock UMI Order No. GAX93-30692.

\bibitem[Rae et~al.(2022)Rae, Borgeaud, Cai, Millican, Hoffmann, Song, Aslanides, Henderson, Ring, Young, Rutherford, Hennigan, Menick, Cassirer, Powell, van~den Driessche, Hendricks, Rauh, Huang, Glaese, Welbl, Dathathri, Huang, Uesato, Mellor, Higgins, Creswell, McAleese, Wu, Elsen, Jayakumar, Buchatskaya, Budden, Sutherland, Simonyan, Paganini, Sifre, Martens, Li, Kuncoro, Nematzadeh, Gribovskaya, Donato, Lazaridou, Mensch, Lespiau, Tsimpoukelli, Grigorev, Fritz, Sottiaux, Pajarskas, Pohlen, Gong, Toyama, de~Masson~d'Autume, Li, Terzi, Mikulik, Babuschkin, Clark, de~Las~Casas, Guy, Jones, Bradbury, Johnson, Hechtman, Weidinger, Gabriel, Isaac, Lockhart, Osindero, Rimell, Dyer, Vinyals, Ayoub, Stanway, Bennett, Hassabis, Kavukcuoglu, and Irving]{rae2022scaling}
Rae, J.~W., Borgeaud, S., Cai, T., Millican, K., Hoffmann, J., Song, F., Aslanides, J., Henderson, S., Ring, R., Young, S., et~al.
\newblock Scaling language models: Methods, analysis \& insights from training gopher, 2022.

\bibitem[Rastegari et~al.(2016)Rastegari, Ordonez, Redmon, and Farhadi]{rastegari2016xnornet}
Rastegari, M., Ordonez, V., Redmon, J., and Farhadi, A.
\newblock Xnor-net: Imagenet classification using binary convolutional neural networks, 2016.

\bibitem[Ruder(2017)]{ruder2017overview}
Ruder, S.
\newblock An overview of gradient descent optimization algorithms, 2017.

\bibitem[Sa et~al.(2018)Sa, Leszczynski, Zhang, Marzoev, Aberger, Olukotun, and Ré]{desa2018highaccuracy}
Sa, C.~D., Leszczynski, M., Zhang, J., Marzoev, A., Aberger, C.~R., Olukotun, K., and Ré, C.
\newblock High-accuracy low-precision training, 2018.

\bibitem[Sakr \& Shanbhag(2018)Sakr and Shanbhag]{sakr2018pertensor}
Sakr, C. and Shanbhag, N.
\newblock Per-tensor fixed-point quantization of the back-propagation algorithm, 2018.

\bibitem[Sanh et~al.(2020)Sanh, Debut, Chaumond, and Wolf]{sanh2020distilbert}
Sanh, V., Debut, L., Chaumond, J., and Wolf, T.
\newblock Distilbert, a distilled version of bert: smaller, faster, cheaper and lighter, 2020.

\bibitem[Shoeybi et~al.(2020)Shoeybi, Patwary, Puri, LeGresley, Casper, and Catanzaro]{shoeybi2020megatronlm}
Shoeybi, M., Patwary, M., Puri, R., LeGresley, P., Casper, J., and Catanzaro, B.
\newblock Megatron-lm: Training multi-billion parameter language models using model parallelism, 2020.

\bibitem[Su et~al.(2024)Su, Ahmed, Lu, Pan, Bo, and Liu]{su2024roformer}
Su, J., Ahmed, M., Lu, Y., Pan, S., Bo, W., and Liu, Y.
\newblock Roformer: Enhanced transformer with rotary position embedding.
\newblock \emph{Neurocomputing}, 568:\penalty0 127063, 2024.

\bibitem[Sun et~al.(2019{\natexlab{a}})Sun, Choi, Chen, Wang, Venkataramani, Srinivasan, Cui, Zhang, and Gopalakrishnan]{sun2019hybrid}
Sun, X., Choi, J., Chen, C.-Y., Wang, N., Venkataramani, S., Srinivasan, V.~V., Cui, X., Zhang, W., and Gopalakrishnan, K.
\newblock Hybrid 8-bit floating point (hfp8) training and inference for deep neural networks.
\newblock \emph{Advances in neural information processing systems}, 32, 2019{\natexlab{a}}.

\bibitem[Sun et~al.(2019{\natexlab{b}})Sun, Choi, Chen, Wang, Venkataramani, Srinivasan, Cui, Zhang, and Gopalakrishnan]{xiao2019hybrid8bit}
Sun, X., Choi, J., Chen, C.-Y., Wang, N., Venkataramani, S., Srinivasan, V.~V., Cui, X., Zhang, W., and Gopalakrishnan, K.
\newblock Hybrid 8-bit floating point (hfp8) training and inference for deep neural networks.
\newblock In \emph{Advances in Neural Information Processing Systems}, volume~32. Curran Associates, Inc., 2019{\natexlab{b}}.
\newblock URL \url{https://proceedings.neurips.cc/paper_files/paper/2019/file/65fc9fb4897a89789352e211ca2d398f-Paper.pdf}.

\bibitem[Team et~al.(2023)Team, Anil, Borgeaud, Wu, Alayrac, Yu, Soricut, Schalkwyk, Dai, Hauth, Millican, Silver, Petrov, Johnson, Antonoglou, Schrittwieser, Glaese, Chen, Pitler, Lillicrap, Lazaridou, Firat, Molloy, Isard, Barham, Hennigan, Lee, Viola, Reynolds, Xu, Doherty, Collins, Meyer, Rutherford, Moreira, Ayoub, Goel, Tucker, Piqueras, Krikun, Barr, Savinov, Danihelka, Roelofs, White, Andreassen, von Glehn, Yagati, Kazemi, Gonzalez, Khalman, Sygnowski, Frechette, Smith, Culp, Proleev, Luan, Chen, Lottes, Schucher, Lebron, Rrustemi, Clay, Crone, Kocisky, Zhao, Perz, Yu, Howard, Bloniarz, Rae, Lu, Sifre, Maggioni, Alcober, Garrette, Barnes, Thakoor, Austin, Barth-Maron, Wong, Joshi, Chaabouni, Fatiha, Ahuja, Liu, Li, Cogan, Chen, Jia, Gu, Zhang, Grimstad, Hartman, Chadwick, Tomar, Garcia, Senter, Taropa, Pillai, Devlin, Laskin, de~Las~Casas, Valter, Tao, Blanco, Badia, Reitter, Chen, Brennan, Rivera, Brin, Iqbal, Surita, Labanowski, Rao, Winkler, Parisotto, Gu, Olszewska, Zhang, Addanki, Miech, Louis,
  Shafey, Teplyashin, Brown, Catt, Attaluri, Balaguer, Xiang, Wang, Ashwood, Briukhov, Webson, Ganapathy, Sanghavi, Kannan, Chang, Stjerngren, Djolonga, Sun, Bapna, Aitchison, Pejman, Michalewski, Yu, Wang, Love, Ahn, Bloxwich, Han, Humphreys, Sellam, Bradbury, Godbole, Samangooei, Damoc, Kaskasoli, Arnold, Vasudevan, Agrawal, Riesa, Lepikhin, Tanburn, Srinivasan, Lim, Hodkinson, Shyam, Ferret, Hand, Garg, Paine, Li, Li, Giang, Neitz, Abbas, York, Reid, Cole, Chowdhery, Das, Rogozińska, Nikolaev, Sprechmann, Nado, Zilka, Prost, He, Monteiro, Mishra, Welty, Newlan, Jia, Allamanis, Hu, de~Liedekerke, Gilmer, Saroufim, Rijhwani, Hou, Shrivastava, Baddepudi, Goldin, Ozturel, Cassirer, Xu, Sohn, Sachan, Amplayo, Swanson, Petrova, Narayan, Guez, Brahma, Landon, Patel, Zhao, Villela, Wang, Jia, Rahtz, Giménez, Yeung, Lin, Keeling, Georgiev, Mincu, Wu, Haykal, Saputro, Vodrahalli, Qin, Cankara, Sharma, Fernando, Hawkins, Neyshabur, Kim, Hutter, Agrawal, Castro-Ros, van~den Driessche, Wang, Yang, yiin Chang,
  Komarek, McIlroy, Lučić, Zhang, Farhan, Sharman, Natsev, Michel, Cheng, Bansal, Qiao, Cao, Shakeri, Butterfield, Chung, Rubenstein, Agrawal, Mensch, Soparkar, Lenc, Chung, Pope, Maggiore, Kay, Jhakra, Wang, Maynez, Phuong, Tobin, Tacchetti, Trebacz, Robinson, Katariya, Riedel, Bailey, Xiao, Ghelani, Aroyo, Slone, Houlsby, Xiong, Yang, Gribovskaya, Adler, Wirth, Lee, Li, Kagohara, Pavagadhi, Bridgers, Bortsova, Ghemawat, Ahmed, Liu, Powell, Bolina, Iinuma, Zablotskaia, Besley, Chung, Dozat, Comanescu, Si, Greer, Su, Polacek, Kaufman, Tokumine, Hu, Buchatskaya, Miao, Elhawaty, Siddhant, Tomasev, Xing, Greer, Miller, Ashraf, Roy, Zhang, Ma, Filos, Besta, Blevins, Klimenko, Yeh, Changpinyo, Mu, Chang, Pajarskas, Muir, Cohen, Lan, Haridasan, Marathe, Hansen, Douglas, Samuel, Wang, Austin, Lan, Jiang, Chiu, Lorenzo, Sjösund, Cevey, Gleicher, Avrahami, Boral, Srinivasan, Selo, May, Aisopos, Hussenot, Soares, Baumli, Chang, Recasens, Caine, Pritzel, Pavetic, Pardo, Gergely, Frye, Ramasesh, Horgan, Badola,
  Kassner, Roy, Dyer, Campos, Tomala, Tang, Badawy, White, Mustafa, Lang, Jindal, Vikram, Gong, Caelles, Hemsley, Thornton, Feng, Stokowiec, Zheng, Thacker, Çağlar Ünlü, Zhang, Saleh, Svensson, Bileschi, Patil, Anand, Ring, Tsihlas, Vezer, Selvi, Shevlane, Rodriguez, Kwiatkowski, Daruki, Rong, Dafoe, FitzGerald, Gu-Lemberg, Khan, Hendricks, Pellat, Feinberg, Cobon-Kerr, Sainath, Rauh, Hashemi, Ives, Hasson, Li, Noland, Cao, Byrd, Hou, Wang, Sottiaux, Paganini, Lespiau, Moufarek, Hassan, Shivakumar, van Amersfoort, Mandhane, Joshi, Goyal, Tung, Brock, Sheahan, Misra, Li, Rakićević, Dehghani, Liu, Mittal, Oh, Noury, Sezener, Huot, Lamm, Cao, Chen, Elsayed, Chi, Mahdieh, Tenney, Hua, Petrychenko, Kane, Scandinaro, Jain, Uesato, Datta, Sadovsky, Bunyan, Rabiej, Wu, Zhang, Vasudevan, Leurent, Alnahlawi, Georgescu, Wei, Zheng, Chan, Rabinovitch, Stanczyk, Zhang, Steiner, Naskar, Azzam, Johnson, Paszke, Chiu, Elias, Mohiuddin, Muhammad, Miao, Lee, Vieillard, Potluri, Park, Davoodi, Zhang, Stanway, Garmon,
  Karmarkar, Dong, Lee, Kumar, Zhou, Evens, Isaac, Chen, Jia, Levskaya, Zhu, Gorgolewski, Grabowski, Mao, Magni, Yao, Snaider, Casagrande, Suganthan, Palmer, Irving, Loper, Faruqui, Arkatkar, Chen, Shafran, Fink, Castaño, Giannoumis, Kim, Rybiński, Sreevatsa, Prendki, Soergel, Goedeckemeyer, Gierke, Jafari, Gaba, Wiesner, Wright, Wei, Vashisht, Kulizhskaya, Hoover, Le, Li, Iwuanyanwu, Liu, Ramirez, Khorlin, Cui, LIN, Georgiev, Wu, Aguilar, Pallo, Chakladar, Repina, Wu, van~der Weide, Ponnapalli, Kaplan, Simsa, Li, Dousse, Yang, Piper, Ie, Lui, Pasumarthi, Lintz, Vijayakumar, Thiet, Andor, Valenzuela, Paduraru, Peng, Lee, Zhang, Greene, Nguyen, Kurylowicz, Velury, Krause, Hardin, Dixon, Janzer, Choo, Feng, Zhang, Singhal, Latkar, Zhang, Le, Abellan, Du, McKinnon, Antropova, Bolukbasi, Keller, Reid, Finchelstein, Raad, Crocker, Hawkins, Dadashi, Gaffney, Lall, Franko, Filonov, Bulanova, Leblond, Yadav, Chung, Askham, Cobo, Xu, Fischer, Xu, Sorokin, Alberti, Lin, Evans, Zhou, Dimitriev, Forbes, Banarse, Tung,
  Liu, Omernick, Bishop, Kumar, Sterneck, Foley, Jain, Mishra, Xia, Bos, Cideron, Amid, Piccinno, Wang, Banzal, Gurita, Noga, Shah, Mankowitz, Polozov, Kushman, Krakovna, Brown, Bateni, Duan, Firoiu, Thotakuri, Natan, Mohananey, Geist, Mudgal, Girgin, Li, Ye, Roval, Tojo, Kwong, Lee-Thorp, Yew, Yuan, Bagri, Sinopalnikov, Ramos, Mellor, Sharma, Severyn, Lai, Wu, Cheng, Miller, Sonnerat, Vnukov, Greig, Beattie, Caveness, Bai, Eisenschlos, Korchemniy, Tsai, Jasarevic, Kong, Dao, Zheng, Liu, Yang, Zhu, Geller, Teh, Sanmiya, Gladchenko, Trdin, Sozanschi, Toyama, Rosen, Tavakkol, Xue, Elkind, Woodman, Carpenter, Papamakarios, Kemp, Kafle, Grunina, Sinha, Talbert, Goyal, Wu, Owusu-Afriyie, Du, Thornton, Pont-Tuset, Narayana, Li, Fatehi, Wieting, Ajmeri, Uria, Zhu, Ko, Knight, Héliou, Niu, Gu, Pang, Tran, Li, Levine, Stolovich, Kalb, Santamaria-Fernandez, Goenka, Yustalim, Strudel, Elqursh, Lakshminarayanan, Deck, Upadhyay, Lee, Dusenberry, Li, Wang, Levin, Hoffmann, Holtmann-Rice, Bachem, Yue, Arora, Malmi,
  Mirylenka, Tan, Koh, Yeganeh, Põder, Zheng, Pongetti, Tariq, Sun, Ionita, Seyedhosseini, Tafti, Kotikalapudi, Liu, Gulati, Liu, Ye, Chrzaszcz, Wang, Sethi, Li, Brown, Singh, Fan, Parisi, Stanton, Kuang, Koverkathu, Choquette-Choo, Li, Lu, Ittycheriah, Shroff, Sun, Varadarajan, Bahargam, Willoughby, Gaddy, Dasgupta, Desjardins, Cornero, Robenek, Mittal, Albrecht, Shenoy, Moiseev, Jacobsson, Ghaffarkhah, Rivière, Walton, Crepy, Parrish, Liu, Zhou, Farabet, Radebaugh, Srinivasan, van~der Salm, Fidjeland, Scellato, Latorre-Chimoto, Klimczak-Plucińska, Bridson, de~Cesare, Hudson, Mendolicchio, Walker, Morris, Penchev, Mauger, Guseynov, Reid, Odoom, Loher, Cotruta, Yenugula, Grewe, Petrushkina, Duerig, Sanchez, Yadlowsky, Shen, Globerson, Kurzrok, Webb, Dua, Li, Lahoti, Bhupatiraju, Hurt, Qureshi, Agarwal, Shani, Eyal, Khare, Belle, Wang, Tekur, Kale, Wei, Sang, Saeta, Liechty, Sun, Zhao, Lee, Nayak, Fritz, Vuyyuru, Aslanides, Vyas, Wicke, Ma, Bilal, Eltyshev, Balle, Martin, Cate, Manyika, Amiri, Kim, Xiong,
  Kang, Luisier, Tripuraneni, Madras, Guo, Waters, Wang, Ainslie, Baldridge, Zhang, Pruthi, Bauer, Yang, Mansour, Gelman, Xu, Polovets, Liu, Cai, Chen, Sheng, Xue, Ozair, Yu, Angermueller, Li, Wang, Wiesinger, Koukoumidis, Tian, Iyer, Gurumurthy, Goldenson, Shah, Blake, Yu, Urbanowicz, Palomaki, Fernando, Brooks, Durden, Mehta, Momchev, Rahimtoroghi, Georgaki, Raul, Ruder, Redshaw, Lee, Jalan, Li, Perng, Hechtman, Schuh, Nasr, Chen, Milan, Mikulik, Strohman, Franco, Green, Hassabis, Kavukcuoglu, Dean, and Vinyals]{geminiteam2023gemini}
Team, G., Anil, R., Borgeaud, S., Wu, Y., Alayrac, J.-B., Yu, J., Soricut, R., Schalkwyk, J., Dai, A.~M., Hauth, A., et~al.
\newblock Gemini: A family of highly capable multimodal models, 2023.

\bibitem[Thoppilan et~al.(2022)Thoppilan, De~Freitas, Hall, Shazeer, Kulshreshtha, Cheng, Jin, Bos, Baker, Du, et~al.]{thoppilan2022lamda}
Thoppilan, R., De~Freitas, D., Hall, J., Shazeer, N., Kulshreshtha, A., Cheng, H.-T., Jin, A., Bos, T., Baker, L., Du, Y., et~al.
\newblock Lamda: Language models for dialog applications.
\newblock \emph{arXiv preprint arXiv:2201.08239}, 2022.

\bibitem[Touvron et~al.(2023)Touvron, Lavril, Izacard, Martinet, Lachaux, Lacroix, Rozière, Goyal, Hambro, Azhar, Rodriguez, Joulin, Grave, and Lample]{touvron2023llama}
Touvron, H., Lavril, T., Izacard, G., Martinet, X., Lachaux, M.-A., Lacroix, T., Rozière, B., Goyal, N., Hambro, E., Azhar, F., et~al.
\newblock Llama: Open and efficient foundation language models, 2023.

\bibitem[Wang et~al.(2019)Wang, Singh, Michael, Hill, Levy, and Bowman]{wang2019glue}
Wang, A., Singh, A., Michael, J., Hill, F., Levy, O., and Bowman, S.~R.
\newblock Glue: A multi-task benchmark and analysis platform for natural language understanding, 2019.

\bibitem[Wang et~al.(2018{\natexlab{a}})Wang, Choi, Brand, Chen, and Gopalakrishnan]{wang20188bit}
Wang, N., Choi, J., Brand, D., Chen, C.-Y., and Gopalakrishnan, K.
\newblock Training deep neural networks with 8-bit floating point numbers.
\newblock In \emph{Advances in Neural Information Processing Systems}, volume~31. Curran Associates, Inc., 2018{\natexlab{a}}.
\newblock URL \url{https://proceedings.neurips.cc/paper_files/paper/2018/file/335d3d1cd7ef05ec77714a215134914c-Paper.pdf}.

\bibitem[Wang et~al.(2018{\natexlab{b}})Wang, Choi, Brand, Chen, and Gopalakrishnan]{wang2018training}
Wang, N., Choi, J., Brand, D., Chen, C.-Y., and Gopalakrishnan, K.
\newblock Training deep neural networks with 8-bit floating point numbers.
\newblock \emph{Advances in neural information processing systems}, 31, 2018{\natexlab{b}}.

\bibitem[Wei et~al.(2023)Wei, Gonugondla, Ahmad, Wang, Ray, Qian, Li, Kumar, Wang, Tian, Sun, Athiwaratkun, Shang, Ramanathan, Bhatia, and Xiang]{wei2023greener}
Wei, X., Gonugondla, S., Ahmad, W., Wang, S., Ray, B., Qian, H., Li, X., Kumar, V., Wang, Z., Tian, Y., et~al.
\newblock Greener yet powerful: Taming large code generation models with quantization, 2023.

\bibitem[Wolf et~al.(2019)Wolf, Debut, Sanh, Chaumond, Delangue, Moi, Cistac, Rault, Louf, Funtowicz, et~al.]{wolf2019huggingface}
Wolf, T., Debut, L., Sanh, V., Chaumond, J., Delangue, C., Moi, A., Cistac, P., Rault, T., Louf, R., Funtowicz, M., et~al.
\newblock Huggingface's transformers: State-of-the-art natural language processing.
\newblock \emph{arXiv preprint arXiv:1910.03771}, 2019.

\bibitem[Wortsman et~al.(2023)Wortsman, Dettmers, Zettlemoyer, Morcos, Farhadi, and Schmidt]{wortsman2023stable}
Wortsman, M., Dettmers, T., Zettlemoyer, L., Morcos, A., Farhadi, A., and Schmidt, L.
\newblock Stable and low-precision training for large-scale vision-language models, 2023.

\bibitem[Wu et~al.(2018)Wu, Li, Chen, and Shi]{wu2018training}
Wu, S., Li, G., Chen, F., and Shi, L.
\newblock Training and inference with integers in deep neural networks, 2018.

\bibitem[Xi et~al.(2023)Xi, Li, Chen, and Zhu]{xi2023training}
Xi, H., Li, C., Chen, J., and Zhu, J.
\newblock Training transformers with 4-bit integers.
\newblock In \emph{Thirty-seventh Conference on Neural Information Processing Systems}, 2023.
\newblock URL \url{https://openreview.net/forum?id=H9hWlfMT6O}.

\bibitem[Xia et~al.(2023)Xia, Zheng, Li, Zhuang, Zhou, Qiu, Li, Lin, and Song]{xia2023flash}
Xia, H., Zheng, Z., Li, Y., Zhuang, D., Zhou, Z., Qiu, X., Li, Y., Lin, W., and Song, S.~L.
\newblock Flash-llm: Enabling cost-effective and highly-efficient large generative model inference with unstructured sparsity.
\newblock \emph{arXiv preprint arXiv:2309.10285}, 2023.

\bibitem[Xia et~al.(2022)Xia, Zhong, and Chen]{xia2022structured}
Xia, M., Zhong, Z., and Chen, D.
\newblock Structured pruning learns compact and accurate models.
\newblock \emph{arXiv preprint arXiv:2204.00408}, 2022.

\bibitem[Xiao et~al.(2023)Xiao, Lin, Seznec, Wu, Demouth, and Han]{xiao2023smoothquant}
Xiao, G., Lin, J., Seznec, M., Wu, H., Demouth, J., and Han, S.
\newblock Smoothquant: Accurate and efficient post-training quantization for large language models, 2023.

\bibitem[Xu et~al.(2023)Xu, Xie, Gu, Chen, Chang, Zhang, Chen, Zhang, and Tian]{xu2023qa}
Xu, Y., Xie, L., Gu, X., Chen, X., Chang, H., Zhang, H., Chen, Z., Zhang, X., and Tian, Q.
\newblock Qa-lora: Quantization-aware low-rank adaptation of large language models.
\newblock \emph{arXiv preprint arXiv:2309.14717}, 2023.

\bibitem[Yao et~al.(2023)Yao, Wu, Li, Youn, and He]{yao2023zeroquant}
Yao, Z., Wu, X., Li, C., Youn, S., and He, Y.
\newblock Zeroquant-v2: Exploring post-training quantization in llms from comprehensive study to low rank compensation.
\newblock \emph{arXiv preprint arXiv:2303.08302}, 2023.

\bibitem[Yu et~al.(2022{\natexlab{a}})Yu, Huang, Wang, Cheng, Chu, and Cui]{yu2022width}
Yu, F., Huang, K., Wang, M., Cheng, Y., Chu, W., and Cui, L.
\newblock Width \& depth pruning for vision transformers.
\newblock In \emph{Proceedings of the AAAI Conference on Artificial Intelligence}, volume~36, pp.\  3143--3151, 2022{\natexlab{a}}.

\bibitem[Yu \& De~Sa(2021)Yu and De~Sa]{Yu2021MCT}
Yu, T. and De~Sa, C.~M.
\newblock Representing hyperbolic space accurately using multi-component floats.
\newblock In Ranzato, M., Beygelzimer, A., Dauphin, Y., Liang, P., and Vaughan, J.~W. (eds.), \emph{Advances in Neural Information Processing Systems}, volume~34, pp.\  15570--15581. Curran Associates, Inc., 2021.
\newblock URL \url{https://proceedings.neurips.cc/paper_files/paper/2021/file/832353270aacb6e3322f493a66aaf5b9-Paper.pdf}.

\bibitem[Yu et~al.(2022{\natexlab{b}})Yu, Guo, Li, Yuan, and Sa]{yu2022mctensor}
Yu, T., Guo, W., Li, J.~C., Yuan, T., and Sa, C.~D.
\newblock Mctensor: A high-precision deep learning library with multi-component floating-point, 2022{\natexlab{b}}.

\bibitem[Zamirai et~al.(2020)Zamirai, Zhang, Aberger, and De~Sa]{zamirai2020revisiting}
Zamirai, P., Zhang, J., Aberger, C.~R., and De~Sa, C.
\newblock Revisiting bfloat16 training.
\newblock \emph{arXiv preprint arXiv:2010.06192}, 2020.

\bibitem[Zhang et~al.(2022)Zhang, Roller, Goyal, Artetxe, Chen, Chen, Dewan, Diab, Li, Lin, Mihaylov, Ott, Shleifer, Shuster, Simig, Koura, Sridhar, Wang, and Zettlemoyer]{zhang2022opt}
Zhang, S., Roller, S., Goyal, N., Artetxe, M., Chen, M., Chen, S., Dewan, C., Diab, M., Li, X., Lin, X.~V., et~al.
\newblock Opt: Open pre-trained transformer language models, 2022.

\bibitem[Zhou et~al.(2021)Zhou, Ma, Zhu, Liu, Zhang, Yuan, Sun, and Li]{zhou2021learning}
Zhou, A., Ma, Y., Zhu, J., Liu, J., Zhang, Z., Yuan, K., Sun, W., and Li, H.
\newblock Learning n:m fine-grained structured sparse neural networks from scratch, 2021.

\end{thebibliography}
\bibliographystyle{icml2023}

\newpage
\appendix
\onecolumn

\section{Floating point units}
\label{appsec:FPU}
Floating-point representation uses a sign bit to indicate positive or negative numbers, an exponent to determine scale, and a mantissa for significant digits, enabling efficient handling of a wide range of numbers with potential for precision errors. Different floating-point formats offer varying benefits and trade-offs. Single Precision (FP$32$) provides wide range and reasonable precision, while consuming more resources. Half Precision (FP$16$) reduces memory usage and improves efficiency, but sacrifices precision and range. Brain floating point (BF$16$) as another $16$-bit format has a much bigger dynamic range (same as FP$32$), while having a worse precision than FP$16$. FP$8$ (two versions) could further reduce resources, suitable for constrained environments, but with even more limited precision and range. 

We present different formats referenced in the paper along with their exponent and mantissa bits.
\begin{table}[ht]
\centering
\caption{Floating-Point Precisions and ULPs}
\label{tab:fp_formats}
\begin{tabular}{lccc}
\toprule
Precision & \#Exponent bits & \makecell{\#Mantissa \\(significand) bits} & $\ulp(1)$ \\
\hline
Single (FP$32$)& $8$ & $23$ & $2^{-23}$ \\
\hline
Half (FP$16$) & $5$ & $10$ & $2^{-10}$ \\
\hline
BF$16$ & $8$ & $7$ & $2^{-7}$ \\
\hline
FP8 E$4$M$3$ & $4$ & $3$ & $2^{-3}$ \\
\hline
FP8 E$5$M$2$ & $5$ & $2$ & $2^{-2}$ \\
\bottomrule
\end{tabular}
\label{tab:float_precisions}
\end{table}

\section{Related Work}
\label{sec:related-work}
\textbf{Low Precision and Quantization-aware Training.} 
Fully quantized training attempts to downscale numerical precisions but not to compromise accuracy, mainly for large-scale training, using FP16 \cite{micikevicius2017mixed}, BF16 \cite{kalamkar2019study}, FP8 \cite{wang2018training,sun2019hybrid}, 
and other combination of integer types \cite{
banner2018scalable, chen2020statistical}.
\citet{micikevicius2018mixed} developed a mixed precision strategy that maintains master weight in FP32 only whereas others are in lower precision of FP16. 
\citet{xi2023training} recently proposed a training method using INT4 but without customized data types, compatitable with contemporary hardwares. In parallel, 
\citet{peng2023fp8lm} proposed a new mixed-precision strategy, gradually incorporating 8-bit gradients, optimizer states in an incremental manner, under distributed settings. When it comes to fine-tuning setings, LoRA \cite{hu2021lora} leverage structure of matrix to update in low-rank.  \citet{dettmers2023qlora,xu2023qa, guo2023lq} proposed various variants of LoRA more in memory and computationally efficient manners. Overall, these works develop training strategies based on numerical structures like low-rank over attention matrices and/or sparsity over parameters in each layer, numerical scale of each variables used for gradient updates. However, they lack of thorough diagnosis on imprecision errors, which has been depriving potential algorithmic developments in numeric precision levels. 
\citet{zamirai2020revisiting} proposed to adopt Kahan summation \cite{kahan2006futile} and stochastic rounding (SR) \cite{croci2022stochastic} to alleviate the influence of imprecision and lost arithmetic at the model parameter update step. 

\textbf{Pruning and Distillation.} Pruning \cite{han2015deep, kurtic2022optimal} removes redundant parameters from the network. The goal is to maintain prediction quality of the model while shrinking its size, and therewith increasing its efficiency. 
distillation \cite{hsieh2023distilling, hinton2015distilling, hsieh2023distilling} transfers knowledge from a large model to a smaller one. Pruning can be combined with distillation approach to further reduce model parameters \cite{sanh2020distilbert, lagunas2021block, xia2022structured}. Structured pruning removes whole components of the network such as neurons, heads, and layers \cite{yu2022width,lagunas2021block, zhou2021learning}. Unstructured pruning removes individual weights of the network with smaller magnitudes \cite{frantar2023sparsegpt, xia2023flash}. Albeit these are useful in reducing computational overhead, distillation and pruning requires either the model already trained as post-hoc method, architecture change than original models or iterative procedures that potentially take longer in an end-to-end manner. 

\textbf{Post-training Quantization. } Quantization compresses the representation of the parameters into low-precision data types, reducing the storage when loading the model in devices. Post-training quantization methods quantize the parameters of the pre-trained model \cite{yao2023zeroquant,xiao2023smoothquant, frantar2022gptq} often with fine-tuning steps \cite{kwon2022alphatuning}. \cite{jacob2018quantization} emulates inference-time quantization, creating a model that can be quantized later post-training . However, these works mostly focus on faster inference, rather than reducing end-to-end training time.

\paragraph{Kahan Summation.}\label{app-para:kahan} The Kahan summation is a standard algorithm in numerical analysis for accurate summation of floating-point numbers, just like the case of adding updates to the parameter. When incorporated with optimization algorithms such as SGD and AdamW, the Kahan algorithm introduces an auxiliary Kahan variable $\vec{c}$ (in the same precision) to track numerical errors at the parameter update step (i.e., $\vec{\theta}_{t} \gets \mathcal{F}^{P}(\vec{\theta}_{t-1} \oplus \vec{\Delta\theta}_{t})$) with $\vec{c}\gets \mathcal{F}^{P}
\left(\vec{\Delta\theta}_{t} \ominus \mathcal{F}^{P}(\vec{\theta}_{t} \ominus \vec{\theta}_{t-1})\right)$, and to compensate the addition results by adding $\vec{c}$ to the next iteration update: $\vec{\Delta\theta}_{t+1}\gets \mathcal{F}^{P}(\vec{\Delta\theta}_{t+1}\oplus\vec{c})$. The Kahan variable $\vec{c}$ accumulates lost small updates until it grows large enough to be added with the model parameters. As pointed in \cite{zamirai2020revisiting}, ``16-bit-FPU training with Kahan summation for model weight
updates have advantages in terms of throughput and memory consumption compared to 32-bit and mixed precision training", despite of the additional auxiliary value.

\paragraph{Stochastic Rounding.}\label{app-para:sr} Different from the deterministic rounding-to-the-nearest behavior, stochastic rounding rounds the number up and down in a probabilistic way. For any $x\in\mathbb{R}$, assume $a_u,a_l\in\mathbb{R}$ be the closest upper and lower neighboring floating-point values of $x$, i.e., $a_l\leq x<a_u=a_l+\ulp(a_l)$, then $\text{SR}(x)=a_l$ with probability $(a_u-x)/(a_u-a_l)=1-(x-a_l)/\ulp(a_l)$ and otherwise rounds up to $a_u$. Stochastic rounding provides an unbiased estimate of the precise value: $\expect{[\text{SR}(x)]} = x$ and alleviates the influence of imprecision by making the addition valid in expectation. Stochastic rounding for model weight updates adds minimal overhead for training and is supported in modern hardwares, such as AWS Trainium instances.

\begin{figure}[t]
\begin{minipage}[t]{0.3\textwidth}
\begin{algorithm}[H]
\caption{\textbf{TwoSum}}
\label{alg:twosum}
\begin{algorithmic}[1]
      \STATE {\bfseries Input:} $P$-bit floats $a$ and $b$
       \STATE $x\leftarrow\mathcal{F}^{P}(a\oplus b)$
       \STATE $b_{\text{virtual}}\leftarrow \mathcal{F}^{P}(x\ominus a)$
       \STATE $a_{\text{virtual}}\leftarrow \mathcal{F}^{P}(x\ominus b_{\text{virtual}})$
      \STATE $b_{\text{roundoff}}\leftarrow \mathcal{F}^{P}(b\ominus b_{\text{virtual}})$
      \STATE $a_{\text{roundoff}}\leftarrow \mathcal{F}^{P}(a\ominus a_{\text{virtual}})$
      \STATE $y\leftarrow \mathcal{F}^{P}(a_{\text{roundoff}}\oplus b_{\text{roundoff}})$
       \STATE {\bfseries Return:} $(x,y)$
    \end{algorithmic}
\end{algorithm}
\end{minipage}
\hfill
\begin{minipage}[t]{0.3\textwidth}
\begin{algorithm}[H]
\caption{\textbf{Split}}
\label{alg:split}
\begin{algorithmic}[1]
\STATE {\bfseries Input:} $P$-bit float $a$ (with $p$-bit mantissa)
\STATE $c\leftarrow \lfloor \frac{p}{2} \rfloor$
\STATE $t\leftarrow \mathcal{F}^{P}(2^{c} \oplus 1)\cdot a$
\STATE $a_{hi}\leftarrow \mathcal{F}^{P}(t \ominus \mathcal{F}^{P}(t\ominus a))$
\STATE $a_{lo}\leftarrow \mathcal{F}^{P}(a \ominus a_{hi})$
\STATE {\bfseries Return:} $(a_{hi}, a_{lo})$
\end{algorithmic}
\end{algorithm}
\end{minipage}
\hfill
\begin{minipage}[t]{0.34\textwidth}
\begin{algorithm}[H]
\caption{\textbf{TwoProd}}
\label{alg:twoprod}
\begin{algorithmic}[1]
\STATE {\bfseries Input:} $P$-bit floats $a$ and $b$
\STATE $x\leftarrow \mathcal{F}^{P}(a \odot b)$
\STATE $(a_{hi}, a_{lo})\leftarrow\textbf{Split}(a)$
\STATE $(b_{hi}, b_{lo})\leftarrow\textbf{Split}(b)$
\STATE $err_1\leftarrow \mathcal{F}^{P}(p \ominus  \mathcal{F}^{P}(a_{hi}\odot b_{hi}))$
\STATE $err_2\leftarrow \mathcal{F}^{P}(err_1 \ominus \mathcal{F}^{P}(a_{lo}\odot b_{hi}))$
\STATE $err_3\leftarrow \mathcal{F}^{P}(err_2 \ominus \mathcal{F}^{P}(a_{hi}\odot b_{lo}))$
\STATE $e\leftarrow \mathcal{F}^{P}(\mathcal{F}^{P}(a_{lo}\odot b_{lo}) \ominus err_3)$
\STATE {\bfseries Return:} $(x,e)$
\end{algorithmic}
\end{algorithm}
\end{minipage}
\end{figure}

\begin{figure}[t]
\begin{minipage}[t]{0.32\textwidth}
\begin{algorithm}[H]
    \caption{\textbf{TwoProdFMA}}
    \label{alg:twoprodfma}
    \begin{algorithmic}[1]
      \STATE {\bfseries Input:} $P$-bit floats $a$ and $b$
      \STATE {\bfseries Requires:} Machine supports FMA
      \STATE $x\leftarrow \mathcal{F}^{P}(a \odot b)$
      \STATE $e\leftarrow \mathcal{F}^{P}(a \odot b \ominus x)$ \text{in FMA}
       \STATE {\bfseries Return:} $(x,e)$
    \end{algorithmic}
\end{algorithm}
\end{minipage}
\hfill
\begin{minipage}[t]{0.3\textwidth}
\begin{algorithm}[H]
   \caption{\textbf{Scaling}}
   \label{alg:scaling}
    \begin{algorithmic}[1]
    \STATE {\bfseries Input:} a float $v$ and a length-$2$ expansion $(a_1, a_2)$
   \STATE $(x, e) \leftarrow \textbf{TwoProdFMA}(a_1, v)$
   \STATE $e \leftarrow \mathcal{F}^{P}(a_2\odot v \oplus e)$
   \STATE $(x, e) \leftarrow \textbf{Fast2Sum}(x, e)$
   \STATE {\bfseries Return:} $(x,e)$
\end{algorithmic}
\end{algorithm}
\end{minipage}
\hfill
\begin{minipage}[t]{0.36\textwidth}
\begin{algorithm}[H]
\caption{\textbf{Mul}}
\label{alg:multmcn}
\begin{algorithmic}[1]
    \STATE {\bfseries Input:} length-$2$ expansions $(a_1, a_2)$ and $(b_1, b_2)$
    \STATE $(x, e) \leftarrow \textbf{TwoProdFMA}(a_1, b_1)$
    \STATE $e \leftarrow \mathcal{F}^{P}\left(e\oplus( (a_1\odot b_2) \oplus (a_2\odot b_1))\right)$
    \STATE $(x, e) \leftarrow \textbf{Fast2Sum}(x, e)$
    \STATE {\bfseries Return:} $(x,e)$
\end{algorithmic}
\end{algorithm}
\end{minipage}
\end{figure}
\section{MCF algorithms}
\label{appsec:mcf_algs}
As noted in Theorem~\ref{thm:fast2sum}, \pythoninline{Fast2Sum} requires $|a|>|b|$ so as to perform the arithmetic correctly. One can also derive the same length-$2$ expansion using \pythoninline{Two-Sum} in Algorithm~\ref{alg:twosum} for any floats $a, b$ without sorting.

Another category of basic MCF algorithms is the multiplication, between i) a float and a float (with \pythoninline{TwoProd} Algorithm~\ref{alg:twoprod}); ii) a float and a length-$2$ expansion (with \pythoninline{Scaling} Algorithm~\ref{alg:scaling}); iii) an expansion and an expansion (with \pythoninline{Mul} Algorithm~\ref{alg:multmcn}), to produce length-$2$ expansions. 

\paragraph{Case i).} \pythoninline{TwoProd} computes the expansion using another basic algorithm \pythoninline{Split} (Algorithm~\ref{alg:split}), which takes a single $P$-bit floating point value and splits it into its high and low parts, both with $\frac{P}{2}$ bits. On a machine which supports the fused-multiply-add (FMA) instruction set, a much more efficient version \pythoninline{TwoProdFMA} Algorithm~\ref{alg:twoprodfma} can be adopted to give the same results. We utilized this efficient \pythoninline{TwoProdFMA} in our implementations as (Bfloat16) FMA is supported on CUDA, e.g., using \pythoninline{torch.addcmul}($-x, a, b$).

\paragraph{Case ii) and iii).} Algorithm~\ref{alg:scaling}~\pythoninline{Scaling} describes the multiplication of a single float with a length-$2$ expansion and Algorithm~\ref{alg:multmcn}~\pythoninline{Mul} the multiplication between $2$ length-$2$ expansions. With FMA enabled, both algorithms run efficiently.

We refer the readers to \cite{yu2022mctensor} for a full list of MCF algorithms.

\section{Further Discussions on Algorithms}
\label{appsec:further_discussions_algorithm}

\paragraph{Equivalence.} The equivalence of using `Kahan-sum in the optimizer' at the model-update step and \strname-light is straightforward, realizing i) the Kahan variable $\vec{c}$ calculation is essentially \pythoninline{Fast2Sum} given $|\vec{\theta}_t|\geq |\Delta\vec{\theta}_{t-1}|$; and ii) next iteration update $\vec{\Delta\theta}_{t+1}$ has similar magnitude as $\vec{c}$ so that lost arithmetic doesn't happen. In contrast, \strname-light doesn't have such concerns using
\pythoninline{Grow}.

\paragraph{Weight Decay.}
\cite{loshchilov2017decoupled} propose AdamW with the decoupled weight decay placed at line 12 in Algorithm~\ref{alg:mcfadamw} for a summed update $\vec{\Delta\theta}_t$, standard libraries including PyTorch and HuggingFace however implement the decoupled weight decay directly to the parameter:
\begin{equation}
\label{eq:ineffective-wd}
\vec{\theta}_t\leftarrow \vec{\theta}_{t-1} - \alpha\lambda\vec{\theta}_{t-1},~~~~\text{or}~~~~\vec{\theta}_t\leftarrow (1 - \alpha\lambda)\vec{\theta}_{t-1}    
\end{equation}
which works as expected using Float$32$, but is usually ineffective in Bfloat$16$ arithmetic due to imprecision and lost arithmetic. For example, a standard choice of the learning rate and weight decay hyper-parameter in GPT-$6.7$B pretraining is $\alpha=1.2e-4$ and $\lambda=0.1$, yielding $\alpha\lambda=1.2e-5$ and causing lost arithmetic in Equation~\ref{eq:ineffective-wd} when Bfloat$16$ is used. In fact, the least $\alpha\lambda$ value to avoid invalid arithmetic is half $\text{ulp}(1.0)$, i.e., $2^{-7}/2\approx 0.0039$. Either decaying the parameter (expansion) with \pythoninline{Grow} or placing the decoupled weight decay term at line 13 following the original AdamW algorithm statement solves the issue, where we chose the latter option in our experiments.

\paragraph{Scalar and Bias Correction.} A rule of thumb to avoid imprecision and lost arithmetic during low precision training is to do as many scalar computations in high precision as possible before casting them to low precision (e.g., PyTorch BFloat$16$ Tensor). For example, in BFloat$16$ AdamW, it's recommended to compute the bias correction scalar terms $1-\beta_1^t$ and $1-\beta_2^t$ in high precision before dividing the low precision momentums. 

\section{Experiments details}
\label{appsec:experi_details}

\makeatletter
    \setlength\@fptop{0\p@}
    \setlength{\@fpsep}{4ex}
\makeatother

\subsection{BERT and RoBERTa}
\label{appssec:bert_roberta}
We pretrain the BERT-base-uncased, BERT-large-uncased and RoBERTa-base model from HuggingFace \cite{wolf2019huggingface} on the Wikipedia-en corpus \cite{Wikiextractor2015}, preprocessed with BERT tokenizer. We follow the standard pipeline to pretrain BERT and RoBERTa with the same configs and hyper-parameters for all precision strategies.
Note that we took these configs and hyper-parameters from open-sourced models in HuggingFace.
We finetuned the pretrained BERT and RoBERTa models following \cite{wang2019glue} with BF$16$ mixed precision through HuggingFace and evaluated the final model on GLUE benchmarks. Particularly, we used $2e-5$ learning rate and a batch size of $32$ evaluated on single Nvidia A100. All tasks were finetuned for $3$ epochs, apart from MRPC which we ran for $5$ epochs.

\subsection{Multi-size GPTs \& OpenLLaMA-$7$B}
\label{appssec:gpt_llama}
We conduct pretraining experiments of 1) GPT at different sizes ranging from $125$M, $1.3$B, $2.7$B to $6.7$B, and 2) OpenLLaMA-7B using NeMo Megatron \cite{kuchaiev2019nemo} with provided standard configs, both on the Wikipedia corpus with HuggingFace GPT2 and LLaMA tokenizer, respectively. We split the dataset into train/val/test with the split ratio $980:10:10$. We trained all models consistently with disabled sequence parallelism, enabled flash attention, rotary positional embedding (of percentage $1.0$)~\cite{su2024roformer}, disabled transformer engine, untied embeddings \& output weights, sequence length of $2,048$, weight decay $0.1$ and pipeline parallelism $1$, for all GPT models and OpenLLaMA-$7$B in our experiments unless otherwise specified. All models were trained with the CosineAnnealing learning rate scheduler with $200$ warmup iterations. We trained all GPT models for $20$k iterations and OpenLLaMA for $9$k iterations due to timing constraints. The dafault value of $\beta$s are $\beta_1=0.9$ and $\beta_2=0.95$ unless otherwise specified, e.g., in ablation experiments. Note that we took these configs from \hyperlink{https://github.com/EleutherAI/gpt-neox/tree/main/configs}{EleutherAI/gpt-neox} \cite{gpt-neox-library}.

\begin{table}
 \centering
 \caption{Pre-training hyperparameters used for BERT and RoBERTa.}
    \begin{tabular}{lclcc} 
    \toprule
    Model & Phase & hyperparameters & Values\\\hline
    \multirow{12}{*}{BERT-base}  & \multirow{6}{*}{Phase-1} & iterations & $28,125$\\
    && warmup steps & $2,000$\\
    && sequence length & $128$ \\
    && global batch size & $16,384$\\
    && learning rate & $4e-4$\\
    && $(\beta_1, \beta_2)$ & $(0.9, 0.999)$\\
    \cmidrule{2-4}
    & \multirow{6}{*}{Phase-2} & iterations & $28,125$\\
    && warmup steps & $2,000$\\
    && sequence length & $512$\\
    && global batch size & $32,768$\\
    && learning rate & $2.8e-4$\\
    && $(\beta_1, \beta_2)$ & $(0.9, 0.999)$\\
    \hline
    \multirow{6}{*}{RoBERTa-base}  & \multirow{6}{*}{Phase-1} & iterations & $28,125$ \\
    && warmup steps & $2,000$ \\
    && sequence length & $512$ \\
    && global batch size & $8,192$\\
    && learning rate & $lr=6e-4$ \\
    && $(\beta_1, \beta_2)$ & $(0.9, 0.98)$\\
    \bottomrule
    \end{tabular}
    \label{tab:bert_roberta_configs}
\end{table}
\begin{table}[ht]
 \centering
 \vspace{-2em}
 \caption{\centering Some configs and hyper-parameters of GPT models and OpenLLaMA-$7$B.
    }
    \begin{tabular}{l|cccccc} 
    \toprule
      Model & \#Layers & HiddenSize & \#AttentionHeads & Global BatchSize & TensorParallelism & $lr$ \\
    \midrule
    GPT-$125$M & $12$ & $768$ & $12$ & $1,024$ & $1$ & $6e-4$ \\
    GPT-$1.3$B & $24$ & $2048$ & $16$ & $1,024$ & $8$ & $2e-4$ \\
    GPT-$2.7$B & $32$ & $2560$ & $32$ & $512$ & $8$ & $1.6e-4$  \\
    GPT-$6.7$B & $32$ & $4096$ & $32$ & $256$ & $8$ & $1.2e-4$ \\
    OpenLLaMA-$7$B & $32$ & $4096$ & $32$ & $256$ & $8$ & $3e-4$ \\
    \bottomrule
    \end{tabular}
    \label{tab:gpt-configs}
    \vspace{-1em}
\end{table}
\paragraph{GPT-$30$B.} For the GPT-$30$B model used in Section~\ref{subsec:efficiency_memory}, it has $56$ layers, hidden size $7168$ and $56$ attention heads. We trained it with a global batchsize $256$, tensor parallelism $8$ and pipeline parallelism $2$, then varied the micro batchsize and sequence length to explore their maxium values without causing $\oom$ on a NVIDIA A100 cluster with $2$ nodes, $8$ GPUs each.  

\section{Additional Results}
\subsection{Memory Statistics}
\label{appssec:memory_stats}
Table~\ref{tab:memory} summarizes the peak (total) memory of all training precision strategies during practical runs with the same hyper-parameters for a fair comparison: Sequence Length $2048$, Global BatchSize $128$ and Micro (per-device) BatchSize $1$. We trained GPTs and OpenLLaMA with TensorParallelism $8$ over $8\times$A$100$ $40$GB, except from GPT-$125$M with TensorParallelism $1$ on $1\times$A$100$ $40$GB. In Table~\ref{tab:memory}, we report the saved memory compared to the mixed-precision option D with the percentage calculated. During real runs, on average, \strname formations (light/plus) use \bm{$23.8\%/15.6\%$} less peak memory compared to option D. The best savings are for largest model OpenLLaMA-$7$B, with savings \bm{$27.8\%/18.5\%$}, respectively. The memory savings match the theoretical calculation in Table~\ref{tab:precision-strategies-breakdown}. 

\begin{table}[htb]
\centering
\caption{\centering Peak (saved) pretraining memory (GB) of precision strategies compared to option D on GPTs and OpenLLaMA-$7$B.}
\label{tab:memory}
\begin{tabular}{l|ccccc}
\toprule
Precision & \multicolumn{4}{c}{GPT} & \multirow{2}{*}{\makecell{OpenLLaMA \\ $7$B}}  \\
Option & $125$M & $1.3$B & $2.7$B & $6.7$B & \\
\hline
A & $-1.1(-26.6\%)$ & $-10.3(-28.9\%)$ & $-20.8(-31.2\%)$ & $-51.2(-35.6\%)$ & $-65.7(-37.2\%)$ \\
B (ours) & $-0.8(-19.3\%)$ & $-7.6(-21.5\%)$ & $-15.6(-23.8\%)$ & $-38.2(-26.6\%)$ & $-49.2(-27.8\%)$\\
C (ours) & $-0.5(-12.1\%)$ & $-5.0(-14.1\%)$ & $-10.1(-15.4\%)$ & $-25.7(-17.9\%)$ & $-32.7(-18.5\%)$\\
\hline
D & $4.4$ & $35.5$ & $65.3$ & $143.7$  & $176.7$ \\
\bottomrule
\end{tabular}
\end{table}


\subsection{OpenLLama 7B pretraining}
\label{appssec:openllama7B_pt}
\begin{figure}[htb]
\centering
\includegraphics[width=\linewidth]{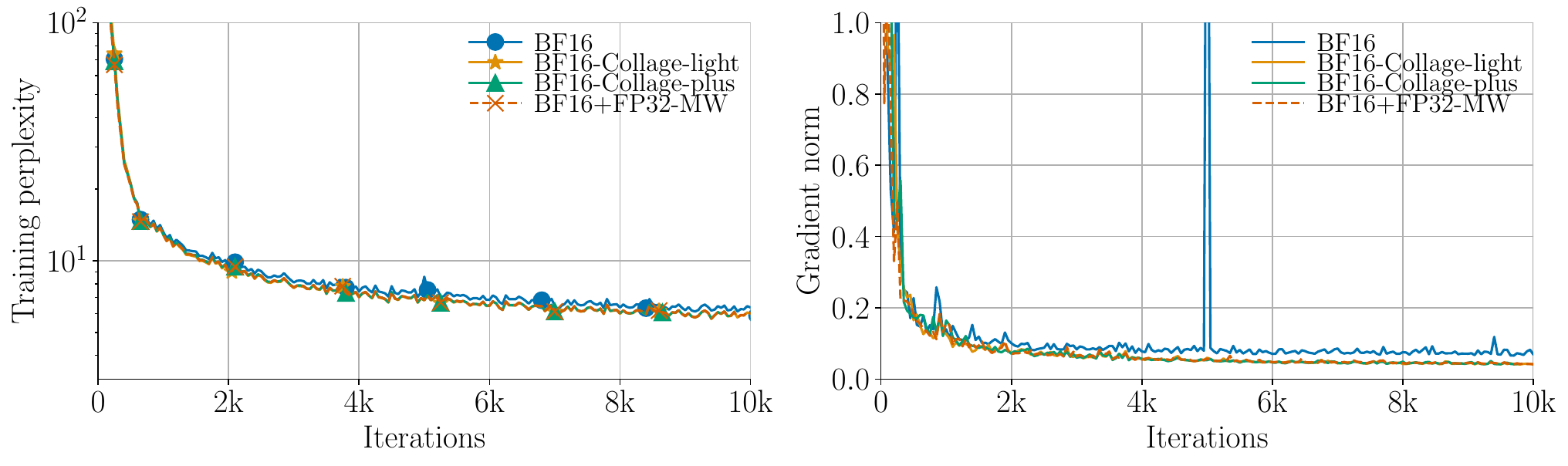}
\vspace{-1em}
\caption{Openllama 7B pretraining (see settings in Section\,\ref{ssec:llama_gpt_results}) with $\beta_2=0.95$. \textbf{Left:} Training perplexity for different precision strategies listed in Table\,\ref{tab:precision-strategies-breakdown}. \textbf{Right:} Model gradient L2 norm across iterations for different strategies. The \strname formations overlap with heavy-weighted FP32 master weights strategy.}
\label{fig:openllama-7B-beta2_0p95}
\end{figure}

\begin{figure}
\centering
\includegraphics[width=\linewidth]{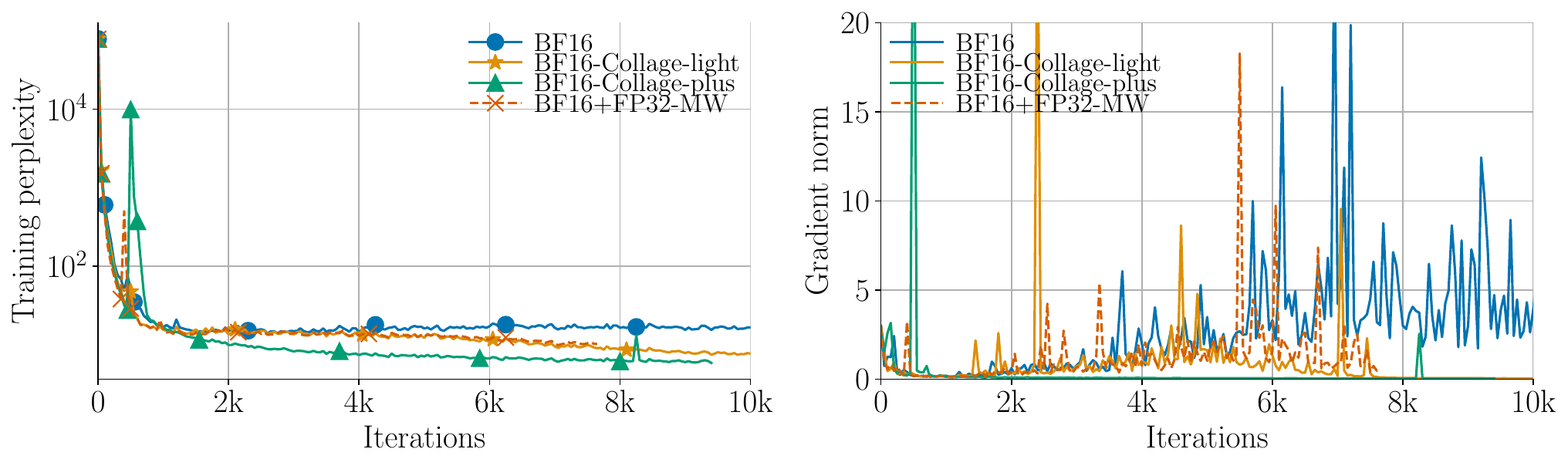}
\vspace{-1em}
\caption{Openllama 7B pretraining (see settings in Section\,\ref{ssec:llama_gpt_results}) with $\beta_2=0.99$. \textbf{Left:} Training perplexity for different precision strategies listed in Table\,\ref{tab:precision-strategies-breakdown}. \textbf{Right:} Model gradient L2 norm across iterations for different strategies. The \strname-plus results in the best train perplexity over iterations while other approaches struggle. The gradient norm blows-up frequently but stays stable for \strname-plus which suggest importance of using multi-components at critical locations.}
\label{fig:openllama-7B-beta2_0p99}
\end{figure}

We provide the pretraining iterations progress for OpenLLama-$7$B (described in Section\,\ref{ssec:llama_gpt_results}) in the \figurename\,\ref{fig:openllama-7B-beta2_0p95}, \ref{fig:openllama-7B-beta2_0p99} for $\beta_2=0.95, 0.99$, respectively. We observe a stable training using \strname-plus when using $\beta_2=0.99$, where other precision strategies show slow convergence. The gradient norm in \figurename\,\ref{fig:openllama-7B-beta2_0p99}\,left show that \strname-plus has stability while other precision strategies encounter transient errors causing blow-ups in gradients.

\subsection{GPT pretraining}
\label{appssec:gpt_pt}
\begin{figure}[htbp]
\centering
\begin{minipage}[b]{0.33\textwidth}
    \centering
    \includegraphics[width=1.02\linewidth]{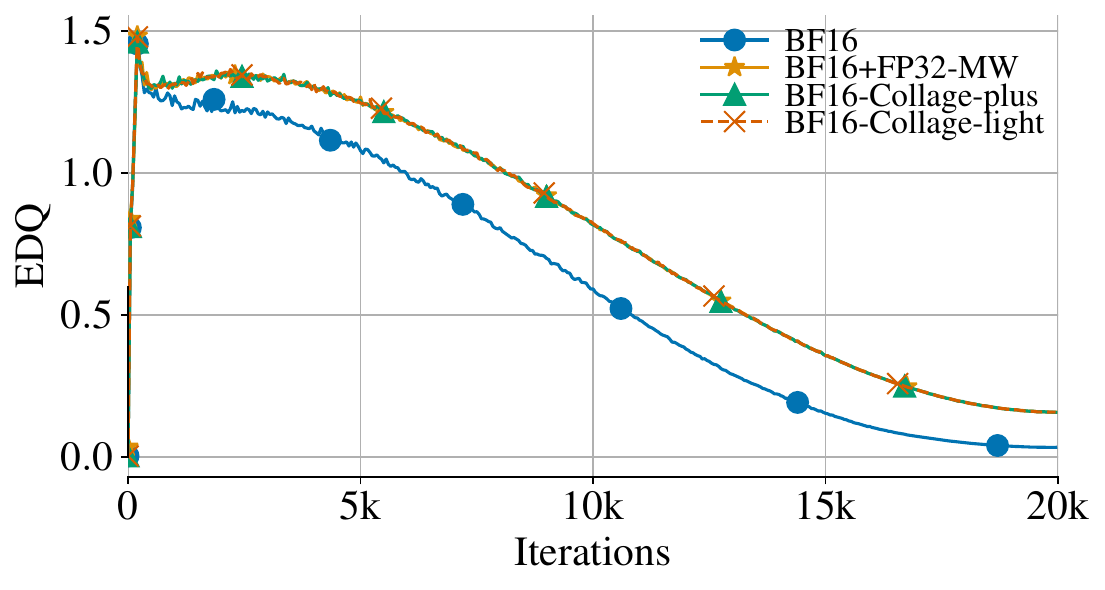}
\end{minipage}
\begin{minipage}[b]{0.33\textwidth}
    \centering
    \includegraphics[width=1.02\linewidth]{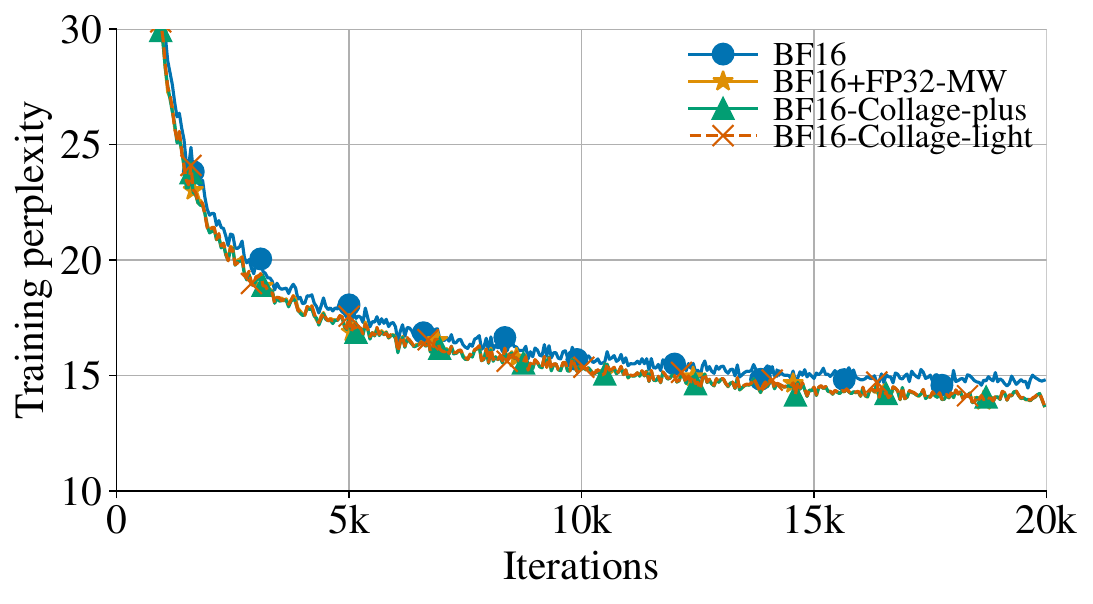}
\end{minipage}
\begin{minipage}[b]{0.33\textwidth}
    \centering
    \includegraphics[width=1.02\linewidth]{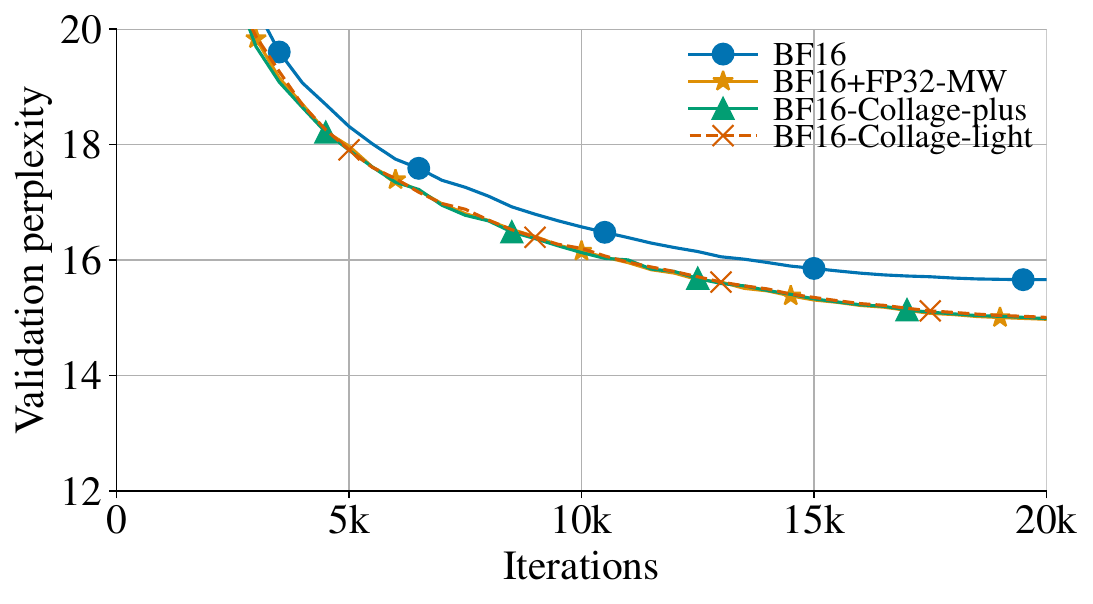}
\end{minipage}
\caption{Pretrainnig progress for GPT $125$M with settings described in Section\,\ref{ssec:llama_gpt_results} and global batch-size=1024, $\beta_2=0.95$. \textbf{Top-left:} EDQ metric vs iterations, \textbf{top-right:} training perplexity vs iterations, and \textbf{bottom:} validation perplexity vs iterations for different precision strategy listed in Table\,\ref{tab:precision-strategies-breakdown}. The proposed \strname formations consistently match the FP32 master weights with much less memory footprint and faster training.}
    \label{fig:gpt125M_beta2_0p95_GBS1024}
\end{figure}

\begin{figure}[htbp]
\centering
\begin{minipage}[b]{0.33\textwidth}
    \centering
    \includegraphics[width=1.02\linewidth]{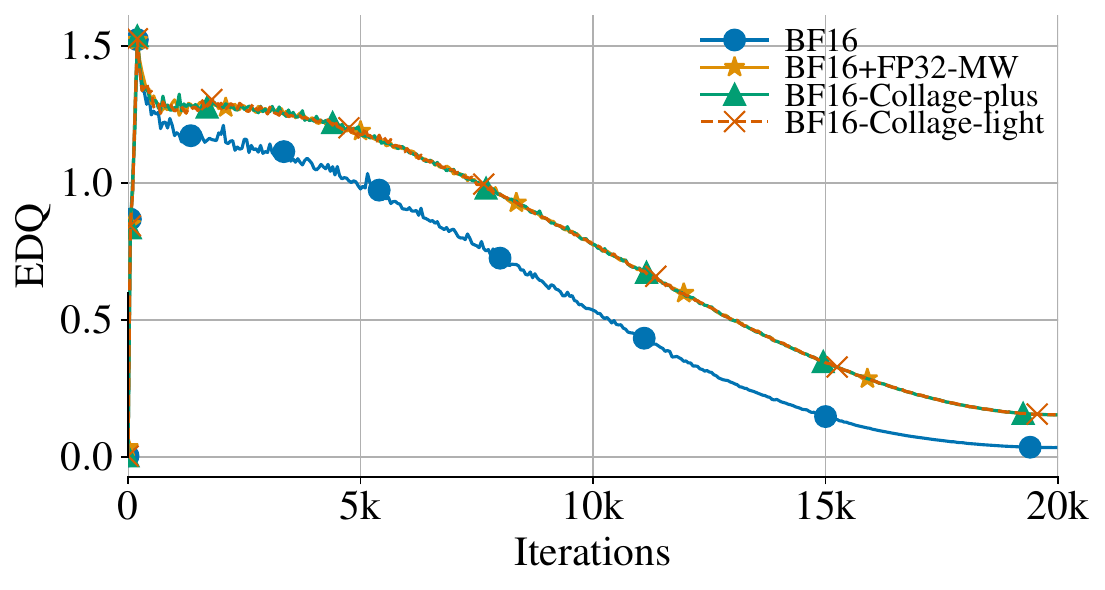}
\end{minipage}
\begin{minipage}[b]{0.33\textwidth}
    \centering
    \includegraphics[width=1.02\linewidth]{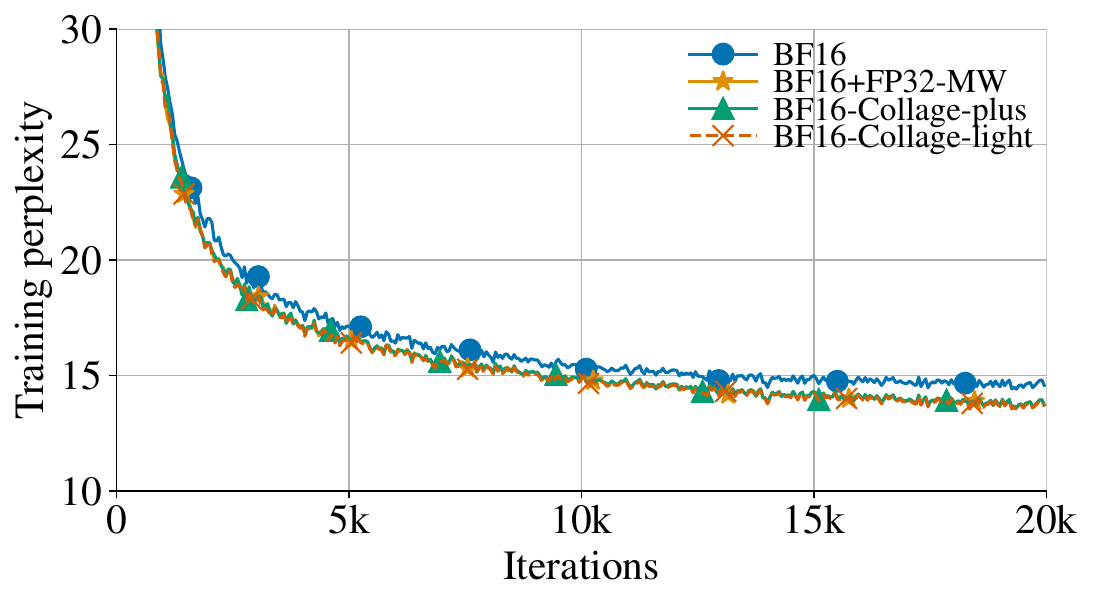}
\end{minipage}
\begin{minipage}[b]{0.33\textwidth}
    \centering
    \includegraphics[width=1.02\linewidth]{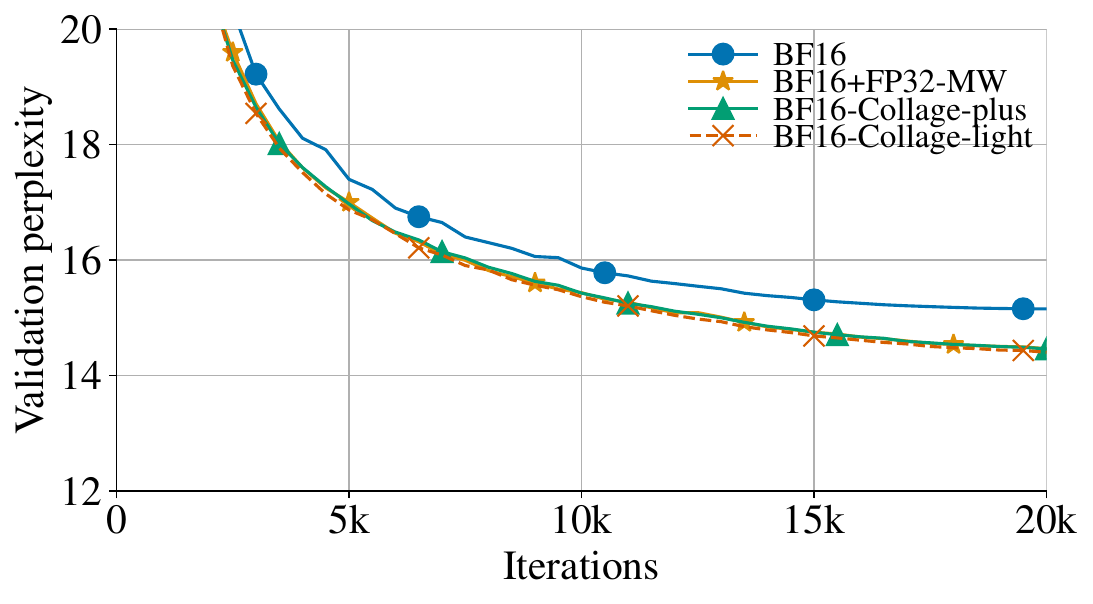}
\end{minipage}
\caption{Pretrainnig progress for GPT $125$M with settings described in Section\,\ref{ssec:llama_gpt_results} and global batch-size=2048, $\beta_2=0.95$. \textbf{Top-left:} EDQ metric vs iterations, \textbf{top-right:} training perplexity vs iterations, and \textbf{bottom:} validation perplexity vs iterations for different precision strategy listed in Table\,\ref{tab:precision-strategies-breakdown}. The proposed \strname formations consistently match the FP32 master weights with much less memory footprint and faster training.}
    \label{fig:gpt125M_beta2_0p95_GBS2048}
\end{figure}

\begin{figure}[htbp]
\centering
\begin{minipage}[b]{0.33\textwidth}
    \centering
    \includegraphics[width=1.02\linewidth]{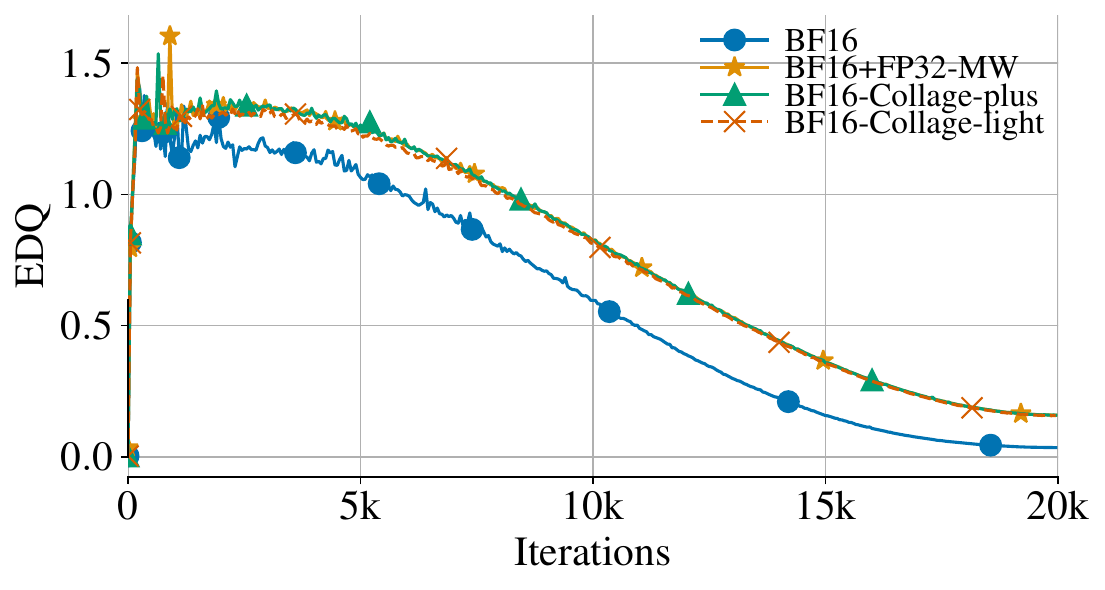}
\end{minipage}
\begin{minipage}[b]{0.33\textwidth}
    \centering
    \includegraphics[width=1.02\linewidth]{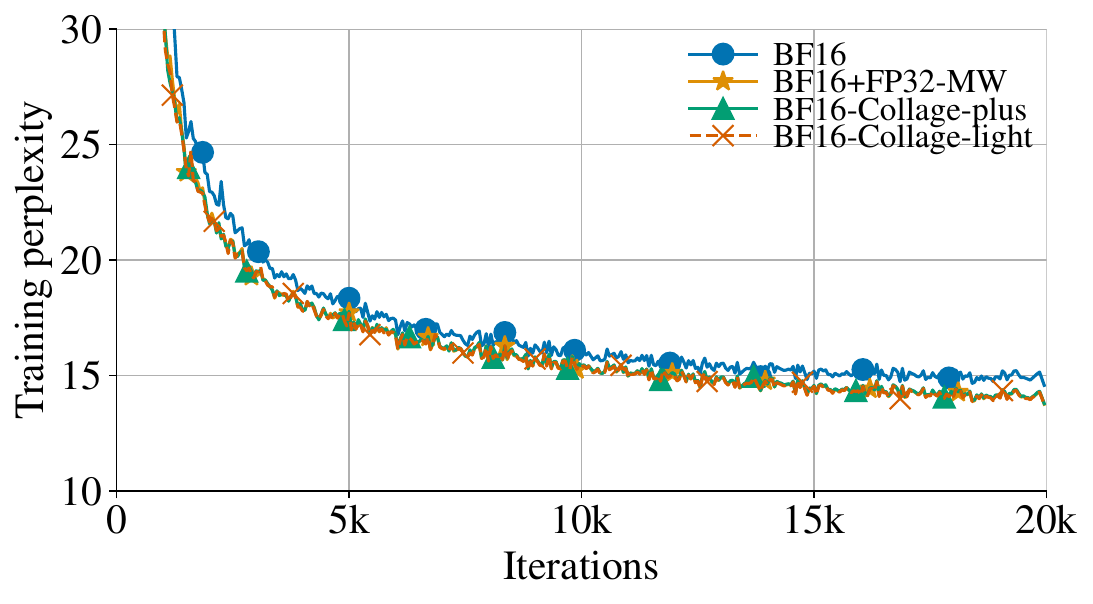}
\end{minipage}
\begin{minipage}[b]{0.33\textwidth}
    \centering
    \includegraphics[width=1.02\linewidth]{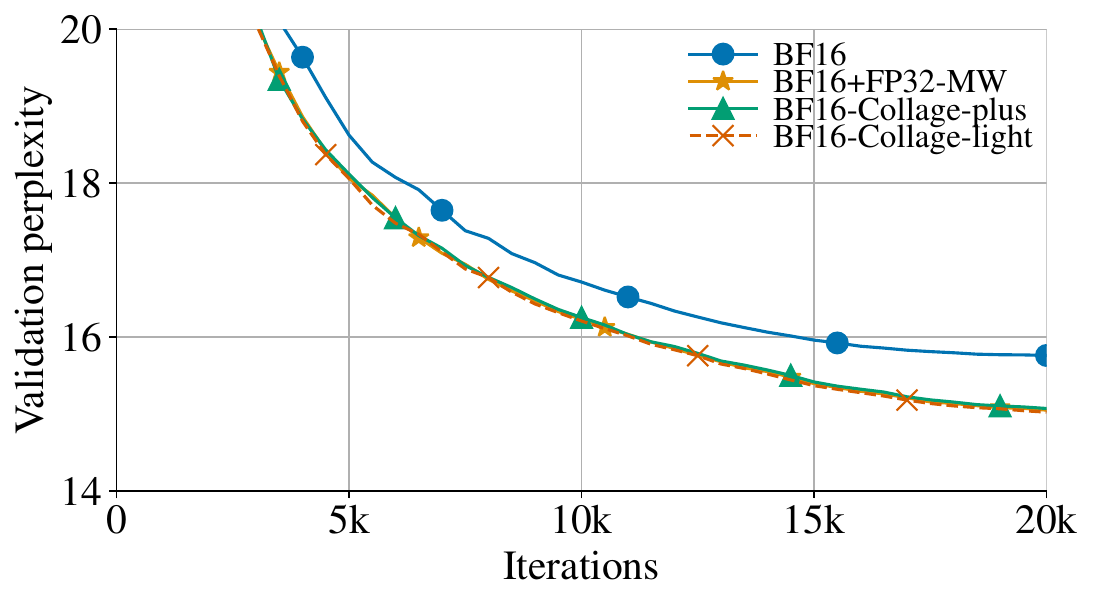}
\end{minipage}
\caption{Pretrainnig progress for GPT $125$M with settings described in Section\,\ref{ssec:llama_gpt_results} and global batch-size=1024, $\beta_2=0.99$. \textbf{Top-left:} EDQ metric vs iterations, \textbf{top-right:} training perplexity vs iterations, and \textbf{bottom:} validation perplexity vs iterations for different precision strategy listed in Table\,\ref{tab:precision-strategies-breakdown}. The proposed \strname formations consistently match the FP32 master weights with much less memory footprint and faster training.}
    \label{fig:gpt125M_beta2_0p99_GBS1024}
\end{figure}

\begin{figure}[htbp]
\centering
\begin{minipage}[b]{0.33\textwidth}
    \centering
    \includegraphics[width=1.02\linewidth]{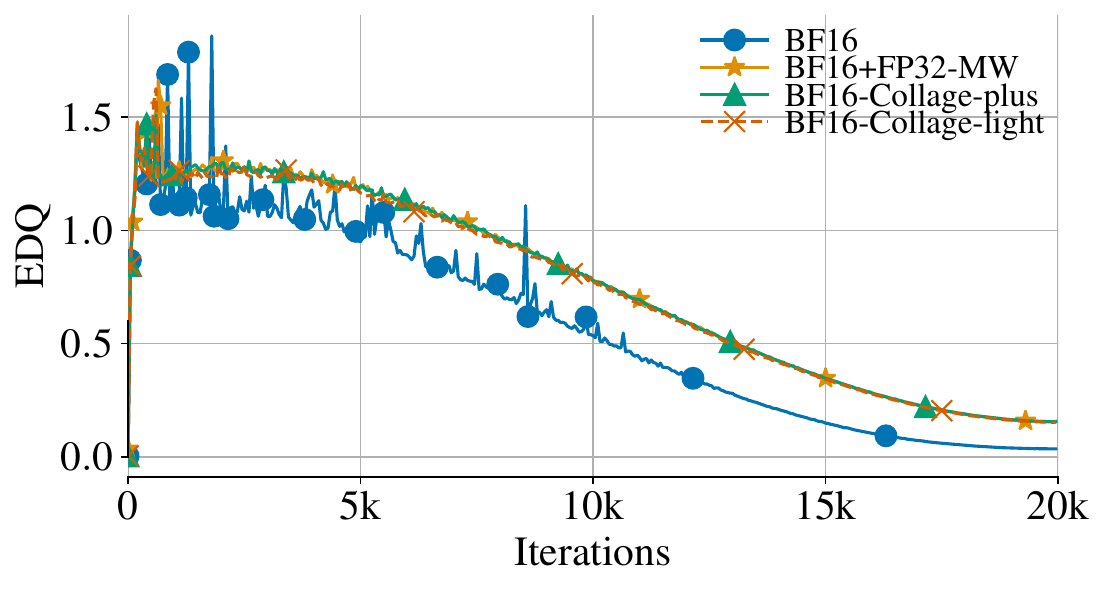}
\end{minipage}
\begin{minipage}[b]{0.33\textwidth}
    \centering
    \includegraphics[width=1.02\linewidth]{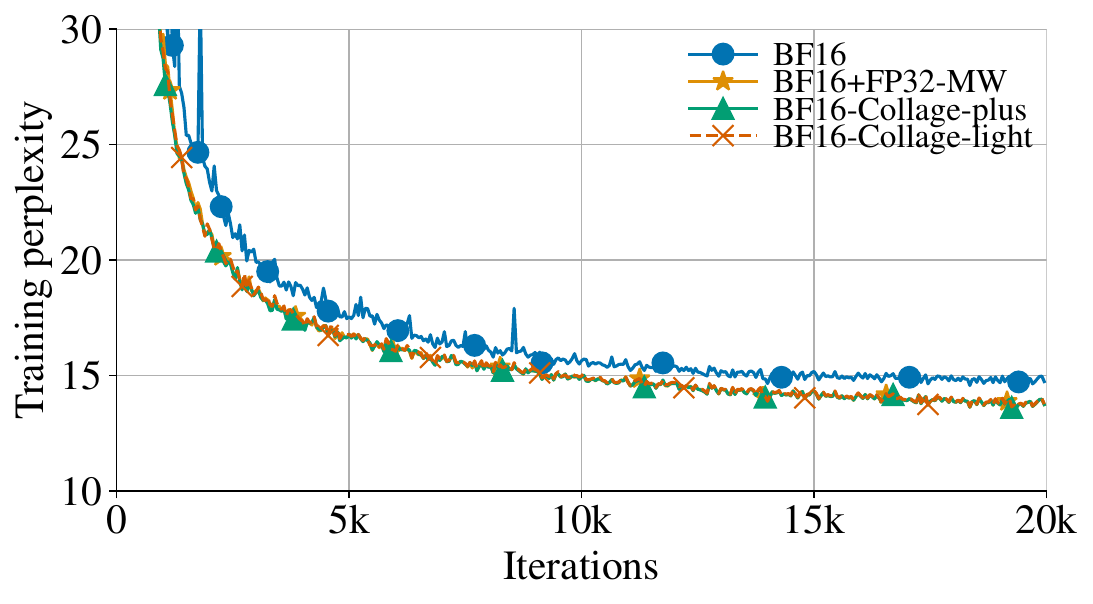}
\end{minipage}
\begin{minipage}[b]{0.33\textwidth}
    \centering
    \includegraphics[width=1.02\linewidth]{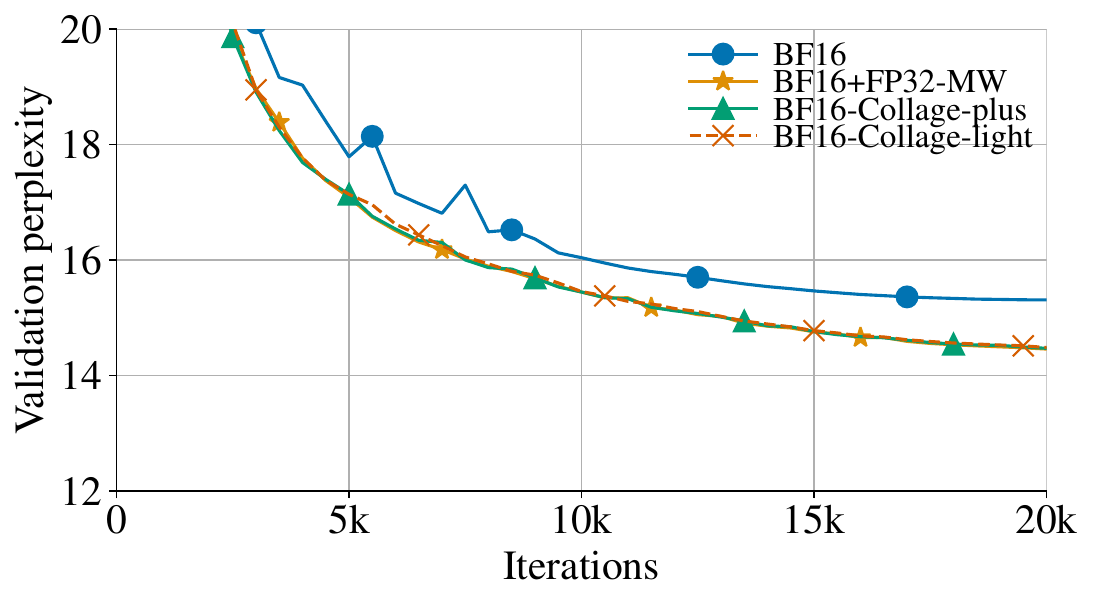}
\end{minipage}
\caption{Pretrainnig progress for GPT $125$M with settings described in Section\,\ref{ssec:llama_gpt_results} and global batch-size=2048, $\beta_2=0.99$. \textbf{Top-left:} EDQ metric vs iterations, \textbf{top-right:} training perplexity vs iterations, and \textbf{bottom:} validation perplexity vs iterations for different precision strategy listed in Table\,\ref{tab:precision-strategies-breakdown}. The proposed \strname formations consistently match the FP32 master weights with much less memory footprint and faster training.}
    \label{fig:gpt125M_beta2_0p99_GBS2048}
\end{figure}

\begin{figure}[htbp]
\centering
\begin{minipage}[b]{0.33\textwidth}
    \centering
    \includegraphics[width=1.02\linewidth]{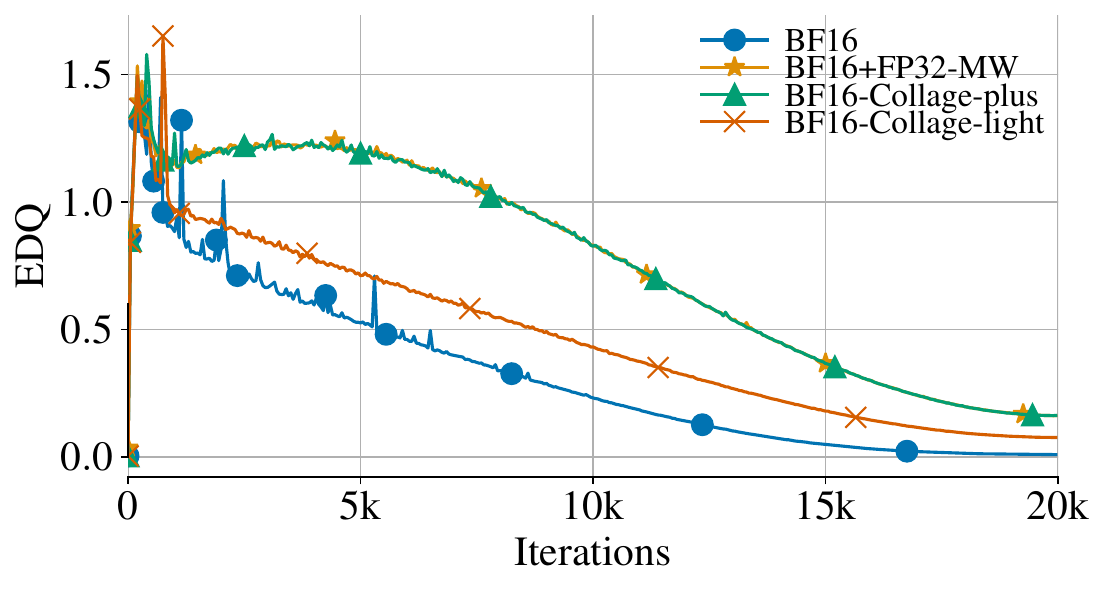}
\end{minipage}
\begin{minipage}[b]{0.33\textwidth}
    \centering
    \includegraphics[width=1.02\linewidth]{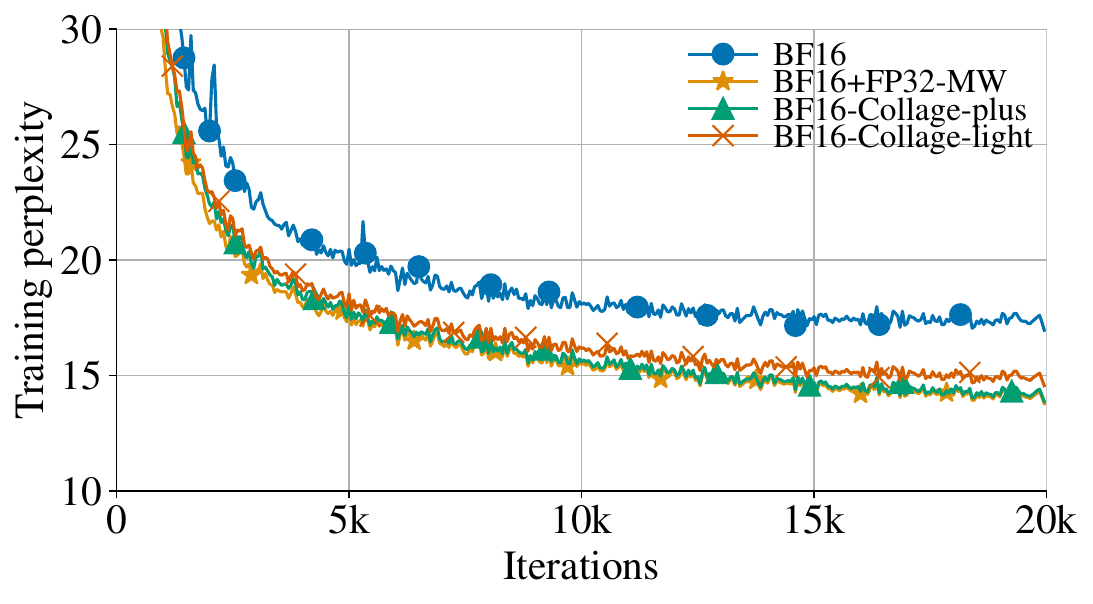}
\end{minipage}
\begin{minipage}[b]{0.33\textwidth}
    \centering
    \includegraphics[width=1.02\linewidth]{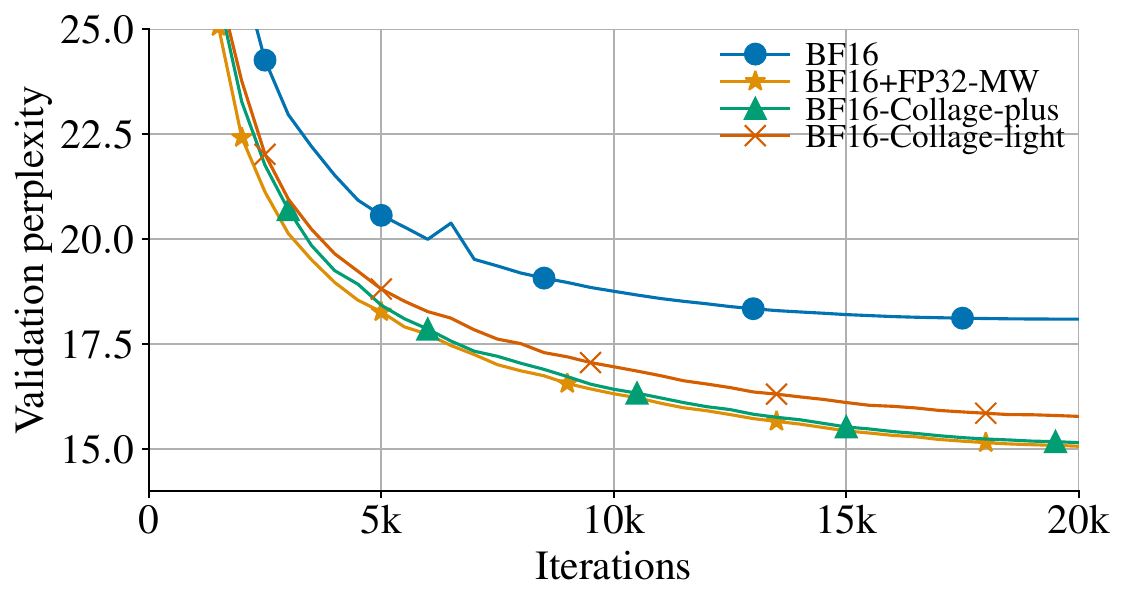}
\end{minipage}
\caption{Pretrainnig progress for GPT $125$M with settings described in Section\,\ref{ssec:llama_gpt_results} and global batch-size=1024, $\beta_2=0.999$. \textbf{Top-left:} EDQ metric vs iterations, \textbf{top-right:} training perplexity vs iterations, and \textbf{bottom:} validation perplexity vs iterations for different precision strategy listed in Table\,\ref{tab:precision-strategies-breakdown}. The proposed \strname formations consistently match the FP32 master weights with much less memory footprint and faster training.}
    \label{fig:gpt125M_beta2_0p999_GBS1024}
\end{figure}

\begin{figure}[htbp]
\centering
\begin{minipage}[b]{0.33\textwidth}
    \centering
    \includegraphics[width=1.02\linewidth]{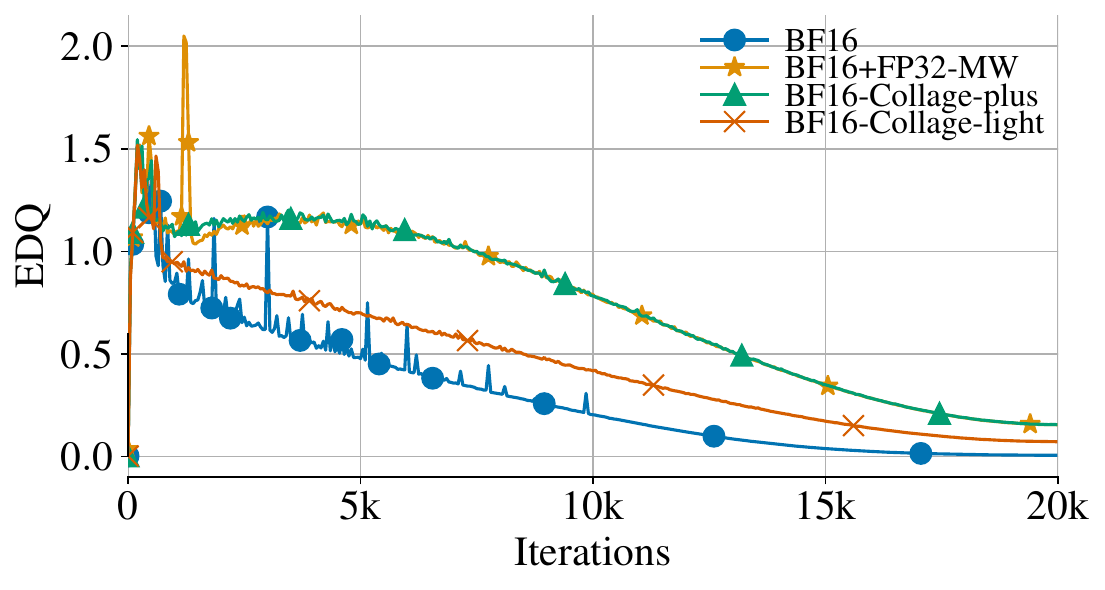}
\end{minipage}
\begin{minipage}[b]{0.33\textwidth}
    \centering
    \includegraphics[width=1.02\linewidth]{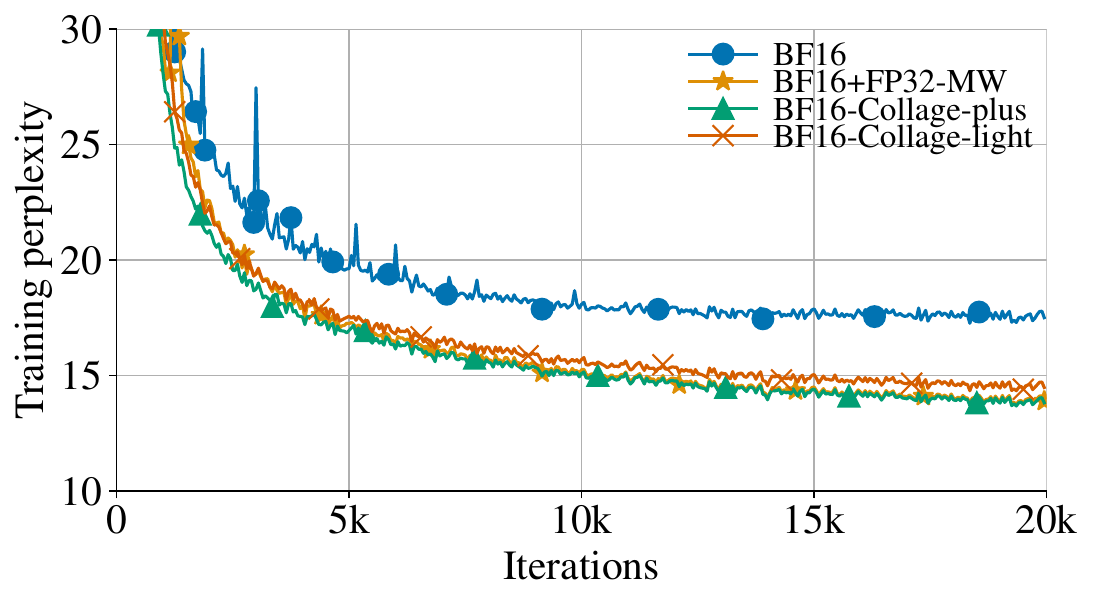}
\end{minipage}
\begin{minipage}[b]{0.33\textwidth}
    \centering
    \includegraphics[width=1.02\linewidth]{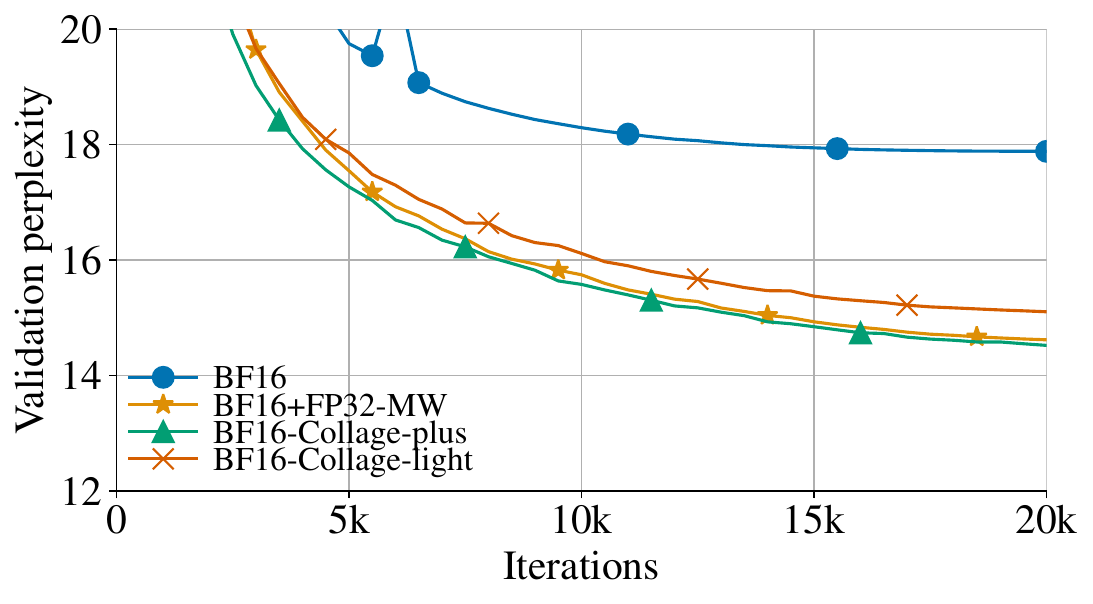}
\end{minipage}
\caption{Pretrainnig progress for GPT $125$M with settings described in Section\,\ref{ssec:llama_gpt_results} and global batch-size=2048, $\beta_2=0.999$. \textbf{Top-left:} EDQ metric vs iterations, \textbf{top-right:} training perplexity vs iterations, and \textbf{bottom:} validation perplexity vs iterations for different precision strategy listed in Table\,\ref{tab:precision-strategies-breakdown}. The proposed \strname formations consistently match the FP32 master weights with much less memory footprint and faster training.}
    \label{fig:gpt125M_beta2_0p999_GBS2048}
\end{figure}

\begin{figure}
    \centering
    \subfigure{
    \includegraphics[width=0.48\linewidth]{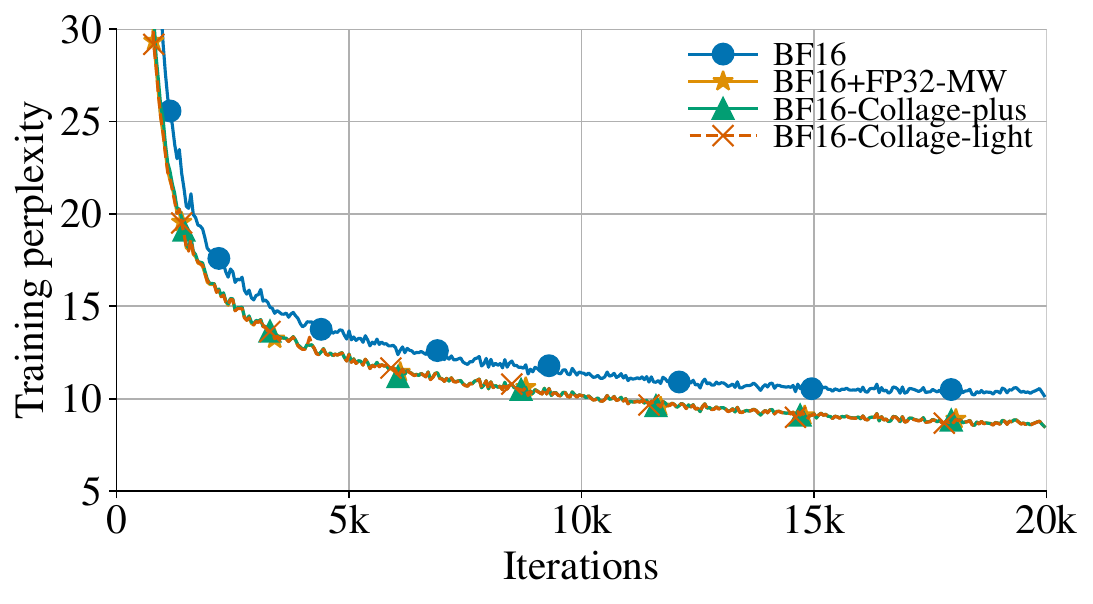}
    }
    ~
    \subfigure{
    \includegraphics[width=0.48\linewidth]{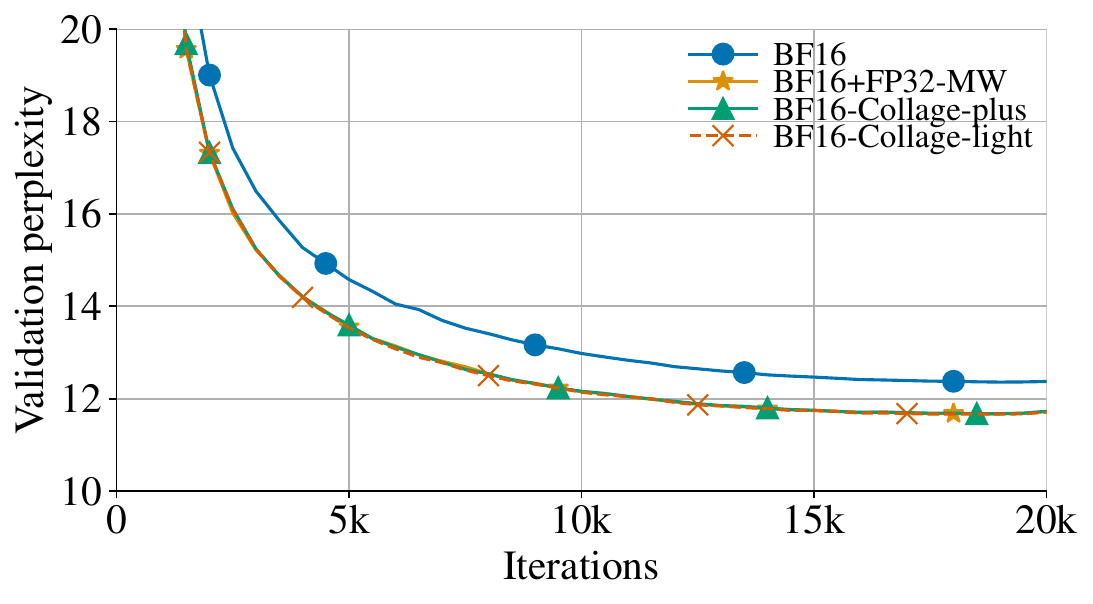}
    }
    \caption{Pretrainnig progress for GPT $1.3$B with settings described in Section\,\ref{ssec:llama_gpt_results} and global batch-size=512, $\beta_2=0.95$. \textbf{Left:} training perplexity vs iterations, and \textbf{right:} validation perplexity vs iterations for different precision strategy listed in Table\,\ref{tab:precision-strategies-breakdown}. The proposed \strname formations consistently match the FP32 master weights with much less memory footprint and faster training.}
    \label{fig:gpt1p3B_beta2_0p95}
\end{figure}

\begin{figure}
    \centering
    \subfigure{
    \includegraphics[width=0.48\linewidth]{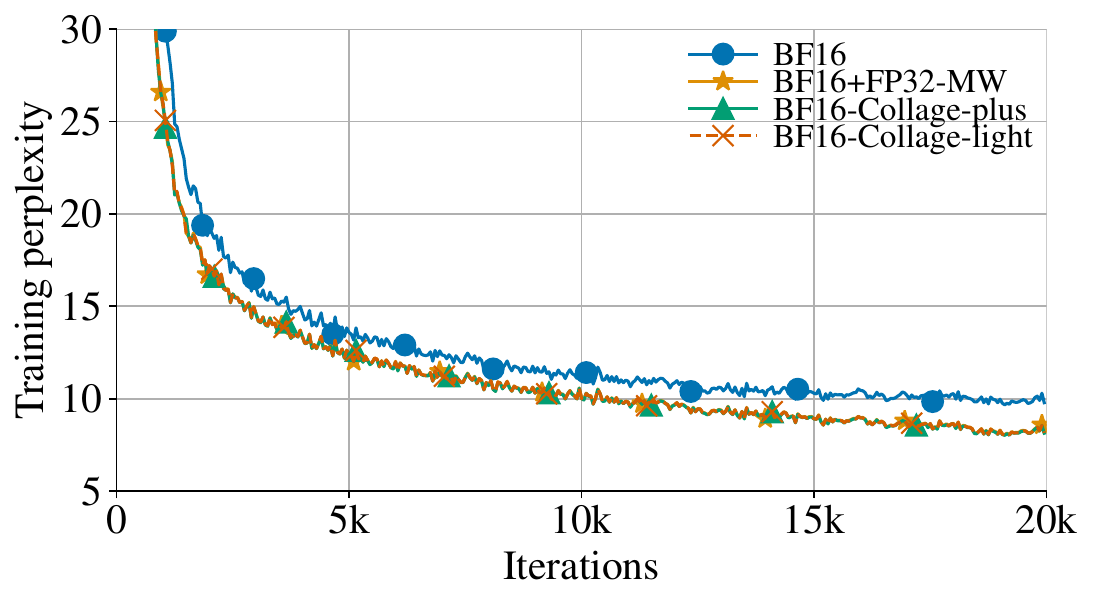}
    }
    ~
    \subfigure{
    \includegraphics[width=0.48\linewidth]{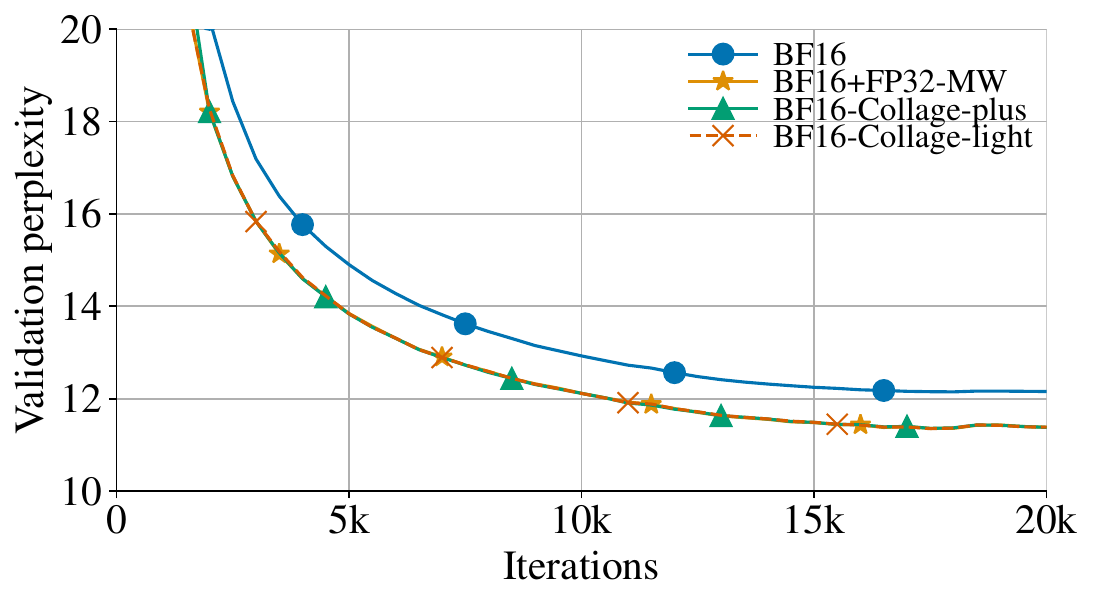}
    }
    \caption{Pretrainnig progress for GPT $2.7$B with settings described in Section\,\ref{ssec:llama_gpt_results} and global batch-size=512, $\beta_2=0.95$. \textbf{Left:} training perplexity vs iterations, and \textbf{right:} validation perplexity vs iterations for different precision strategy listed in Table\,\ref{tab:precision-strategies-breakdown}. The proposed \strname formations consistently match the FP32 master weights with much less memory footprint and faster training.}
    \label{fig:gpt2p7B_beta2_0p95}
\end{figure}

\begin{figure}
    \centering
    \subfigure{
    \includegraphics[width=0.48\linewidth]{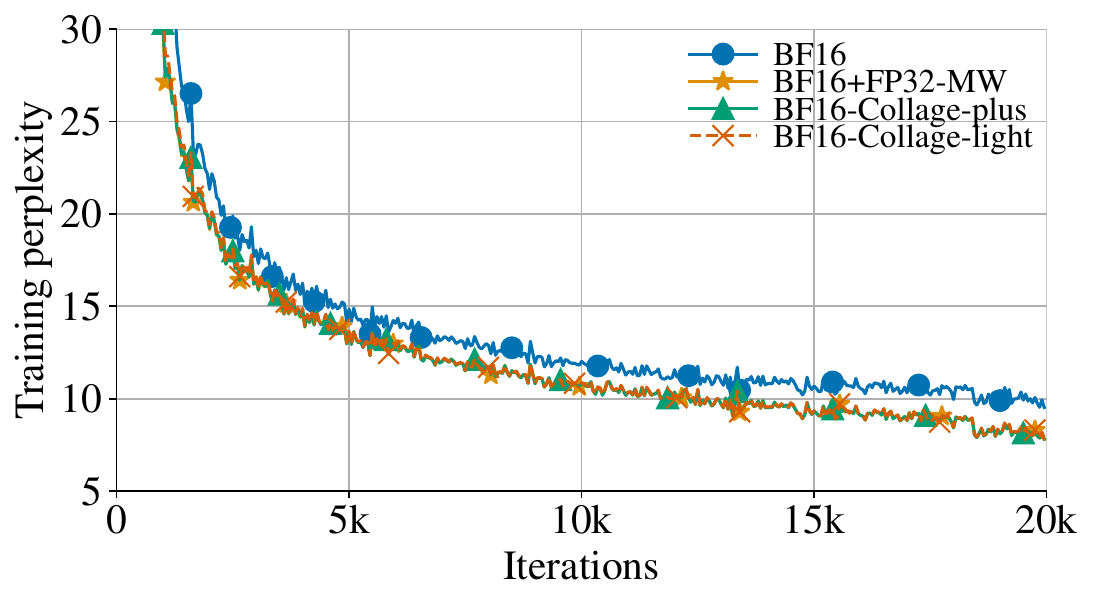}
    }
    ~
    \subfigure{
    \includegraphics[width=0.48\linewidth]{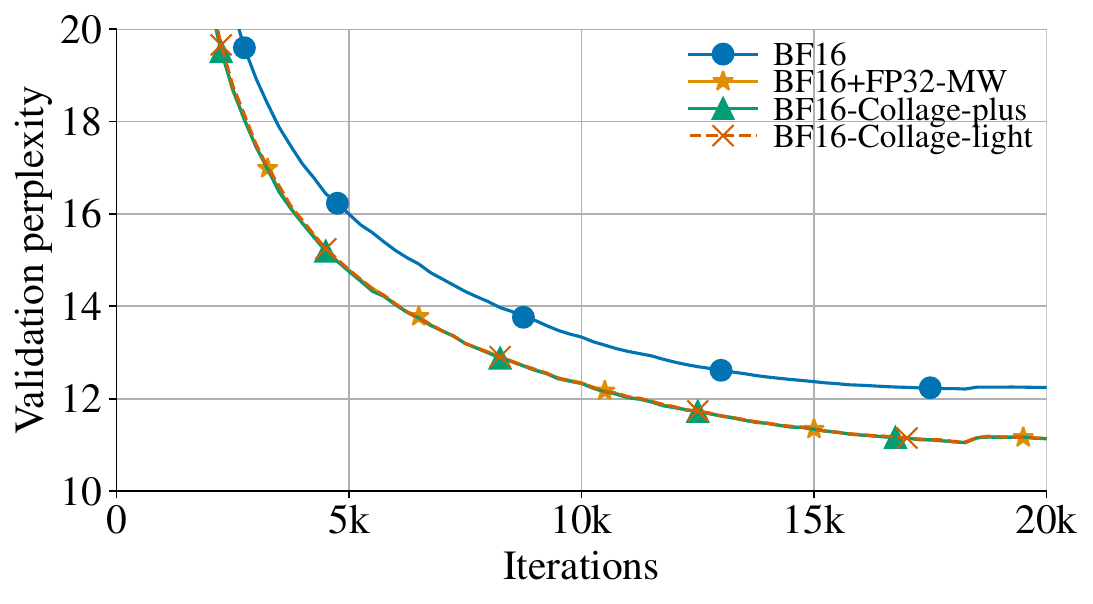}
    }
    \caption{Pretrainnig progress for GPT $6.7$B with settings described in Section\,\ref{ssec:llama_gpt_results} and global batch-size=256, $\beta_2=0.95$. \textbf{Left:} training perplexity vs iterations, and \textbf{right:} validation perplexity vs iterations for different precision strategy listed in Table\,\ref{tab:precision-strategies-breakdown}. The proposed \strname formations consistently match the FP32 master weights with much less memory footprint and faster training.}
    \label{fig:gpt6p7B_beta2_0p95}
\end{figure}

The pretraining progress of GPT $125$M for various settings of $\beta_2$ and global batch sizes is provided in \figurename\,\ref{fig:gpt125M_beta2_0p95_GBS1024}\,\ref{fig:gpt125M_beta2_0p95_GBS2048}\,\ref{fig:gpt125M_beta2_0p99_GBS1024}\,\ref{fig:gpt125M_beta2_0p99_GBS2048}\,\ref{fig:gpt125M_beta2_0p999_GBS1024}\,\ref{fig:gpt125M_beta2_0p999_GBS2048}. For pretraining of GPT $1.3$B, see \figurename\,\ref{fig:gpt1p3B_beta2_0p95}. For pretraining of $2.7$B, see \figurename\,\ref{fig:gpt2p7B_beta2_0p95}. For pretraining of $6.7$B, see \figurename\,\ref{fig:gpt6p7B_beta2_0p95}.

\end{document}